\documentclass[fleqn,10pt]{wlscirep}
\usepackage[utf8]{inputenc}
\usepackage[T1]{fontenc}
\usepackage{bm}
\usepackage{rotating}
\usepackage{marvosym}
\usepackage{multirow}
\usepackage[table]{xcolor}
\title{A Breast Vision Pathology Foundation Model for Real-world Clinical Utility}

\author[1,*]{Yingxue Xu}
\author[2,*]{Zhengyu Zhang}
\author[3,*]{Xiuming Zhang}
\author[4]{Mengwei Xu}
\author[1]{Fengtao Zhou}
\author[1]{Yihui Wang}
\author[1]{Jiabo Ma}
\author[1]{Yi Xin}
\author[2]{Danyi Li}
\author[2]{Chengyu Lu}
\author[2]{Zhijian Cen}
\author[2]{Ying Tan}
\author[5,6]{Qingbing Yao}
\author[6,8]{Qi Wang}
\author[4]{Zizhao Gao}
\author[7]{Yong Zhang}
\author[8]{Jingjing Chen}
\author[9]{Feifei Liu}
\author[10]{Qian Xu}
\author[11]{Yi Dai}
\author[12]{Hongxuan Tan}
\author[1]{Cheng Jin}
\author[1]{Huajun Zhou}
\author[1]{Zhengrui Guo}
\author[1]{Ling Liang}
\author[1]{Hongyi Wang}
\author[13]{Yingcong Chen}
\author[1]{Xi Wang}
\author[14, \Letter]{Zhenhui Li}
\author[12, \Letter]{Ronald Cheong Kin Chan}
\author[6,15,16,17 \Letter]{Ning Mao}
\author[18, \Letter]{Muyan Cai}
\author[4, \Letter]{Zhe Wang}
\author[2, \Letter]{Li Liang}
\author[1,19,20,21,22 \Letter]{Hao Chen}
\affil[1]{Department of Computer Science and Engineering, The Hong Kong University of Science and Technology, Hong Kong SAR, China}
\affil[2]{Department of Pathology, Nanfang Hospital, School of Basic Medical Sciences, Southern Medical University, Guangzhou,  China}
\affil[3]{Department of Pathology, The First Affiliated Hospital, School of Medicine, Zhejiang University, Hangzhou, China}
\affil[4]{State Key Laboratory of Holistic Integrative Management of Gastrointestinal Cancers, Department of Pathology, School of Basic Medicine and Xijing Hospital, Fourth Military Medical University, Xi'an, China}
\affil[5]{School of Information and Electronic Engineering, Shandong Technology and Business University, Yantai, China}
\affil[6]{Big Data and Artificial Intelligence Laboratory, Yantai Yuhuangding Hospital, Qingdao University, Yantai, Shandong, China}
\affil[7]{Department of Pathology, Cancer Hospital of Dalian University of Technology, Cancer Hospital of China Medical University, Liaoning Cancer Hospital \& Institute, Shenyang, Liaoning, China}
\affil[8]{Department of Radiology, Qingdao University Hospital, Qingdao, China}
\affil[9]{Department of Ultrasound, Binzhou Medical University Hospital, Yantai, Shandong Province, China}
\affil[10]{Department of Computed Tomography and Magnetic Resonance, The Fourth Hospital of Hebei Medical University, Shijiazhuang, Hebei, China}
\affil[11]{Department of Medical Imaging, Peking University Shenzhen Hospital, Shenzhen, Guangdong, China}
\affil[12]{Department of Anatomical and Cellular Pathology, The Chinese University of Hong Kong, Hong Kong SAR, China}
\affil[13]{AI Thrust, Information Hub, The Hong Kong University of Science and Technology (Guangzhou), Guangzhou, China}
\affil[14]{Department of Radiology, The Third Affiliated Hospital of Kunming Medical University, Yunnan Cancer Hospital, Kunming, China}
\affil[15]{Department of Radiology, Yantai Yuhuangding Hospital, Qingdao University, Yantai, Shandong, China}
\affil[16]{Shandong Provincial Key Medical and Health Laboratory of Intelligent Diagnosis and Treatment for Women's Diseases, Yantai Yuhuangding Hospital, Qingdao University, Yantai, Shandong, China}
\affil[17]{Faculty of Applied Sciences, Macao Polytechnic University, Macao, China}
\affil[18]{Department of Pathology, State Key Laboratory of Oncology in South China, Guangdong Provincial Clinical Research Center for Cancer, Sun Yat-sen University Cancer Center, Guangzhou, China}
\affil[19]{Department of Chemical and Biological Engineering, Hong Kong University of Science and Technology, Hong Kong SAR, China}
\affil[20]{Division of Life Science, Hong Kong University of Science and Technology, Hong Kong SAR, China}
\affil[21]{HKUST Shenzhen-Hong Kong Collaborative Innovation Research Institute, Futian, Shenzhen, China}
\affil[22]{State Key Laboratory of Nervous System Disorders, The Hong Kong University of Science and Technology, Hong Kong SAR, China}
\affil[*]{Contributed Equally (Co-first)}
\affil[\Letter]{Corresponding Authors}
\affil[ ]{\textbf{Lead Contact: Hao Chen (jhc@ust.hk)}}

\begin{abstract}
Pathology foundation models have shown strong retrospective performance, but whether such systems can support clinically relevant use remains unclear. This challenge is particularly important in breast cancer, where pathological assessment serves as the gold standard for diagnosis and guides treatment planning, surgical decision-making and risk stratification across pre-, intra- and post-operative stages. Here we present \textbf{BRAVE}, a breast-adaptive pathology foundation model developed and evaluated using a total resource of 101,638 breast whole-slide images from 32 sources across Asia, Europe and North America. We assessed BRAVE across 34 tasks in 82 cohorts spanning pre-operative biopsy, intra-operative frozen section and post-operative resection, using an evidence chain comprising retrospective benchmarking, clinically challenging scenarios, workflow-oriented clinical impact simulations, prospective observational validation with the thresholds locked in the retrospective cohorts and crossover pathologist-AI interaction studies. Across these settings, BRAVE supported practical roles in the clinical workflow, including safe exclusion of low-risk cases from routine review, AI-assisted second-review rescue of initially missed positives and prioritization of cases for further assessment. In prospective observational validation across three centres, BRAVE reduced review workload by identifying high-confidence cases that could be safely excluded, including 76.9\% of negative cases in pre-operative malignancy detection at an NPV of 0.953, 70.1\% of negative cases in intra-operative malignancy detection at an NPV of 0.973, while triaging 78.8\% of post-operative subtyping cases as high-confidence, clear-cut cases at an NPV of 1.000. In crossover reader studies, AI assistance improved average balanced accuracy from 88.5\% to 95.1\% (OR=3.14, $P<0.001$), corresponding to more than a threefold increase in the odds of a correct diagnosis, while also improving efficiency, confidence and inter-rater agreement. BRAVE-derived scores further provided independent prognostic information beyond standard clinicopathological assessment for disease-free survival (adjusted HR=4.79, $P<0.001$) and overall survival (adjusted HR=8.14, $P<0.001$). These findings establish a workflow-aligned framework for evaluating breast pathology foundation models across retrospective, prospective and human-AI settings, providing a practical roadmap for translating such systems to support diagnostic, triage and prognostic decisions under clinically relevant conditions.
\end{abstract}
\begin{document}

\flushbottom
\maketitle

\thispagestyle{empty}
\section*{Introduction}
Pathology foundation models have expanded the scope of computational pathology, but their relevance to routine clinical practice remains incompletely established~\cite{xu2024whole,wang2024pathology,vorontsov2024foundation,ma2025generalizable,xu2025multimodal}. Most research has emphasized retrospective performance in curated cohorts, whereas translating these systems into practice requires rigorous, multi-dimensional validation. First, models must demonstrate generalizability across the diverse clinical tasks, institutions and specimen types encountered in real pathological workflows. Second, they must exhibit robustness in clinically challenging scenarios. Third, they must demonstrate actionable clinical utility beyond retrospective benchmarking~\cite{de2023perspectives,you2025clinical,campanella2025clinical}. Addressing these translational gaps is particularly crucial in breast pathology, where histological assessment is central to breast cancer management, serving as the gold standard for diagnosis and guiding treatment planning and risk stratification from pre-operative biopsy to post-operative assessment~\cite{gradishar2024breast}.

In breast cancer care, the pathological tasks encountered at pre-operative biopsy, intra-operative frozen section and post-operative surgical resection serve distinct clinical purposes~\cite{gradishar2024breast}. These settings differ in tissue context, time constraints, diagnostic demands and tolerance for error. Pre-operative biopsies support differential diagnosis, biomarker inference and treatment planning from limited tissue. Intra-operative frozen sections must provide rapid guidance despite preparation-related artefacts~\cite{zhao2025clinical}. Post-operative resections support a broader range of morphological, molecular and prognostic assessments. Accordingly, clinically useful AI may take different forms across the pathway, including second-review support in pre-operative biopsy, safe exclusion from routine review to enable rapid intra-operative decision-making, and case prioritization in post-operative assessment to focus molecular testing on higher-risk cases. However, many AI studies evaluate these scenarios together or focus on only a narrow subset of tasks~\cite{chen2024towards,lu2024visual}, making it difficult to determine whether strong retrospective performance will translate into clinically meaningful utility under the actual workflow. Moreover, such evaluations rarely account for the diverse and challenging scenarios encountered in routine practice, such as preparation artefacts or tissue deformation. These are precisely the conditions under which algorithmic robustness is most severely tested.

Another unresolved issue is whether pathology AI can be validated in a cancer-specific and workflow-aligned manner that is sufficiently close to clinical use. Existing pathology foundation models are often trained on pan-cancer collections~\cite{xu2024whole,wang2024pathology}, which provide breadth but may not adequately represent the spectrum of breast-specific morphology and clinical questions. At the same time, prospective evidence remains limited~\cite{huang2023visual,yan2025pathorchestra,neidlinger2025benchmarking}, and validation rarely extends from retrospective benchmarking to challenging scenarios, clinical impact simulations, prospective observational validation with locked thresholds and pathologist-AI collaboration within the same study. As a result, the clinical value of AI for supporting routine diagnosis, guiding treatment decisions, surgical decision-making and contributing additional prognostic information has not been established consistently across multiple clinical settings~\cite{huang2025pathologist,campanella2025real}.

Here we present \textbf{BRAVE}, a \textbf{BR}e\textbf{A}st \textbf{V}ision pathology foundation mod\textbf{E}l developed as a domain-focused framework to establish clinical utility. Starting from a pancancer pathology foundation model, we further adapted BRAVE using 57,271 breast whole-slide images (WSIs). Across adaptive pretraining and downstream analyses, this study drew on a total resource of over 100K breast WSIs collected from 32 sources across Asia, Europe and North America (Figure~\ref{fig:overview}c and Extended Data Tables~\ref{tab:data_src} and \ref{tab:data_pretrain}). We then structured the evaluation as an evidence chain aligned with the clinical pathway and the depth of pathological inference. We assessed 34 tasks across 82 cohorts (Figure~\ref{fig:overview}d) spanning pre-operative, intra-operative and post-operative settings, from routine morphological diagnosis to treatment planning, molecular prediction and prognosis (Figure~\ref{fig:overview}b). This framework incorporated retrospective benchmarking, analyses of clinically challenging scenarios, retrospective workflow-oriented clinical impact simulations, prospective observational validation with the thresholds locked in the retrospective cohorts, and crossover pathologist-AI interaction (Figure~\ref{fig:overview}a) across the breast pathology workflow, with the aim of testing not only whether the model generalized retrospectively, but also whether it could be examined under conditions closer to practical use.

Using this framework, retrospective clinical impact simulations across pre-operative, intra-operative and post-operative settings supported practical applications including safe exclusion of low-risk cases from routine review, AI-assisted second-review rescue of initially missed positives, and prioritization of higher-risk cases. To confirm these simulated benefits under real-world conditions, in the registered prospective observational validation with locked thresholds across three centres, BRAVE retained clinically actionable utility across multiple workflow settings, saving review workload by safely excluding 76.9\% high-confidence negatives of pre-operative malignancy detection (NPV=0.953), 70.1\% negatives of intra-operative malignancy assessments (NPV=0.973), while 78.8\% of post-operative subtyping cases were triaged as high-confidence, clear-cut cases at an NPV of 1.000. In biomarker second review, 75.0\% of missed Ki67-positive cases were rescued with only a small number of additional review burden (Figure~\ref{fig:pros}). In a crossover reader study covering pre-, intra- and post-operative tasks, AI assistance improved average balanced accuracy from 88.5\% to 95.1\% (OR=3.14, 95\% CI: 2.22--4.46, $P<0.001$), corresponding to more than a threefold increase in the odds of a correct diagnosis, while also improving efficiency, confidence and inter-rater agreement (Figure~\ref{fig:reader_study} and Extended Data Table~\ref{tab:gee}). In prognostic analyses, BRAVE-derived scores provided independent prognostic information beyond standard clinicopathological assessment, remaining significantly associated with disease-free survival (adjusted HR=4.79, 95\% CI: 2.53--7.04, $P<0.001$) and overall survival (adjusted HR=8.14, 95\% CI: 3.90--12.39, $P<0.001$) after adjustment for routine variables (Figure~\ref{fig:surv}). These findings provide a workflow-aligned and clinically grounded framework for evaluating breast pathology foundation models across retrospective, prospective and human-AI settings, establishing a practical roadmap for translating such systems to support diagnostic, triage and prognostic decisions under clinically relevant conditions.

\section*{Results}
\subsection*{Study design and pathological task landscape}
\begin{figure}
    \centering
    \includegraphics[width=1.0\linewidth]{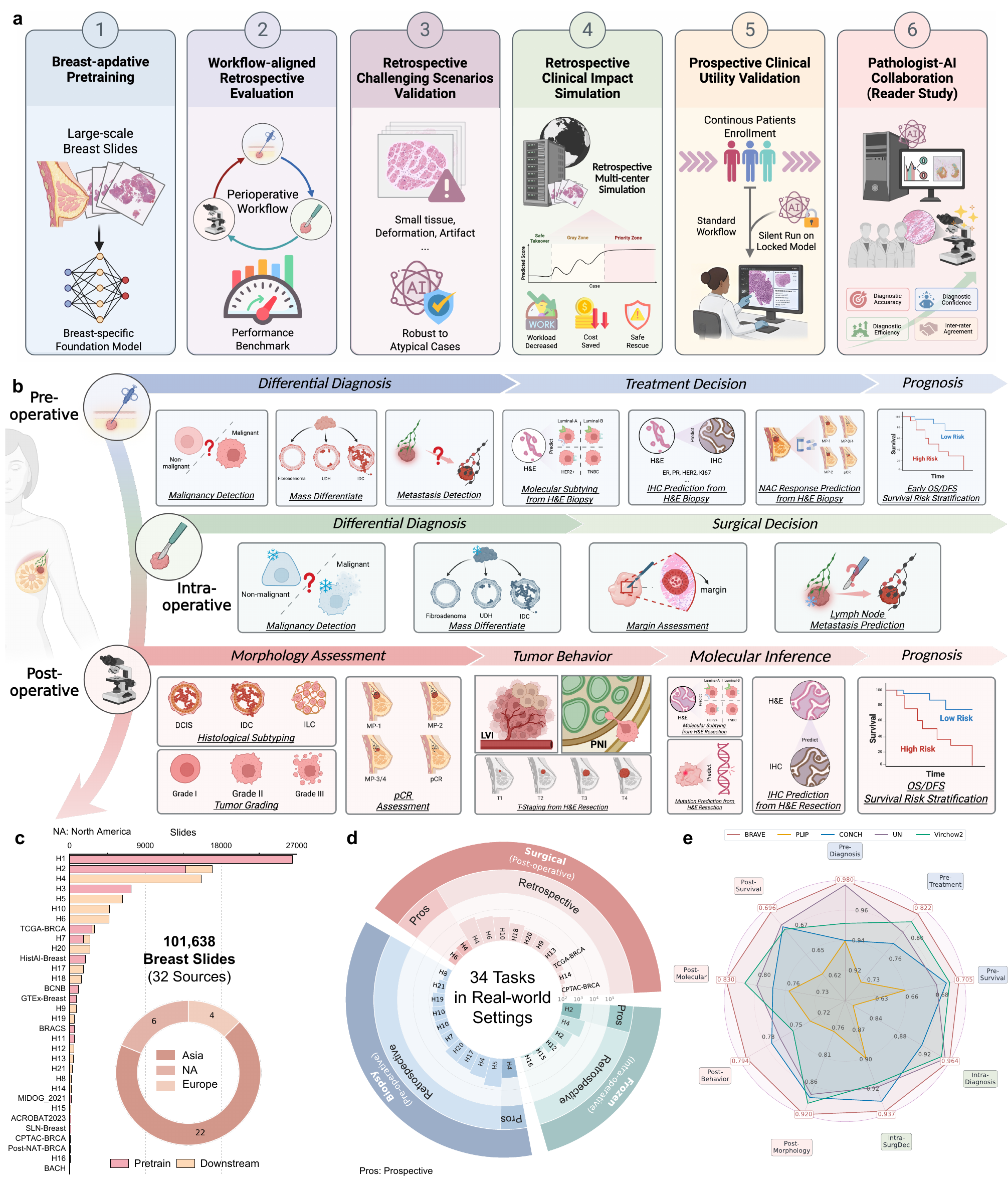}
\end{figure}
\begin{figure}
    \centering
    \caption{\textbf{Overview of BRAVE and study design.} \textbf{a}, Study design comprising breast-adaptive pretraining, workflow-aligned retrospective evaluation, retrospective validation in challenging scenarios, retrospective clinical impact simulation, prospective clinical utility validation, and pathologist-AI collaboration in reader studies. \textbf{b}, Clinical task landscape across pre-operative, intra-operative and post-operative breast cancer care. \textbf{c}, Distribution of 101,638 breast WSIs from 32 sources used for adaptive pretraining and downstream analyses, together with their geographic composition (Asia, North America and Europe). \textbf{d}, Center distribution across 34 real-world tasks, stratified by pre-operative, intra-operative and post-operative stage and by retrospective versus prospective cohorts. \textbf{e}, Comparison of BRAVE with representative pathology foundation models across clinical-purpose categories, summarized by averaged Macro-AUC for classification tasks and C-index for survival prediction.}
    \label{fig:overview}
\end{figure}

\subsubsection*{Study design for pretraining and evidence-chain validation of clinical utility}

We first performed breast-adaptive pretraining using 57,271 breast whole-slide images to obtain the breast-specific BRAVE model. The study was then designed as a stepwise validation framework intended to move from this adaptation stage toward evidence of clinical utility. Specifically, we evaluated BRAVE in workflow-aligned retrospective tasks, examined robustness in challenging scenarios, simulated stage-specific clinical application, and then assessed prospective performance with the thresholds locked in the retrospective cohorts, together with pathologist-AI collaboration in reader studies (Figure~\ref{fig:overview}a). This design was intended to determine not only whether BRAVE achieved strong benchmark performance, but also whether its outputs remained stable, transferable and actionable in settings that more closely approximate real diagnostic workflows.

\subsubsection*{Clinical task landscape and data resources}

The evaluated task landscape spanned pre-operative biopsy assessment, intra-operative frozen-section support, and post-operative resection-based diagnosis and molecular inference, together with prognostic stratification across pre-operative and post-operative settings (Figure~\ref{fig:overview}b). To support this evaluation, we assembled 44,367 breast WSIs and organized them into 34 real-world tasks with retrospective and prospective cohorts across these clinical stages (Figure~\ref{fig:overview}c--d). This breadth allowed BRAVE to be assessed across distinct clinical purposes rather than within a single narrow endpoint, while also enabling comparison with representative pathology foundation models across various clinical tasks (Figure~\ref{fig:overview}e).

\subsection*{Pre-operative evaluation supports diagnosis and treatment planning}
\begin{figure}
    \centering
    \includegraphics[width=1.0\linewidth]{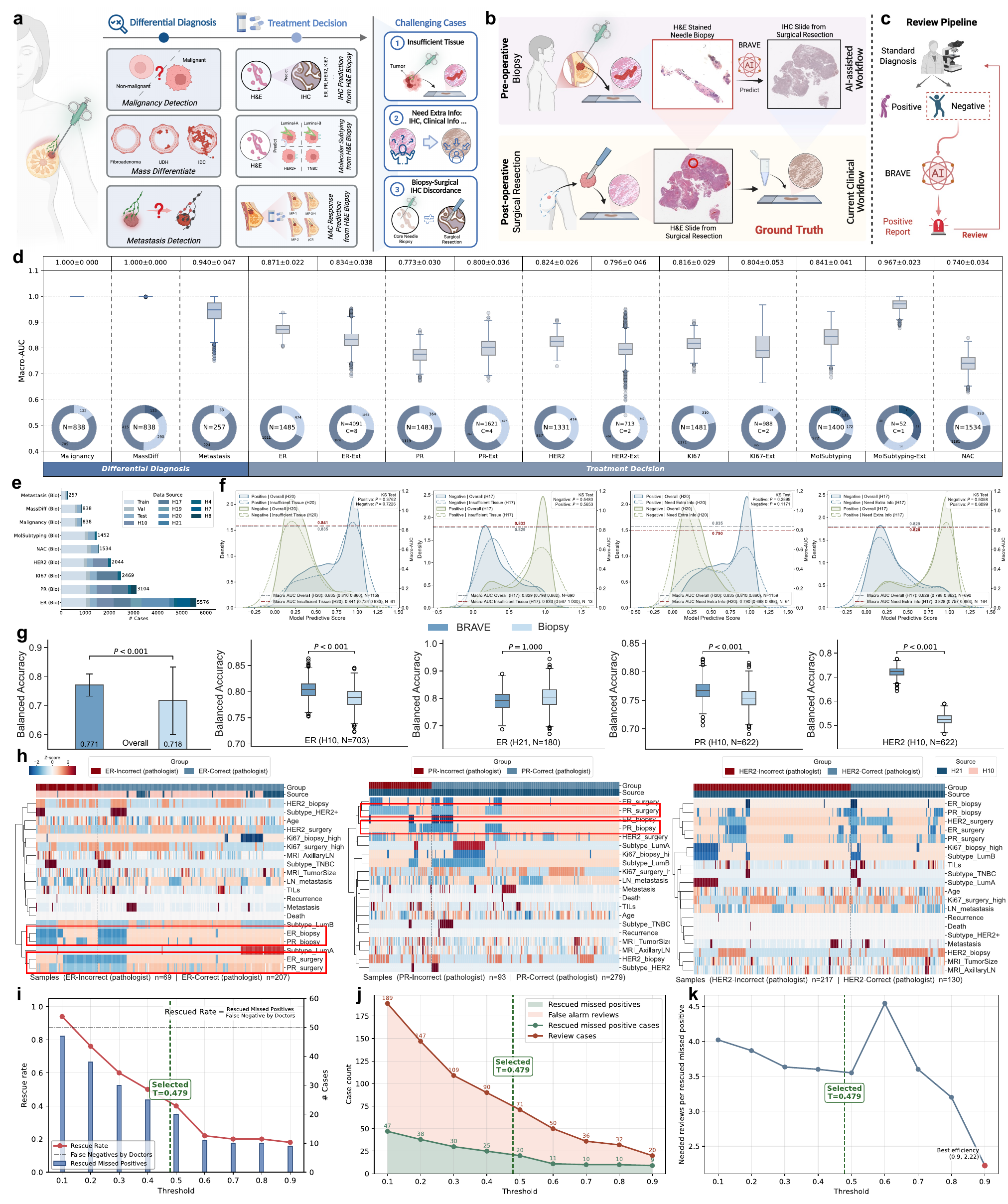}
\end{figure}
\begin{figure}
    \centering
    \caption{\textbf{BRAVE supports pre-operative biopsy decision-making, with simulation of biomarker second review.} \textbf{a}, Pre-operative diagnostic and treatment-related uses of BRAVE, together with representative challenges in biopsy-based assessment. \textbf{b}, AI-assisted workflow for biopsy-based prediction of surgical biomarker status, intended to reduce biopsy-surgical IHC discordance. BRAVE predicts final surgical biomarker status from pre-operative H\&E needle biopsies only. \textbf{c}, Review pipeline for AI-assisted second review, in which biopsy cases initially assessed as negative in routine practice are flagged by BRAVE for review with the aim of rescuing cases that are ultimately positive at surgery. \textbf{d}, Performance of BRAVE on pre-operative tasks in internal and external cohorts. The suffix "-Ext" denotes external validation results. $N$ indicates the sample size (case), and $C$ represents the number of medical centers involved. \textbf{e}, Data distribution for every task across train, val, test and all external cohorts. \textbf{f}, Robustness of BRAVE's performance in ER prediction across the total cohort and the difficult subgroup. A Kolmogorov-Smirnov (KS) test of the prediction scores yielded a $P$-value > 0.05, indicating no statistically significant difference in score distribution and supporting model stability in challenging cases. \textbf{g}, Balanced Accuracy comparison between BRAVE (using H\&E biopsy only) and Biopsy from pathologists (standard H\&E + IHC workflow) for predicting surgical resection biomarker status. Post-operative surgical IHC status served as the ground truth for all patients who underwent direct surgical resection without prior neoadjuvant therapy. \textbf{h}, Exploratory error-pattern analysis comparing biopsy-based pathologist assessment with surgical ground truth. Rows indicate biopsy and surgical biomarker labels and columns indicate cases grouped by whether pathologists were correct or incorrect, showing that pathologist-error groups were enriched for biopsy-positive but surgically lower-expression cases. \textbf{i}--\textbf{k}, Retrospective threshold-based simulation of AI-assisted second review for ER, evaluating whether BRAVE could identify pathologist-missed positive cases under different review burdens. \textbf{i}, Rescue rate across thresholds, with bars indicating the number of rescued missed positives. \textbf{j}, Threshold-dependent trade-off between rescued missed positive cases and total review burden. \textbf{k}, Review efficiency across thresholds, expressed as the number of reviews required per rescued missed positive case. The selected threshold is indicated by the dashed line. Related analyses for PR and HER2 are shown in Extended Data Figure~\ref{fig:pr_her2_prevspost_analysis}.}
    \label{fig:preop}
\end{figure}

We first asked whether BRAVE could provide clinically useful support across the breadth of pre-operative biopsy practice, from routine diagnosis to treatment-related assessment. We then evaluated whether these signals remained informative in challenging biopsy scenarios (Figure~\ref{fig:preop}a). Finally, we simulated a second-review workflow to assess its potential clinical impact (Figure~\ref{fig:preop}b-c). In this pre-operative evaluation, BRAVE was assessed on H\&E-stained core-needle biopsy specimens across diagnostic, IHC biomarker, molecular-subtyping and treatment-response tasks. Across 26 retrospective cohorts from 9 centers, the pre-operative evaluation covered malignancy detection, mass differentiation, metastasis detection, IHC biomarker prediction, molecular subtyping, and NAC response, providing a stage-specific test bed for both benchmark performance and workflow-level utility (Extended Data Tables~\ref{tab:result_pre} and~\ref{tab:pre_cmp}; see Methods - \hyperlink{methods_downstream_eval_cohorts}{Downstream Evaluation Cohorts for Clinical Validation}). Detailed data for all panels in Figure~\ref{fig:preop} are provided in the Extended Data.

Due to different clinical purposes and diagnostic concordance, clinically relevant reference standard differed task from task. Detailed data eligibility, reference standard and label assignment can be found in Section \hyperlink{methods_reference_standard}{Methods - Reference standard and label assignment}. For pre-operative routine diagnostic tasks, ground-truth labels were defined from the biopsy diagnosis itself. For pre-operative molecular prediction tasks, the internal cohorts were restricted to direct-surgery patients where reference labels could be derived from the corresponding post-operative surgical pathology together with ancillary molecular test results (standard workflow). Retrospective external validation cohorts included both direct-surgery and neoadjuvant chemotherapy (NAC)-treated patients. Accordingly, labels were assigned from post-operative resection findings in direct-surgery cohorts and from pre-treatment biopsy assessment in NAC-treated cohorts. This is because NAC can alter morphology and biomarker expression, and thus the post-operative specimen no longer directly reflects the tumour state at the time of biopsy. Therefore, for NAC-treated cases, the pre-treatment biopsy provides the closest available reference to the tumour state at the time of prediction. For NAC response prediction, labels were defined from the final post-operative pathological response assessment. All cases included for analyses are evaluable with definitive reference diagnostic conclusion.

\subsubsection*{Performance benchmark across pre-operative tasks}

Across pre-operative tasks, BRAVE showed strong performance for both routine diagnosis and treatment-related assessment (Figure~\ref{fig:preop}d-e). In internal evaluation, macro-AUC reached 1.000 (95\% CI, 1.000-1.000) for malignancy detection, 1.000 (0.998-1.000) for mass differentiation, and 0.940 (0.820-1.000) for metastasis detection, while treatment-related endpoints also remained strong, including ER prediction at 0.871 (0.826-0.912), PR prediction at 0.773 (0.709-0.829), HER2 prediction at 0.824 (0.771-0.872), Ki67 prediction at 0.816 (0.753-0.873), molecular subtyping at 0.841 (0.756-0.912), and NAC response prediction at 0.740 (0.671-0.805) (Extended Data Table~\ref{tab:result_pre}). These results indicate that a single breast-adaptive model can support heterogeneous pre-operative biopsy tasks ranging from routine differential diagnosis to treatment-related inference.

This performance was also maintained across external cohorts from multiple centers. In the aggregated external analyses shown in Figure~\ref{fig:preop}d, mean macro-AUC was 0.834$\pm$0.038 for ER prediction across 7 retrospective centers, 0.800$\pm$0.036 for PR prediction across 4 centers, and 0.796$\pm$0.046 for HER2 prediction across 2 centers. Molecular subtyping also reached 0.967$\pm$0.023 in H4, although this result should be viewed in the context of the small external H4 cohort. Additionally, individual cohort results were broadly consistent with these aggregated external analyses (Extended Data Tables~\ref{tab:result_pre} and~\ref{tab:pre_cmp}). Taken together, these results support the potential of BRAVE to generalize across diverse biopsy cohorts while preserving clinically relevant performance across both diagnostic and treatment-oriented tasks.

\subsubsection*{Robustness in challenging pre-operative scenarios}

We next asked whether this performance remained stable in biopsy cases that had already been flagged as difficult in routine pathology reports, including specimens with limited tissuen (refer to `Insufficient Tissue' in Figure~\ref{fig:preop}f) and cases in which pathologists considered H\&E-based assessment insufficiently clear for confident interpretation (refer to `Need Extra Info' in Figure~\ref{fig:preop}f). We also examined scenarios in which biomarker discordance may arise from intratumoral heterogeneity (ITH), because core biopsies sample only a limited portion of the lesion and may therefore not fully represent the final surgical specimen (Figure~\ref{fig:preop}a). Notably, all these challenging cases remained evaluable and retained definitive reference labels. For pre-operative challenging cases, the difficulty arose at the initial biopsy, but the final label was assigned from the eventual definitive clinical conclusion established through subsequent diagnostic workup and follow-up. Detailed case eligibility and label-assignment rules are described in Methods - \hyperlink{methods_challenge}{Identification of Challenging Scenarios} and \hyperlink{methods_reference_standard}{Reference standard and label assignment}.

We used ER prediction as the primary robustness example because it was evaluated across the largest number of centers and provided sufficient difficult cases for subgroup analysis (Figure~\ref{fig:preop}f). In these ER analyses, subgroup performance remained close to that of the full cohort. For `Insufficient Tissue' cases in H20, the subgroup macro-AUC was 0.841 (95\% CI, 0.724-0.933), compared with 0.835 (0.810-0.860) in the full cohort. For `Need Extra Info' cases in H17, the subgroup macro-AUC was 0.828 (0.757-0.893), compared with 0.829 (0.798-0.862) in the full cohort. Kolmogorov-Smirnov testing also showed no significant shift in score distribution in either comparison ($P>0.05$ for all comparisons, Extended Data Table~\ref{tab:ks_stat_pre}). This statistical equivalence indicates that the model's predictive ability did not degrade when faced with these specific challenges.

These findings suggest that BRAVE remained informative even in biopsy cases that had been considered difficult in routine practice. This robustness is important because the intended clinical role of a pre-operative system is not limited to ideal biopsy specimens. Its practical value lies in remaining informative in challenging cases where tissue is limited, or pathologists would otherwise consider these cases difficult and require additional IHC workup or further evaluation. The subgroup analyses therefore suggest that BRAVE may retain useful signals even in more ambiguous biopsy-based scenarios.

Beyond these predefined difficult subgroups, pre-operative biopsy assessment can also be challenged by biopsy-surgical discordance arising from limited sampling and intratumoral heterogeneity. This represents a distinct but clinically important source of uncertainty not captured by routine difficult-case labels alone. We therefore next examined whether BRAVE could better align biopsy-stage assessment with the eventual surgical reference and whether this could support AI-assisted second review before treatment.

\subsubsection*{Biopsy second review to reduce missed positives before treatment}

In clinical practice, routine biopsy-based biomarker assessment may still miss cases that are ultimately positive at surgery. One reason is that limited biopsy sampling may not fully capture intratumoral heterogeneity. As a result, the biopsy specimen may not fully represent the tumour's overall biomarker landscape, leading to biopsy-surgical discordance\cite{li2025her2discordance} and delayed identification of truly positive patients for treatment planning (Figure~\ref{fig:preop}a). To assess whether BRAVE could match or improve upon routine clinical practice, we compared its biomarker predictions based on H\&E biopsy with pathologists' standard workflow (H\&E and IHC), using surgical biomarker status as the reference. Results in Figure~\ref{fig:preop}g (Extended Data Table~\ref{tab:res_cmp_human_ai}) showed that BRAVE using H\&E alone achieved higher overall balanced accuracy for predicting surgical biomarker status (0.771$\pm$0.031 versus 0.718$\pm$0.113, $P<0.001$). To understand why biopsy- based assessment from pathologists performed less well, we performed an exploratory error-pattern analysis (Figure~\ref{fig:preop}h), in which cases were grouped according to whether the pathologist's biopsy-based assessment agreed with the final surgical result. We found that, in discordant cases (Incorrect groups), when the surgical specimen (ground-truth) was ultimately low-expression or negative, pathologists tended to overestimate the likelihood of positivity on the biopsy, possibly reflecting a cautious interpretation on biopsy section with limited tissues. This pattern is consistent with the lower balanced accuracy observed for pathologists, because balanced accuracy gives equal weight to sensitivity and specificity, and systematic overcalling of positivity reduces performance on the low-expression or negative side.

To reduce biopsy-surgical discordance, we designed an AI-assisted second-review pipeline (Figure~\ref{fig:preop}c) to test whether BRAVE could help rescue missed positive cases before treatment, because positive biomarker signals can guide treatment selection and missing them may delay timely therapy (Figure~\ref{fig:preop}i-k). In this AI-assisted second-review pipeline (Figure\ref{fig:preop}c), cases initially assessed as negative in routine practice were re-evaluated by BRAVE, where those with prediction scores above a certain threshold were flagged for review. We then retrospectively simulated this workflow across a range of candidate thresholds to evaluate the trade-off between rescue rate (the proportion of missed positive cases that would be rescued), review burden (the proportion of cases that would require review), and review efficiency (the number of reviews needed per rescued case). Under predefined constraints requiring a rescue rate of at least 40\% while limiting review burden to at most 40\%, the selected threshold of $T=0.479$ achieved a rescue rate of 44.0\% at a review burden of 35.7\%, corresponding to 22 rescued false-negative cases with 74 review cases (Extended Data Table~\ref{tab:threshold_er_pr}). 3.4 cases needed to be reviewed to rescue one missed positive case. These analyses support evaluating biopsy second review as a plausible second-review strategy in clinical workflow for BRAVE, in which model-triggered review may help rescue surgically positive biomarker cases that would otherwise be missed in routine practice and thereby reduce the risk of delayed treatment. More analyses for PR and HER2 are shown in Extended Data Figure~\ref{fig:pr_her2_prevspost_analysis}. Compared with ER, PR showed broadly similar threshold-dependent trade-offs, whereas HER2 used a higher selected threshold and required fewer reviews per rescued missed positive case. This is consistent with prior reports that HER2 assessment is particularly prone to biopsy-surgical discordance, reflected in contemporary HER2 assessment guidance~\cite{li2025her2discordance,wolff2023her2}.
\subsection*{Intra-operative evaluation supports time-critical surgical decisions}
\begin{figure}
    \centering
    \includegraphics[width=1.0\linewidth]{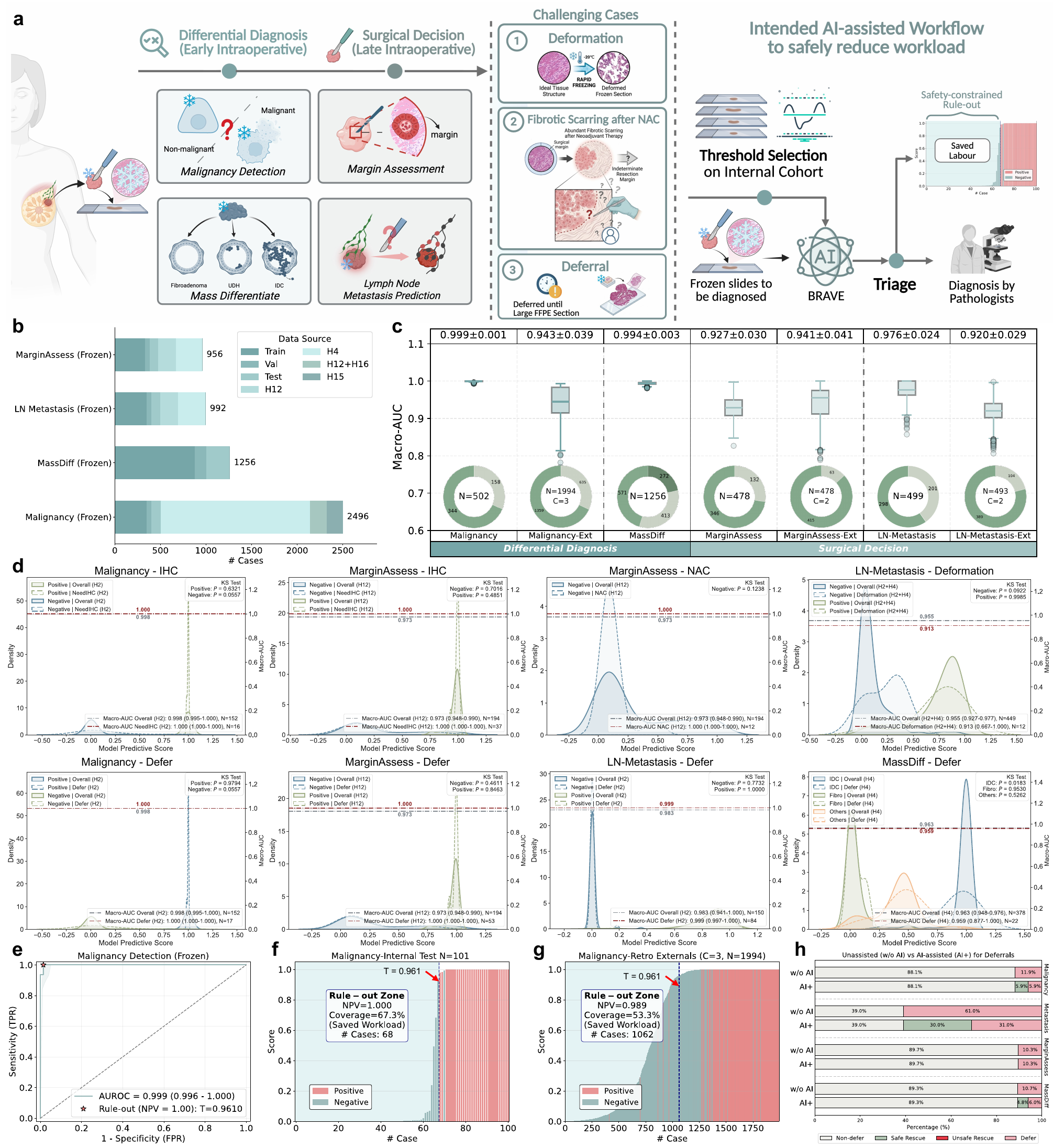}
\end{figure}
\begin{figure}
    \centering
    \caption{\textbf{BRAVE supports intra-operative frozen-section diagnosis, with simulation of safety-constrained rule-out triage.} \textbf{a}, Clinical roles of BRAVE during the intra-operative stage, spanning early differential diagnosis and later surgical decision support, together with representative frozen-section challenges including tissue deformation during rapid freezing, fibrotic scarring after neoadjuvant chemotherapy (NAC), and cases deferred in routine practice. The panel also illustrates the intended AI-assisted workflow, in which thresholds selected on the internal cohort are used to triage frozen slides and safely reduce workload by ruling out low-risk cases before pathologist review. \textbf{b}, Data distribution for each intra-operative task across training, validation, test, and external cohorts. \textbf{c}, Performance of BRAVE on intra-operative tasks in internal and external cohorts. The suffix "-Ext" denotes external validation results. $N$ indicates the sample size (case), and $C$ represents the number of medical centers involved. \textbf{d}, Robustness of BRAVE in representative challenging intra-operative scenarios, including cases requiring additional IHC workup, NAC-associated fibrosis, deformation-related frozen-section artifacts, and cases deferred in routine practice. Density plots show predictive score distributions for the overall cohorts and their corresponding difficult subgroups, with Kolmogorov-Smirnov (KS) test results shown in each panel. \textbf{e}, Receiver operating characteristic curve for threshold selection to safely rule out low-risk cases in the malignancy-detection task, with the selected operating threshold marked to satisfy a safety-constrained negative predictive value. \textbf{f}, Case-level internal-test simulation of low-risk case rule-out in the malignancy-detection task, showing the threshold-defined rule-out zone, the resulting negative predictive value, and workload coverage. \textbf{g}, External validation of the same low-risk case rule-out threshold for the malignancy-detection task across all retrospective external cohorts, showing preserved rule-out performance and coverage. \textbf{h}, Task-level comparison between the conventional frozen-section workflow and AI-assisted triage in subsets with deferral annotations. Stacked bars partition cases into non-deferred cases, safely rescued cases that could be ruled out by BRAVE, unsafe rescues, and cases that would still require deferral to routine pathologist review.}
    \label{fig:intraop}
\end{figure}

We then turned to the intra-operative stage, where frozen-section interpretation must support rapid diagnosis and immediate surgical decisions despite frequent freezing-related artifacts (Figure~\ref{fig:intraop}a). In this setting, BRAVE was evaluated directly on the original digitized rapid-frozen H\&E slides rather than on reprocessed permanent FFPE sections derived from the frozen tissue remnants. Postoperative paraffin-section diagnosis was used as the reference standard because it provides the final definitive pathological assessment. Across 11 retrospective cohorts from 5 centers, this evaluation covered malignancy detection, mass differentiation, margin assessment, and lymph node metastasis prediction, allowing us to examine overall task performance, robustness in difficult frozen-section cases, and the potential utility of safety-constrained rule-out triage (Extended Data Tables~\ref{tab:result_intra} and~\ref{tab:intra_cmp}). Detailed data underlying all panels in Figure~\ref{fig:intraop} are provided in the Extended Data.

\subsubsection*{Performance benchmark across intra-operative tasks}

For intra-operative tasks, BRAVE performed strongly in both early differential diagnosis and later surgical decision support (Figure~\ref{fig:intraop}b-c). In internal evaluation, macro-AUC reached 0.999 (95\% CI, 0.996-1.000) for malignancy detection, 0.994 (0.986-0.999) for mass differentiation, 0.927 (0.862-0.981) for margin assessment, and 0.976 (0.923-1.000) for lymph node metastasis prediction (Extended Data Table~\ref{tab:result_intra}). The slightly lower performance for margin assessment than for malignancy detection is clinically expected. As described in Methods, malignancy detection is performed earlier in the intra-operative workflow to diagnose the main tumor mass, whereas margin assessment is performed later on the surgical resection edges, where the target lesion may be limited to minute residual tumor foci. Despite this higher level of difficulty, BRAVE still maintained strong performance for margin assessment. These results suggest that BRAVE can support multiple intra-operative frozen-section tasks, from rapid lesion classification to surgical decision support.

A similar pattern was seen in external frozen-section cohorts. In the aggregated analyses shown in Figure~\ref{fig:intraop}c, mean macro-AUC was 0.943$\pm$0.039 for malignancy detection across 3 retrospective centers, 0.941$\pm$0.041 for margin assessment across 2 centers, and 0.920$\pm$0.029 for lymph node metastasis prediction across 2 centers. BRAVE was often the best-performing or tied-best foundation model in these retrospective comparisons (Extended Data Tables~\ref{tab:result_intra} and~\ref{tab:intra_cmp}). External validation therefore points to reasonably consistent performance across distinct frozen-section cohorts.

\subsubsection*{Robustness in challenging intra-operative scenarios}

We then examined whether these signals remained informative in frozen-section cases that were difficult in routine practice, including deformation-related artifacts (refer to `Deformation' in Figure~\ref{fig:intraop}d), neoadjuvant chemotherapy-associated fibrosis (refer to `NAC' in Figure~\ref{fig:intraop}d), and cases deferred by pathologists (refer to `Defer' in Figure~\ref{fig:intraop}d) during intra-operative assessment (Figure~\ref{fig:intraop}a). We also evaluated subgroups in which routine interpretation required additional IHC workup (refer to `NeedIHC' in Figure~\ref{fig:intraop}d). Notably, all these challenging intra-operative cases remained evaluable, and their reference labels were defined from the final diagnosis established on subsequent permanent paraffin sections. These subgroups capture common sources of uncertainty in frozen-section assessment encountered in routine practice (see Methods - \hyperlink{methods_challenge}{Identification of Challenging Scenarios} and \hyperlink{methods_reference_standard}{Reference standard and label assignment}).

Across difficult cases flagged in routine pathology reports, performance generally remained close to that of the corresponding full cohorts (Figure~\ref{fig:intraop}d). For example, in H12 margin assessment, the NeedIHC subgroup reached a macro-AUC of 1.000 (95\% CI, 1.000-1.000) versus 0.973 (0.948-0.990) in the overall cohort, with no significant shift in score distribution by Kolmogorov-Smirnov testing ($P=0.7016$ for negative cases and $P=0.4851$ for positive cases). In deformation-associated lymph-node cases, subgroup macro-AUC was 0.913 (0.667-1.000) versus 0.955 (0.927-0.977) in the full cohort, again without significant score-distribution differences ($P=0.0922$ for negative cases and $P=0.9985$ for positive cases; Extended Data Table~\ref{tab:ks_stat_intra}). Defer-related analyses were broadly similar, with macro-AUCs usually remaining close to those of the full cohorts. In deferred mass-differentiation cases that represent a harder-to-interpret subset for pathologists, although the IDC score distribution slightly differed from that of the full cohort ($P=0.0183$), subgroup macro-AUC remained similar (0.959 versus 0.963). Overall, BRAVE remained informative with high performance in many frozen-section cases considered difficult in routine practice, although the score-distribution shifts seen in some smaller deferred subsets suggest room for further improvement in finer-grained tasks such as mass differentiation.

\subsubsection*{Safety-constrained rule-out triage to reduce intra-operative workload}

In our routine practice, intra-operative pathological review can account for a substantial portion of operative waiting time, sometimes approaching one-third to one-half, so reducing diagnostic turnaround could accelerate surgical decision-making and help mitigate risks associated with prolonged intra-operative waiting. Given such strict time constraints in intra-operative frozen-section assessment, we asked whether BRAVE could safely rule out low-risk cases and thereby reduce immediate review burden (Figure~\ref{fig:intraop}a). We first selected an internal threshold that let BRAVE rule out cases only when no malignant cases were missed, which corresponded to a rule-out NPV of 1.000 and an operating threshold of $T=0.961$ in the internal test cohort (Figure~\ref{fig:intraop}e). At this threshold, malignancy rule-out covered 67.3\% of internal cases (68 of 101) without false-negative rule-out cases, and external validation retained a rule-out NPV of 0.989 while still covering 53.3\% of cases (1,062 of 1,994) across 3 retrospective centers (Figure~\ref{fig:intraop}f-g and Extended Data Table~\ref{fig:intra_rule_out_results}). This level of coverage suggests that pathologist slide-review workload could be substantially reduced while maintaining safety.

The broader rule-out analysis in Extended Data Figure~\ref{fig:intra_all_rule_out} indicates that the same safety-constrained framework can also be applied to lymph node metastasis prediction, margin assessment, and mass differentiation, although the achievable workload reduction differs substantially by task. For example, external rule-out coverage reached 68.5\% for lymph node metastasis prediction but only 8.8\% for margin assessment under the selected thresholds. This lower coverage for margin assessment is clinically reasonable, because margin status directly informs the need for additional intra-operative resection. As a result, a more conservative threshold for margin assessment may therefore be appropriate. Furthermore, margin assessment is intrinsically challenging because it requires detecting minute residual tumor clusters at the surgical edges, making high-confidence rule-out fundamentally more difficult. Beyond task-level workload reduction, this triage framework also demonstrated potential in diagnosing difficult cases. Specifically, in subsets with deferral annotations, AI-assisted triage reclassified some cases that would otherwise have been deferred into a safely ruled-out group (Figure~\ref{fig:intraop}h), suggesting potential to support definitive intra-operative decision-making rather than postponing decision until postoperative review and a second operation. Overall, these results suggest that BRAVE may help reduce intra-operative review workload and accelerate surgical decision-making by safely ruling out low-risk cases.
\subsection*{Post-operative evaluation extends from morphology to molecular inference}
\begin{figure}
    \centering
    \includegraphics[width=1.0\linewidth]{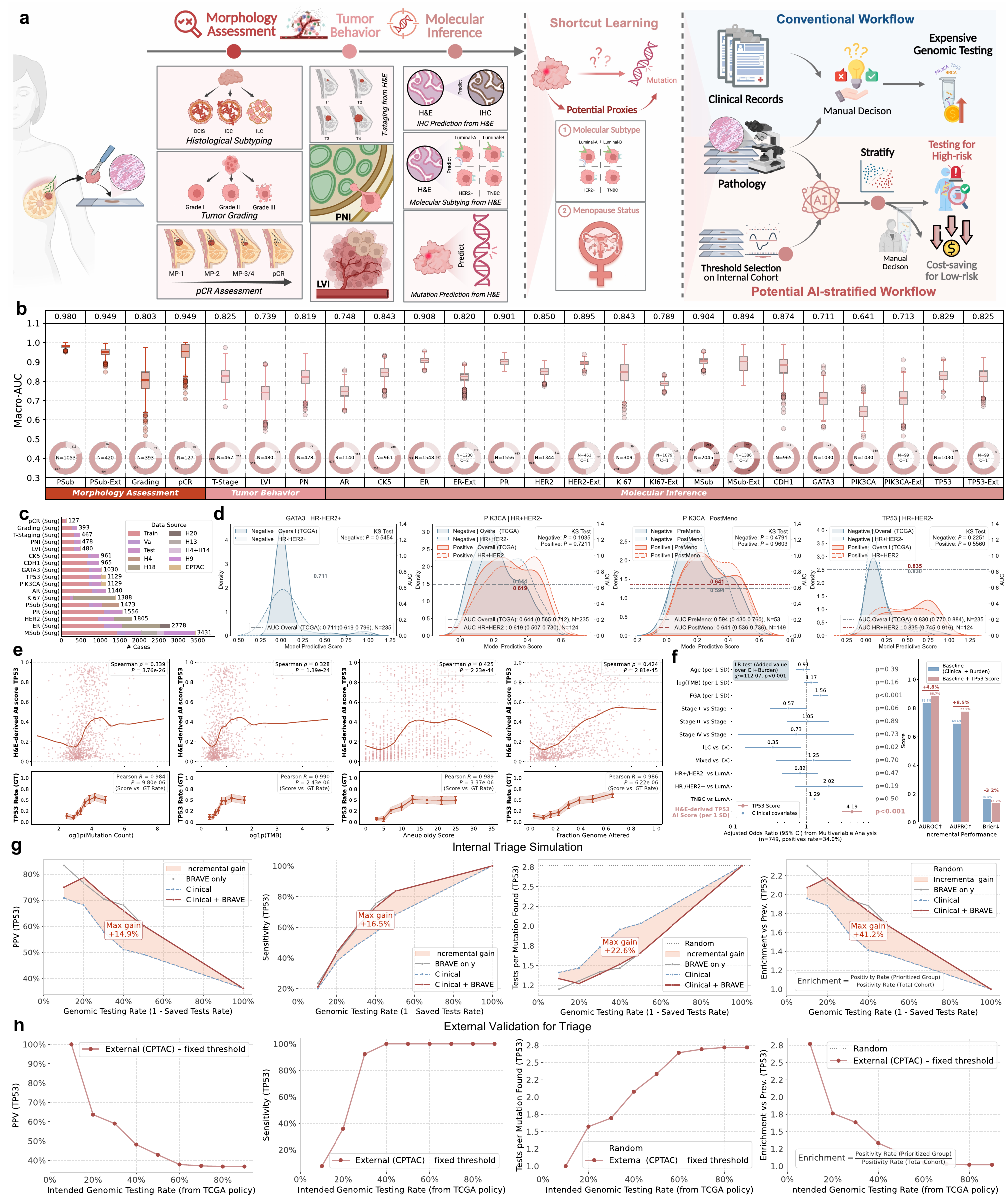}
\end{figure}
\begin{figure}
    \caption{\textbf{BRAVE supports post-operative pathology assessment and AI-guided prioritization of molecular testing.} \textbf{a}, Clinical roles of BRAVE during the post-operative stage, spanning morphology assessment, tumor-behavior assessment, and molecular inference from surgical specimens. The panel also illustrates a potential shortcut-learning concern in mutation prediction and contrasts the conventional workflow for costly genomic testing with a potential AI-stratified workflow, in which BRAVE-derived scores are used to prioritize high-risk cases for testing while reducing unnecessary testing in lower-risk cases. \textbf{b}, Performance of BRAVE on post-operative tasks in internal and external cohorts. Tasks are grouped into morphology assessment, tumor behavior, and molecular inference. The suffix "-Ext" denotes external validation results. $N$ indicates the sample size (case), and $C$ represents the number of medical centers involved. \textbf{c}, Data distribution for each post-operative task across training, validation, test, and external cohorts. \textbf{d}, Robustness of representative post-operative molecular-inference tasks in clinically relevant subgroups. Density plots show predictive score distributions for the overall cohorts and subgroup cohorts, with Kolmogorov-Smirnov (KS) test results shown in each panel. \textbf{e}, Biological association between H\&E-derived AI scores from BRAVE and genomic-instability-related variables. Top, scatter plots with smoothed trends showing the relationship between BRAVE-derived TP53 scores and mutation count, tumor mutational burden, aneuploidy score, and fraction genome altered. Bottom, binned comparisons showing concordance between AI score trends and ground-truth TP53 mutation rates. \textbf{f}, Independent predictive value validation of the H\&E-derived TP53 score beyond clinicopathologic variables and genomic burden. Left, adjusted odds ratios from the aggregate model including clinical covariates, genomic burden, and the AI-derived TP53 score. Right, incremental predictive performance after adding the AI score to the baseline model. \textbf{g,h}, Internal simulation and external validation of TP53 genomic-testing prioritization. Panel g compares BRAVE alone, current clinical stratification, and the integrated clinical + BRAVE strategy across testing rates, with shaded regions indicating incremental gain from integrating BRAVE with the clinical baseline. Panel h applies fixed thresholds derived from the BRAVE-only ranking in the internal TCGA cohort to CPTAC. Curves show positive predictive value, sensitivity, tests required per mutation found, and enrichment over cohort prevalence. Panels e-h are shown using TP53 as an illustrative example.}
    \label{fig:postop}
\end{figure}

We finally examined whether BRAVE could support post-operative interpretation of surgical resection specimens, spanning routine morphologic assessment, inference of tumor behavior, and molecularly related prediction, and whether these H\&E-derived signals could be used to prioritize downstream genomic testing (Figure~\ref{fig:postop}a). Across 27 retrospective cohorts drawn from multiple centers and public datasets, the post-operative benchmark included pathological subtyping, grading, pCR assessment, T-stage prediction, invasion-related features, biomarker prediction, molecular subtyping, and mutation prediction, thereby covering both established surgical pathology readouts and treatment-related molecular inference (Extended Data Tables~\ref{tab:result_post} and~\ref{tab:post_cmp}). Detailed data underlying all panels in Figure~\ref{fig:postop} are provided in the Extended Data.

\subsubsection*{Performance benchmark across post-operative tasks}

On post-operative tasks, BRAVE showed stronger performance in morphology assessment and in several biomarker- and subtype-related inference tasks, whereas tumor-behavior and mutation-prediction endpoints were more heterogeneous (Figure~\ref{fig:postop}b-c). In internal evaluation, macro-AUC reached 0.980 (95\% CI, 0.962-0.993) for pathological subtyping, 0.949 (0.841-1.000) for pCR assessment, 0.908 (0.875-0.938) for ER prediction, 0.901 (0.866-0.936) for PR prediction, and 0.904 (0.867-0.935) for molecular subtyping. Mutation-prediction performance was likewise variable, with macro-AUCs of 0.874 (95\% CI, 0.745-0.953) for CDH1, 0.829 (0.770-0.884) for TP53, and 0.641 (0.565-0.712) for PIK3CA. Tumor-behavior tasks such as T-stage prediction, lymphovascular invasion, and perineural invasion reached macro-AUCs of 0.825 (0.738-0.901), 0.739 (0.615-0.848), and 0.819 (0.719-0.903), respectively (Extended Data Table~\ref{tab:result_post}). These results suggest that post-operative H\&E provides stronger signal for morphologic, biomarker-related, and some mutation-prediction tasks than for tumor-behavior endpoints. Even for the more difficult endpoints, the model still retained biologically and clinically informative signal (see Figure~\ref{fig:postop}d--h).

A similar pattern was seen in external cohorts. Pathological subtyping remained strong in H4 (macro-AUC, 0.949; 95\% CI, 0.907-0.984), HER2 prediction reached 0.895 (0.866-0.922) in H18, molecular subtyping ranged from 0.842 (0.801-0.878) in H9 to 0.937 (0.888-0.974) in H4+H14, TP53 mutation prediction reached 0.825 (0.733-0.903) in CPTAC, and PIK3CA mutation prediction reached 0.713 (0.595-0.814) in CPTAC (Extended Data Table~\ref{tab:result_post}). BRAVE was also frequently the best-performing or tied-best foundation model in retrospective comparisons (Extended Data Table~\ref{tab:post_cmp}). These external results suggest that BRAVE generalizes across several post-operative tasks, where performance is generally stronger for tasks with clearer morphologic correlates in H\&E.

\subsubsection*{Robustness in challenging post-operative scenarios}
Given that mutation-prediction tasks showed measurable predictive ability in the post-operative benchmark, we next examined them more closely, as AI provides distinct added value by inferring molecular alterations that pathologists cannot routinely identify during manual review. We thus asked whether BRAVE retained predictive discrimination within clinically relevant subgroup contexts rather than mainly serving as a subtype proxy (Figure~\ref{fig:postop}d). This question was important because several mutation targets evaluated here are non-randomly distributed across breast-cancer subtypes: PIK3CA alterations are frequently observed in hormone-receptor-positive disease~\cite{griffith2018erpositive,kalinsky2009pik3ca}, GATA3 alterations are enriched in luminal or ER-positive tumors~\cite{griffith2018erpositive}, and TP53 shows marked variation across breast-cancer subtypes~\cite{tcga2012breast}. Figure~\ref{fig:postop}d showed that, for PIK3CA, performance remained similar within HR+HER2- tumors, with macro-AUC of 0.619 (95\% CI, 0.507-0.730) versus 0.644 (0.565-0.712) in the full TCGA cohort. Menopausal status is another clinically relevant source of heterogeneity for PIK3CA in breast cancer~\cite{chollet2016breast}, so we additionally examined premenopausal and postmenopausal subgroups, in which macro-AUCs remained comparable at 0.594 (0.430-0.760) and 0.641 (0.536-0.736), respectively. These subgroup comparisons showed no significant shift in score distribution by Kolmogorov-Smirnov testing (all $P>0.10$). A similar pattern was observed for TP53 in the HR+HER2- subgroup, where macro-AUC was 0.835 (0.745-0.916) versus 0.830 (0.770-0.884) in the full cohort, again without significant score-distribution differences ($P=0.2251$ for the negative class and $P=0.5560$ for the positive class; Extended Data Table~\ref{tab:ks_stat_post}). We also examined GATA3 in an HR-HER2+ subgroup, where GATA3 alterations are less typically enriched than in luminal or ER-positive disease. Although positive cases were too sparse in this subgroup for a stable subgroup AUC estimate, the negative-class score distribution still showed no significant shift relative to the full cohort ($P=0.5454$). Overall, these analyses suggest that BRAVE captured morphologic information associated with mutation status, rather than simply relying on breast-cancer subtype as a proxy.

To further test whether BRAVE-derived mutation signals reflected broader biological structure rather than only shortcut proxies, we next took TP53 as an illustrative example because the biological correlates of p53 dysfunction are relatively well documented, making it appropriate to test whether the model had learned these known associations~\cite{tcga2012breast,levine2020p53,hertel2025p53mitotic}. We therefore examined whether the H\&E-derived TP53 score tracked genomic features associated with p53 dysfunction (Figure~\ref{fig:postop}e). As loss of p53 function is often accompanied by increased genomic burden, typically reflected by more mutations, higher mutational load, and broader chromosomal alteration, we asked whether the H\&E-derived AI TP53 score reflected corresponding associations with mutation count, tumor mutational burden, aneuploidy score, and fraction genome altered. The H\&E-derived AI TP53 score showed moderate positive associations with mutation count (Spearman $\rho=0.339$, $P=3.76\times10^{-26}$), tumor mutational burden ($\rho=0.328$, $P=1.39\times10^{-24}$), aneuploidy score ($\rho=0.425$, $P=2.23\times10^{-44}$), and fraction genome altered ($\rho=0.424$, $P=2.81\times10^{-45}$), while binned score trends remained highly concordant with ground-truth TP53 mutation rates (Pearson $R=0.984$ to 0.990; all $P<10^{-5}$; Extended Data Table~\ref{tab:trend_tp53}). These results showed that higher BRAVE-derived TP53 scores were consistently associated with higher mutation count, tumor mutational burden, aneuploidy score, and fraction genome altered, with the very small $P$ values indicating that these positive associations were statistically significant. This pattern suggests that the model was capturing broader consequences of TP53 dysfunction rather than simply matching mutation labels.

We then asked whether the H\&E-derived AI TP53 score remained an independent predictor beyond standard clinicopathologic variables and genomic burden (Figure~\ref{fig:postop}f). Multivariable analysis showed that it remained independently associated with TP53 mutation status after adjustment for age, genomic burden, stage, histologic subtype, and molecular subtype (odds ratio, 4.19; 95\% CI, 3.12-5.61; $P<0.001$), and significantly improved the baseline model by likelihood-ratio testing ($\chi^2=112.07$, $P<0.001$), with AUROC and AUPRC gains of 4.8\% and 8.5\%, respectively (Extended Data Table~\ref{tab:multivariate_analysis}). To sum up, these results suggest that BRAVE-derived mutation signals not only capture biologically meaningful information, but also retain independent predictive value beyond standard clinicopathologic variables and genomic burden, with the small $P$ values supporting that this added contribution remained statistically significant after adjustment.

Overall, this set of analyses addressed the robustness question at three levels. We first asked whether mutation prediction mainly reflected subtype proxies and found that predictive discrimination was largely preserved across subtype-relevant subgroups (Figure~\ref{fig:postop}d). This supports the robustness of the mutation-prediction signal across clinically relevant subgroup contexts. We next asked whether the BRAVE-derived mutation signals aligned with established biological associations. Because TP53-related genomic correlates have been extensively documented, we used TP53 as an illustrative example and found consistent associations with mutation count, tumor mutational burden, aneuploidy score, and fraction genome altered (Figure~\ref{fig:postop}e). Finally, we asked whether the H\&E-derived AI score retained independent predictive value beyond clinicopathologic variables and genomic burden, and found that it remained independently informative in multivariable analysis, with the statistically significant added contribution (Figure~\ref{fig:postop}f).

\subsubsection*{BRAVE-guided prioritization of molecular testing for cost-saving}
Having found that BRAVE-derived mutation signals aligned with established biological associations, we next asked whether these signals could be translated into more efficient genomic-testing strategies in practice.
Genomic testing is not universally affordable or accessible, particularly in resource-limited settings. We therefore asked whether BRAVE-derived scores could help prioritize testing toward higher-risk patients while reducing testing in lower-risk groups (Figure~\ref{fig:postop}a,g,h). Following previous analyses, here we also took TP53 as an example. In internal simulations across multiple testing-rate settings, both BRAVE alone and BRAVE combined with clinical variables outperformed clinical stratification alone for prioritizing genomic testing. To illustrate the practical trade-off between testing capacity and mutation capture, we highlight the 20\% and 30\% testing-rate scenarios as representative resource-constrained settings. At a 20\% genomic testing rate, the integrated  clinical + BRAVE strategy increased positive predictive value from 68.1\% to 78.7\% (+10.6\%) compared to current clinical stratification alone (Extended Data Table~\ref{tab:triage_internal}). At a 30\% genomic testing rate, the same integrated strategy increased positive predictive value from 57.7\% to 71.8\% (+14.1\%), while reducing the number of tests required per mutation found from 1.7 to 1.4. These internal simulations suggest that BRAVE-derived scores not only align with biologically meaningful mutation-associated signals, but may also enable more precise prioritization of post-operative TP53 testing while reducing unnecessary testing in lower-risk groups.

We next examined how BRAVE-alone thresholds corresponding to different intended testing rates in TCGA performed in CPTAC as an external validation. For the threshold corresponding to 20\% testing in TCGA, 20.4\% of CPTAC cases were selected, with positive predictive value of 63.6\%, 1.76-fold enrichment over baseline prevalence, and 1.6 tests required per mutation found. For the threshold corresponding to 30\% testing in TCGA, 56.5\% of CPTAC cases were selected, yielding 92.3\% sensitivity, 59.0\% positive predictive value, 1.63-fold enrichment and 1.7 tests per mutation found (Extended Data Table~\ref{tab:triage_ext}). Although these external metrics were slightly attenuated relative to the internal simulations, the prioritization signal remained clinically meaningful after transfer, still showing meaningful enrichment over baseline prevalence and a favorable trade-off between tests required per mutation found and sensitivity. In a nutshell, these results initially support BRAVE-derived scores as a potential tool for prioritizing post-operative genomic testing in resource-limited settings.
\subsection*{Prospective observational validation confirms clinical utility}
\begin{figure}
    \centering
    \includegraphics[width=1.0\linewidth]{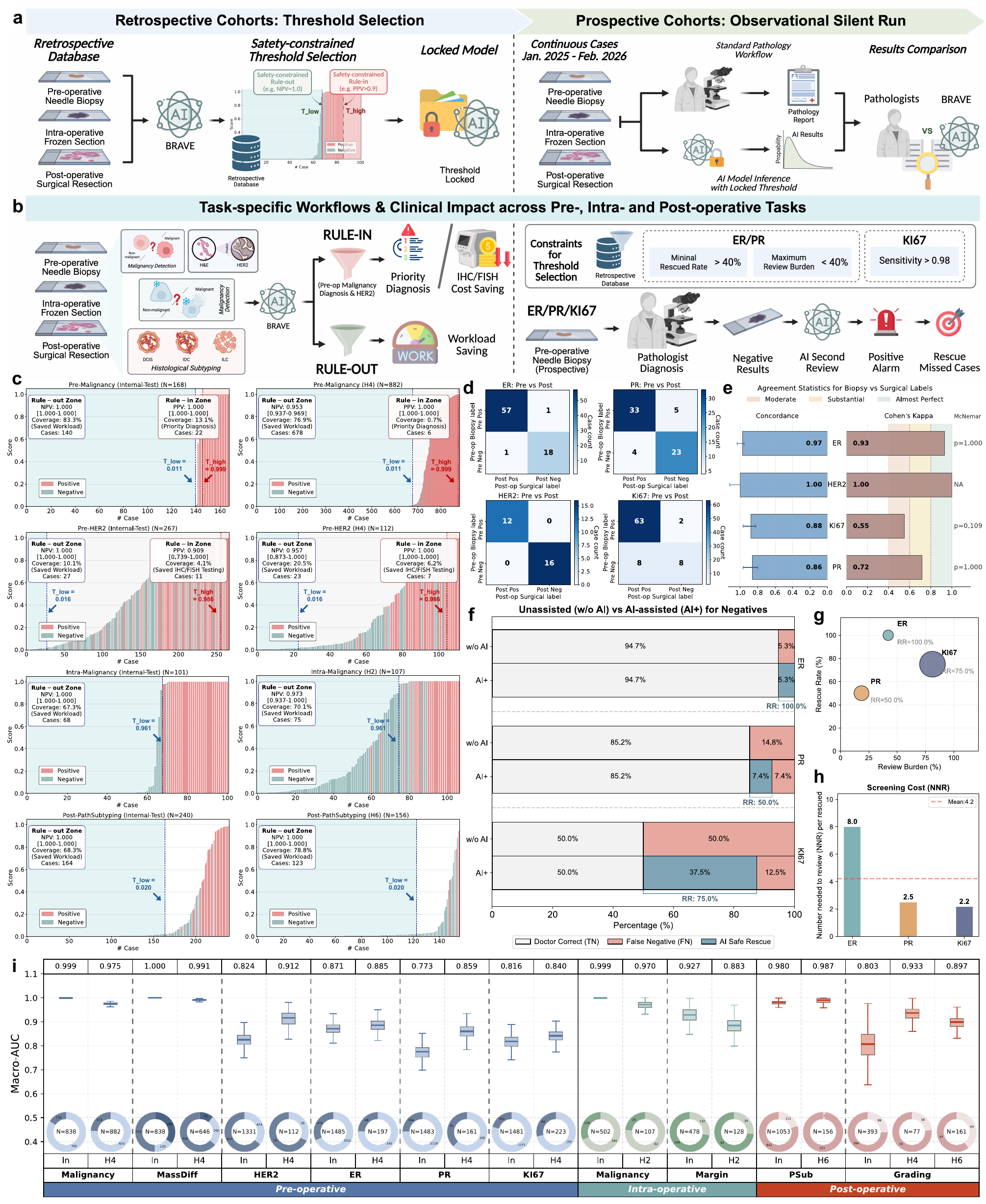}
\end{figure}
\begin{figure}
    \centering
    \caption{\textbf{Prospective observational validation of locked BRAVE workflows across pre-, intra-, and post-operative settings.} \textbf{a}, Study design for prospective observational validation. Task-specific thresholds were selected in retrospective internal cohorts under predefined safety or utility constraints, then locked before silent-run evaluation on consecutive prospective cases. BRAVE predictions were generated without altering the routine pathology workflow and were compared with the corresponding clinical or surgical reference standards. \textbf{b}, Overview of the workflow strategies examined across tasks from pre-operative, intra-operative, and post-operative settings. Left, representative rule-in and rule-out workflows carried forward from retrospective analyses for prospective validation. Right, pre-operative AI-assisted second-review workflows for biomarker assessment, together with representative operating constraints used during retrospective threshold selection. \textbf{c}, Prospective validation results for the workflow strategies shown in \textbf{b} (left), with the corresponding internal-test and prospective score distributions, locked threshold-defined rule-out or rule-in zones, coverage, and negative or positive predictive values shown for representative pre-, intra-, and post-operative tasks. \textbf{d}--\textbf{h}, Prospective validation of AI-assisted second-review workflows for pre-operative biomarker assessment. \textbf{d}, Confusion matrices comparing prospective biopsy-based biomarker labels assigned in routine practice with post-operative surgical labels for ER, PR, HER2, and Ki67. \textbf{e}, Agreement statistics for biopsy versus surgical biomarker labels in the prospective cohort, including concordance, Cohen's kappa, and McNemar testing. \textbf{f}, Comparison of unaided and AI-assisted second review for cases initially assessed as negative, showing doctor-correct negatives, false negatives, and AI-safe rescues for ER, PR, and Ki67 using locked thresholds. \textbf{g}, Efficiency and review burden of prospective AI-assisted second-review workflows across biomarkers, with bubble size indicating the number of rescued missed positive cases. \textbf{h}, Screening cost, expressed as the number needed to review (NNR) per rescued missed positive case. \textbf{i}, Reference performance across representative pre-, intra-, and post-operative tasks in internal and prospective cohorts. $N$ indicates the sample size (case).}
    \label{fig:pros}
\end{figure}

The next step was to test whether workflow strategies selected retrospectively would remain actionable when transferred to a prospective observational setting (Figure~\ref{fig:pros}a,b). In a registered prospective study across three centers, task-specific thresholds were fixed in advance in internal cohorts based on the previous retrospective analyses and then applied without retuning to consecutive prospective cases. When ground-truth labels became available at the end of the prospective study, predictive performance was additionally evaluated for reference against the retrospective cohorts.

To ensure the validation reflected diverse clinical demands, we selected tasks that could benefit most from AI assistance at each clinical stage (Figure~\ref{fig:pros}b) for prospective validation. In the pre-operative biopsy setting, where timely patient stratification and treatment planning are critical to avoid treatment delays, we selected malignancy detection and HER2 assessment for a dual-threshold triage strategy: a low threshold to safely rule out high-confidence low-risk cases and reduce review workload, and a high threshold to rule in high-risk cases for priority assessment so as to avoid delays in the treatment window. Particularly for HER2, confidently prioritized positive cases could also help reduce the need for additional confirmatory testing such as FISH, thereby lowering downstream molecular-testing costs. We additionally evaluated second-review workflows for typical biomarkers, aiming to reduce the risk of missing false-negative cases that could otherwise delay appropriate treatment decisions. For the intra-operative stage, where strict time constraints limit extensive human review, we prioritized malignancy assessment to test the feasibility of safety-constrained rule-out triage to accelerate the surgical decision-making. Finally, for the post-operative stage, we selected high-volume diagnostic tasks such as pathologic subtyping to verify whether AI could reliably triage clear-cut and low-risk cases from the heavy workload of routine surgical pathology.

\subsubsection*{Prospective validation of locked triage workflows}

We first established task-specific operating thresholds using internal retrospective cohorts, enforcing strict constraints to maximize safety for rule-out and clinical certainty for rule-in. Specifically, rule-out thresholds were selected under an NPV constraint of 1.0, whereas rule-in thresholds were selected under a PPV constraint of 1.0 for pre-operative malignancy detection and 0.90 for pre-operative HER2 prediction, as detailed in Methods (\hyperlink{methods_pros_triage}{Prospective validation of locked triage workflows}). When these locked thresholds were applied directly to consecutive prospective cases, they maintained their triage utility (Figure~\ref{fig:pros}c and Extended Data Table~\ref{tab:pros_triage_threshold_perform}). In pre-operative malignancy detection, the low threshold safely excluded 76.9\% negatives of prospective H4 cases from routine review while preserving a high negative predictive value (NPV) of 0.953 (95\% CI, 0.937-0.969). At the other extreme, owing to the strict zero-false-positive constraint used during internal threshold selection, the high threshold prioritized a small but definitive subset of cases for immediate assessment with a perfect positive predictive value (PPV) of 1.000. This small rule-in coverage was intentional, because the rule-in zone was restricted to cases for which the AI was highly confident so as to avoid congesting clinical resources and creating unnecessary additional review workload. Similarly, this patient stratification extended to pre-operative HER2 prediction, where locked thresholds excluded 20.5\% of negative cases (NPV 0.957) and prioritized 6.3\% (PPV 1.000). The approach also generalized robustly to intra- and post-operative rule-out tasks for workload reduction: the AI safely excluded 70.1\% of intra-operative malignancy assessments (NPV 0.973) and 78.8\% of post-operative subtyping cases (NPV 1.000). These results indicate that retrospectively selected thresholds could still support operationally meaningful triage zones after transfer to consecutive prospective cases.

\subsubsection*{Prospective validation of AI-assisted biomarker second review}
We next asked whether the AI-assisted biomarker second-review strategy identified retrospectively would remain clinically usable in the prospective cohort. As described in our retrospective pre-operative analysis of AI-assisted second review (Figure~\ref{fig:preop}c,i--k and Extended Data Figure~\ref{fig:pr_her2_prevspost_analysis}), thresholds were selected by balancing the rescue rate of missed positive cases against the additional review burden. For this prospective validation of AI-assisted biomarker second review, we locked these operating thresholds from those previous retrospective second-review simulations in advance to test their clinical utility. For Ki67, given prior reports of substantial discordance between core needle biopsy and surgical specimens in breast cancer\cite{kalvala2022ki67concordance}, we instead selected the threshold retrospectively in the internal cohort before the prospective study commenced under a conservative sensitivity constraint (sensitivity $\geq 0.98$) to minimize the risk of missing truly positive cases. Detailed threshold selection policies and the resulting thresholds are described in the Methods section (\hyperref[sec:methods_pros_second_review]{Prospective validation of AI-assisted biomarker second review}).

To determine whether it was meaningful to proceed with further analysis of AI-assisted second review in this prospective setting, we first examined the degree of discordance between biopsy-stage biomarker assessments and the definitive surgical result in prospective cohorts. We therefore compared routine biopsy-based pathologist labels with post-operative surgical labels in the prospective cohort (Figure~\ref{fig:pros}d--e and Extended Data Tables~\ref{tab:pros_biomarker_paired_distribution} and~\ref{tab:pros_biomarker_concordance_stats}). Agreement was high for ER and HER2, with concordance of 97.4\% and 100.0\% and Cohen's kappa values of 0.930 and 1.000, respectively. Agreement was lower for PR and Ki67, with concordance of 86.2\% and 87.7\% and kappa values of 0.716 and 0.546. The discordance for PR and Ki67 confirmed that a meaningful proportion of biopsy-stage calls deviated from the surgical reference. Notably, no biopsy--surgery discordance was observed for HER2 in this prospective cohort, likely reflecting the relatively small number of biopsy cases in the current prospective series that had proceeded to surgery and become available for paired analysis. HER2 was therefore excluded from the subsequent second-review analysis.

We next asked whether this discordance mainly reflected a systematic tendency toward over-calling or under-calling. If the errors were systematically skewed, it would indicate a need for calibration rather than case-by-case second reviews using AI. However, McNemar testing showed no significant directional bias for either biomarker (ER: $p = 1.000$; PR: $p = 1.000$; Ki67: $p = 0.109$), suggesting that the discordance was not driven by systematic bias. Instead, it reflected sporadic, case-level errors, which are well suited for an AI-assisted supplemental review. Given that BRAVE was trained using accurate post-operative surgical labels as the objective ground truth rather than potentially discordant biopsy evaluations, we reasoned it would be well equipped to uncover the true biomarker status. We therefore evaluated whether an AI-assisted second review could successfully identify these missed cases.

Using the locked thresholds, AI-assisted second review can rescue missed biomarker-positive cases among specimens initially assessed as negative by pathologists (Figure~\ref{fig:pros}f--h). For ER, the selected threshold rescued all biopsy-negative but surgery-positive cases (100\% rescue rate). Notably, 8 second reviews per rescued missed case (number needed to review [NNR] of 8.0) suggests that the retrospectively simulated threshold may have been overly sensitive. For PR, AI-assisted second review rescued 50.0\% missed positive cases at the NNR of 2.5. For Ki67, the selected threshold rescued 75.0\% missed positive cases at the NNR of 2.2. These NNRs indicate that a modest number of additional reviews per rescued case was sufficient to identify a substantial proportion of missed positives. Although the review burden and rescue rate varied across biomarkers, the results together show that the locked-threshold second-review strategy remained capable of rescuing a meaningful fraction of biopsy-stage misses in prospective use, especially for Ki67 with higher discordance in routine practice.

\subsubsection*{Reference performance after prospective ground-truth collection}

After the prospective study period and subsequent ground-truth collection, we also summarized performance of every task for reference rather than for threshold selection (Figure~\ref{fig:pros}i and Extended Data Table~\ref{tab:pros_auc}). Macro-AUC remained high for representative prospective tasks, including 0.975 (95\% CI, 0.965-0.984) for pre-operative malignancy detection in H4, 0.912 (0.847-0.966) for pre-operative HER2 prediction in H4, 0.970 (0.933-0.994) for intra-operative malignancy detection in H2, 0.987 (0.955-1.000) for post-operative pathologic subtyping in H6, and 0.897 (0.843-0.944) and 0.933 (0.867-0.982) for post-operative grading in H6 and H4, respectively.

Since these cases were collected consecutively after threshold locking and their definitive labels became available only after follow-up, these reference metrics serve as a post-study confirmation of the model's overall discriminative ability, rather than being used to tune the model or establish triage thresholds. Nevertheless, they broadly support the impression that BRAVE retained informative signals in prospective cohorts across representative tasks from the three clinical stages.
\subsection*{Pathologist-AI collaboration improves diagnostic performance}
\begin{figure}
    \centering
    \includegraphics[width=0.95\linewidth]{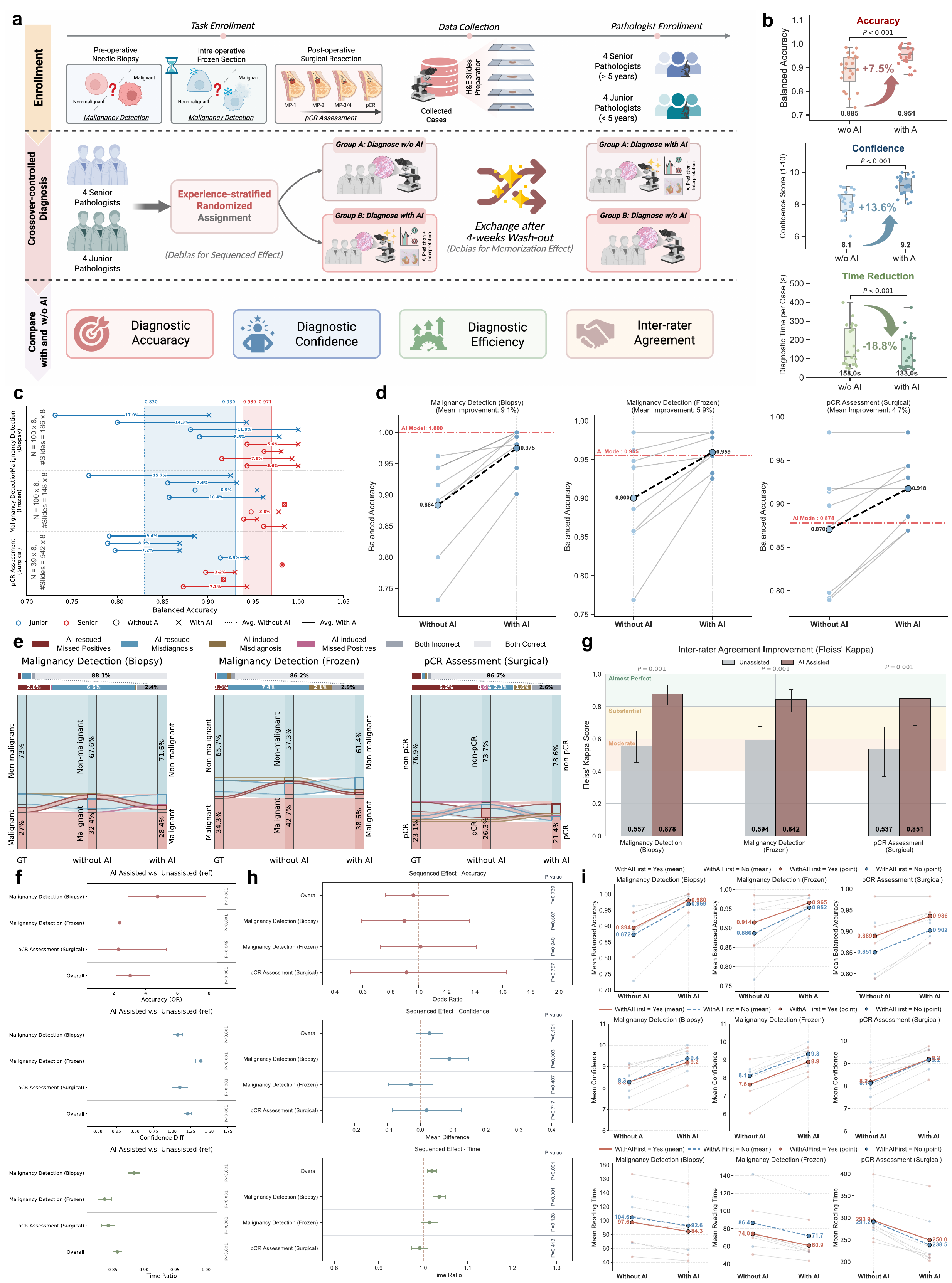}
\end{figure}
\begin{figure}
    \centering
    \caption{\textbf{Real-world Pathologist-AI Reader Study}. \textbf{a}, Cross-over reader study design. \textbf{b}, Overall comparison of balanced accuracy, confidence, and reading time between unassisted and AI-assisted performance from 8 pathologists across three diagnostic tasks spanning pre-operative, intra-operative, and post-operative settings. \textbf{c}, Balanced accuracy stratified by reader experience across three tasks, comparing 4 junior ($<$5 years experience) and 4 senior ($>$5 years experience) pathologists with and without AI assistance. Circles indicate unassisted performance and crosses indicate AI-assisted performance for individual pathologists. Dashed and solid lines represent the group mean for junior and senior pathologists, respectively. \textbf{d}, Task-level comparison of balanced accuracy between unassisted and AI-assisted reading. Each point represents an individual pathologist, and grey lines connect the same pathologist's performance across the two conditions. The black dashed lines indicate the average performance within each group. \textbf{e}, Changes in diagnostic classifications between unassisted and AI-assisted reading across the three tasks. \textbf{f}, Generalized Estimating Equation (GEE) analyses of diagnostic accuracy (odds ratio, OR), reading time (time ratio, TR), and confidence (confidence difference, Diff). Models used an exchangeable correlation structure, adjusted for period, task, and experience level, and treated pathologist as the clustering unit. The without-AI condition served as the reference. OR $>$ 1 indicates higher odds of correct diagnosis with AI assistance, TR $<$ 1 indicates reduced reading time with AI, and Diff $>$ 0 indicates increased confidence. $P$ values are from GEE. \textbf{g}, Comparison of inter-rater agreement (Fleiss's $\kappa$), where $\kappa<0.2 = $poor; $0.21-0.40 = $fair; $0.41-0.60 = $moderate; $0.61-0.80 = $substantial; $\kappa>0.8 = $almost perfect. A non-parametric bootstrap approach (1,000 replicates) was used to test the significance of the difference between the two Fleiss' $\kappa$ coefficients. \textbf{h-i}, Sequence-effect analyses of the crossover reader study. \textbf{h}, GEE-based estimates of sequence effects on diagnostic accuracy, confidence, and reading time, with corresponding $P$ values shown for the overall cohort and for each task. \textbf{i}, Descriptive performance statistics stratified by AI-intervention sequence (WithAIFirst = Yes or No), shown separately for accuracy, confidence, and reading time across the three tasks. Small points represent individual pathologists, and large points with connecting lines represent group means.}
    \label{fig:reader_study}
\end{figure}

\subsubsection*{Crossover reader study across clinical stages}

To determine whether AI assistance could directly improve pathologist accuracy, efficiency, confidence, and inter-rater agreement under routine diagnostic conditions, we conducted a crossover multi-reader study spanning pre-, intra-, and post-operative diagnostic tasks. In the unassisted condition, pathologists made decisions based only on the provided H\&E slides, whereas in the AI-assisted condition they reviewed the same H\&E slides together with BRAVE's predicted result (probability score) and confidence heatmap.

We enrolled 8 pathologists, including 4 senior ($>$5 years experience) and 4 junior ($<$5 years) readers, to interpret three representative tasks, with cases selected by class-balanced random sampling from the corresponding collected test sets to ensure a balanced case composition for reader evaluation. As a result, malignancy detection on biopsy ($N=100$ cases), malignancy detection on frozen section ($N=100$ cases), and post-operative pCR assessment ($N=39$ cases) (Figure~\ref{fig:reader_study}a) were established. A randomized crossover design with a 4-week wash-out period was used so that each reader completed the study both with and without AI assistance, yielding 1,600 individual reader interpretations for biopsy, 1,600 for frozen section, and 624 for pCR assessment (cases $\times$ readers $\times$ conditions), for a total of 3,824 individual reader interpretations across the study. Notably, for the intra-operative frozen-section task, the reading interface enforced a 10-minute time limit per case to mimic the time-critical nature of intra-operative consultation. The predefined handling rules for any timed-out reads are described in \hyperlink{methods_reader_tasks}{Methods - Reading tasks and reference standard}. Although these timeout-handling rules were prespecified, no timed-out cases occurred in this reader study.

Overall, results (Figure~\ref{fig:reader_study}b) have shown that AI assistance was associated with higher diagnostic performance, greater confidence, and shorter reading time (Extended Data Tables~\ref{tab:all_reader_results_biopsy}--\ref{tab:all_reader_results_surg}). Mean balanced accuracy increased from 0.885 to 0.951 (+7.5\%, $P<0.001$), mean confidence increased from 8.1 to 9.2 (+13.6\%, $P<0.001$), and mean reading time decreased from 158.0s to 133.0s (-18.8\%, $P<0.001$).

\subsubsection*{AI assistance improved performance across experience levels and tasks}

Experience-stratified analysis showed improvements in both reader groups, with larger gains (+10\%) among junior pathologists (mean balanced accuracy 0.830 to 0.930) than among senior pathologists (0.940 to 0.971) (Figure~\ref{fig:reader_study}c). In biopsy malignancy detection, mean balanced accuracy increased from 0.826 to 0.956 among junior readers and from 0.942 to 0.994 among senior readers. In frozen-section diagnosis, the corresponding changes were 0.842 to 0.943 for junior readers and 0.959 to 0.976 for senior readers. In pCR assessment, mean balanced accuracy increased from 0.823 to 0.892 among junior readers and from 0.918 to 0.943 among senior readers. These results indicate that AI assistance benefited both groups while narrowing the performance gap between less- and more experienced readers.

Across various tasks (Figure~\ref{fig:reader_study}d), balanced accuracy improved from 0.884 to 0.975 for biopsy malignancy detection (+9.1\%, $P<0.001$), from 0.900 to 0.959 for frozen-section malignancy detection (+5.9\%, $P<0.001$), and from 0.870 to 0.918 for pCR assessment on surgical resection (+4.7\%, $P=0.048$) (Extended Data Tables~\ref{tab:all_reader_results_biopsy}--\ref{tab:all_reader_results_surg}). AI performance is shown for reference (1.000 for biopsy, 0.955 for frozen section, and 0.878 for pCR). Although AI performance was more modest in pCR assessment, AI assistance still improved reader performance, suggesting that AI support could provide complementary information in this more difficult post-treatment setting. Across the three tasks, larger reader gains were observed in settings where AI also performed better.

Decision-trajectory analysis further clarified the clinical meaning of these gains (Figure~\ref{fig:reader_study}e). In biopsy malignancy detection, AI assistance corrected 9.2\% of the initial 11.6\% error burden observed without AI, including 2.6\% of all reads that had missed malignancy and 6.6\% of misclassifications. Clinically, this results in fewer missed malignancies, which helps minimize diagnostic delays, and fewer false-positive diagnoses, thereby avoiding more unnecessary clinical interventions for benign cases. 

In frozen-section diagnosis, AI assistance corrected 8.7\% out of the initial 11.6\% error burden, including 1.3\% of all reads that had missed malignancy and 7.4\% of misclassifications, although 2.1\% of reads were newly misclassified with AI assistance. Despite this trade-off, AI assistance produced a net reduction in diagnostic error, moving frozen-section interpretations closer to the final diagnosis. This is clinically important in the intra-operative setting, where both missed malignancy and unnecessary additional resection can directly affect surgical decision making and patient outcomes.

In post-operative pCR assessment, AI assistance corrected 8.5\% out of the initial 11.1\% error burden, including 6.2\% of all reads that had incorrectly classified non-pCR cases as pCR and 2.3\% of reads in the opposite direction, while introducing smaller countervailing errors (0.6\% and 1.6\%). These changes are clinically relevant in both directions: reducing false pCR calls may help avoid premature treatment de-escalation in patients with residual disease, whereas rescuing missed pCR calls may help avoid unnecessary overtreatment. Overall, the trajectory plots show that AI assistance substantially reduced the initial error burden across all three tasks, while residual errors remained, highlighting its value as an assistive tool that still requires pathologist oversight.

\subsubsection*{Statistical analyses supported robust gains in reader performance}

We next used Generalized Estimating Equations (GEE) to account for repeated measurements within readers while adjusting for period, task, and experience level~\cite{hardin2002generalized}. Since each pathologist evaluated the same set of cases under both conditions, their readings are intrinsically correlated. This statistical approach specifically models this intra-reader correlation to prevent individual biases from skewing the results, while simultaneously controlling for potential confounders such as reader experience, specific diagnostic task, and study period (batch effect). Consequently, this isolates the true independent benefit of AI assistance.

Results (Figure~\ref{fig:reader_study}f and Extended Data Table~\ref{tab:gee}) showed that AI assistance remained independently associated with higher diagnostic accuracy, shorter reading time, and greater confidence, after adjustment for period, task, and experience level. This indicates that the AI tool provides a robust and independent benefit to pathologist performance, confirming that the improvements were truly driven by the AI itself, regardless of the pathologist's experience level or the inherent difficulty of the task. The specific impact of the crossover reading sequence is further explained below. For diagnostic accuracy, the overall odds ratio (OR) was 3.14 (95\% CI, 2.215--4.461; $P<0.001$), with task-specific ORs of 4.88 for biopsy malignancy detection, 2.50 for frozen-section diagnosis, and 2.41 for pCR assessment ($P=0.048$). This means that pathologists were more than three times as likely to make a correct diagnosis when using AI assistance compared to reading without it, and this substantial improvement was consistent across all three diagnostic tasks. AI assistance also reduced reading time consistently, with time ratios of 0.855 for biopsy, 0.807 for frozen section, 0.820 for pCR, and 0.829 overall (all $P<0.001$). This indicates that pathologists completed their evaluations approximately 17\% faster across the board when supported by the AI. Confidence increased in parallel, with overall confidence difference of 1.10 (95\% CI, 1.058--1.147; $P<0.001$), and task-specific differences of 1.01 for biopsy, 1.23 for frozen section, and 1.02 for pCR. In other words, AI assistance boosted the pathologists' self-reported certainty by about a full point on a 10-point scale.

AI assistance also improved agreement among readers (Figure~\ref{fig:reader_study}g and Extended Data Table~\ref{tab:kappa}). Without AI, inter-rater agreement was in the moderate range for all three tasks (Fleiss's $\kappa$ 0.556 for biopsy, 0.592 for frozen section, and 0.534 for pCR). With AI assistance, agreement increased to 0.876 for biopsy, 0.841 for frozen section, and 0.841 for pCR (all $P\leq 0.001$ for the difference). This represents a leap from ``moderate'' to ``almost perfect'' agreement, indicating that AI support reduced subjective discrepancies and standardized interpretations across different pathologists.

Finally, sequence-effect analyses supported the robustness of the crossover design (Figure~\ref{fig:reader_study}h,i and Extended Data Tables~\ref{tab:seq_effect_stat} and~\ref{tab:seq_effect_desc}). No significant sequence effect was observed for diagnostic accuracy either overall or within any individual task (all $P\geq 0.607$). Figure~\ref{fig:reader_study}i likewise showed no statistically significant overall difference ($P=0.739$) between the two intervention sequences, despite modest descriptive differences in absolute performance. For secondary endpoints, sequence-related effects were observed for reading time overall and for confidence in biopsy assessment, whereas the remaining confidence and reading-time comparisons were not significant. Importantly, these sequence-related effects were not observed for diagnostic accuracy, suggesting that the observed accuracy gains from AI assistance were not materially influenced by memory effects. A longer wash-out interval may nevertheless help reduce possible carryover effects in secondary endpoints in future crossover studies.

Overall, these findings indicate that BRAVE functioned as an effective assistive tool for improving diagnostic performance across pre-operative, intra-operative, and post-operative settings. AI assistance improved accuracy, shortened reading time, increased confidence, and strengthened inter-rater agreement, with particularly large gains among junior readers and in tasks where model performance was also stronger. Decision-trajectory analyses further showed that these improvements translated into correction of a substantial proportion of the initial diagnostic error burden across all three tasks, while GEE analyses confirmed that the benefits remained after adjustment for period, task, and reader experience. The absence of a significant sequence effect on diagnostic accuracy further supports the robustness of these findings. Taken together, these results support the view that BRAVE can serve as a practical assistive tool for pathologists and may help mitigate pathology workforce constraints in resource-limited settings.
\subsection*{BRAVE-derived scores support prognostic risk stratification}
\begin{figure}
    \centering
    \includegraphics[width=1.0\linewidth]{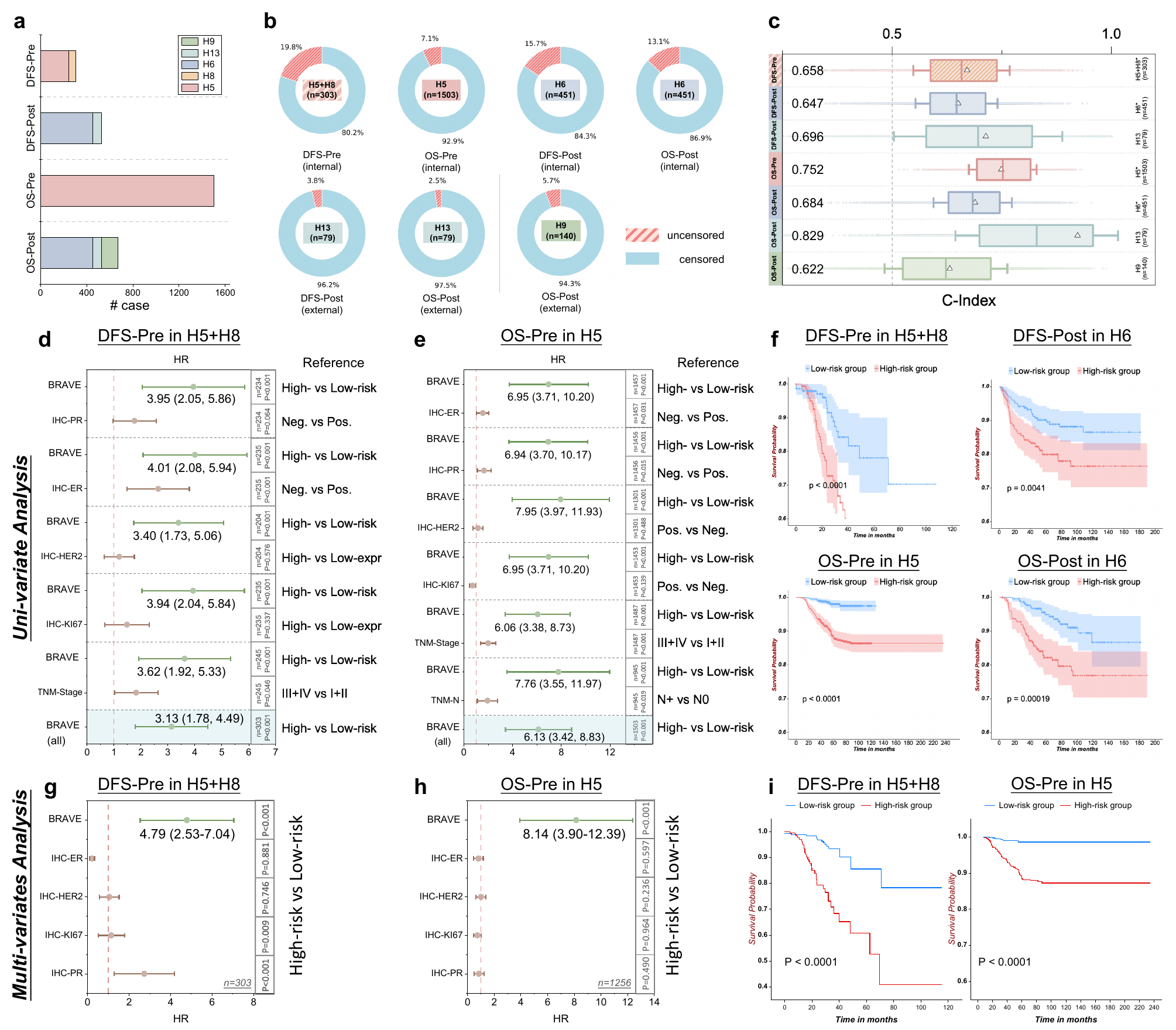}
    \caption{\textbf{Performance of BRAVE for survival prediction}. \textbf{a}, Distribution of patients across 5 centers. \textbf{b}, Distribution of censoring across 7 survival prediction cohorts. \textbf{c}, C-index distributions from 5,000 bootstrap resamples across the 7 cohorts. Boxes indicate the first and third quartiles, the horizontal line indicates the median, the triangle indicates the mean, and whiskers indicate the standard deviation. * denotes the internal cohort used to train the MIL model. \textbf{d-f}, Univariable analyses. Hazard ratios (HRs) for BRAVE and clinicopathological variables in (\textbf{d}) DFS-Pre in the H5+H8 cohort and (\textbf{e}) OS-Pre in the H5 cohort, both based on pre-operative data. The final row highlighted in blue shows the HR for BRAVE calculated using all patients in the corresponding cohort. Error bars indicate 95\% confidence intervals (CIs), and \textit{P} values were estimated from 1,000 bootstrap resamples. Pos. and Neg. indicate positive and negative, respectively. (\textbf{f}) Kaplan-Meier curves for the internal cohorts (five folds), with \textit{P} values from the log-rank test. \textbf{g-h}, Multivariable analyses using Cox proportional hazards models~\cite{harrell2001cox}. HRs for all variables in (\textbf{g}) DFS-Pre in the H5+H8 cohort and (\textbf{h}) OS-Pre in the H5 cohort, both based on pre-operative data. Error bars indicate 95\% CIs, and \textit{P} values were estimated from 1,000 bootstrap resamples. Statistical significance for HRs was defined as a 95\% CI excluding 1. \textbf{i}, Survival curves adjusted for ER, PR, HER2, and Ki67, estimated using the Breslow estimator.}
    \label{fig:surv}
\end{figure}
\subsubsection*{Prognostic stratification performance across pre-operative and post-operative cohorts}

Here we extended the model's capability from diagnosis to prognosis, evaluating whether AI could extract sub-visual features to predict patient survival for stratification. We curated a comprehensive survival analysis cohort spanning multiple centers (H5, H6, H8, H9, H13) covering both Disease-Free Survival (DFS) and Overall Survival (OS) in pre-operative (Pre) and post-operative (Post) settings (Figure~\ref{fig:surv}a and Extended Data Table~\ref{tab:task_dist_surv_data}).

First, BRAVE showed consistent prognostic stratification performance across 7 survival prediction cohorts (Figure~\ref{fig:surv}c and Extended Data Table~\ref{tab:surv_results}). We assessed the Concordance Index (C-index) using 5,000 bootstrap iterations to estimate statistical uncertainty. In the internal cohorts, the model achieved a C-index of 0.752 for pre-operative OS prediction in H5 and 0.658 for pre-operative DFS prediction in the combined H5+H8 cohort. In post-operative evaluation, BRAVE also retained predictive value across external cohorts, with C-index values of 0.696 for DFS-Post in H13, 0.829 for OS-Post in H13, and 0.622 for OS-Post in H9, supporting the transferability of the learned morphological features across centers.

\subsubsection*{BRAVE showed stronger and independent prognostic value than routine clinicopathological  variables}

We next assessed whether BRAVE provided stronger and independent prognostic value compared with routine clinicopathological variables. Kaplan-Meier analysis of the internal cohorts (Figure~\ref{fig:surv}f) showed significant separation between the high-risk and low-risk groups across all four settings, with log-rank $P<0.0001$ for DFS-Pre and OS-Pre, $P=0.0041$ for DFS-Post, and $P=0.00019$ for OS-Post (all $P<0.05$). To compare the stratification strength of BRAVE directly against established clinicopathological variables on the same effect-size scale, we next performed univariable Cox regression analysis (Figure~\ref{fig:surv}d,e and Extended Data Table~\ref{tab:surv_uni_hr_dfs},~\ref{tab:surv_uni_hr_os}), verified by 1,000 bootstrap iterations for 95\% CI. Figures~\ref{fig:surv}d--e have shown that the BRAVE risk score emerged as one of the strongest predictors. For pre-operative OS (Figure~\ref{fig:surv}e and Extended Data Table~\ref{tab:surv_uni_hr_os}), the hazard ratio (HR) for high-risk versus low-risk patients was 6.13 (95\% CI, 3.42--8.83; $P<0.001$), exceeding those of TNM stage and TNM-N. This means that patients classified as high risk by BRAVE had more than a sixfold higher risk of death than those classified as low risk, and that this separation was markedly stronger than that provided by conventional staging variables alone. Similarly, in DFS-Pre analysis, BRAVE maintained strong predictive value with an HR of 3.13 (95\% CI, 1.78--4.49; $P<0.001$) (Figure~\ref{fig:surv}d and Extended Data Table~\ref{tab:surv_uni_hr_dfs}). These univariable results indicate that BRAVE provided stronger prognostic stratification than several routinely used clinicopathological variables.

Since IHC markers (ER, PR, HER2, and Ki67) are well-established prognostic factors in breast cancer, a key question is whether BRAVE's prognostic signal reflects these markers as proxies or captures independent prognostic information beyond them. To address this, we performed multivariable Cox analysis adjusting for these four IHC variables. In DFS-Pre, the adjusted HR for BRAVE was 4.79 (95\% CI, 2.53--7.04; $P<0.001$) (Figure~\ref{fig:surv}g), and in OS-Pre the adjusted HR was 8.14 (95\% CI, 3.90--12.39; $P<0.001$) (Figure~\ref{fig:surv}h). In the OS-Pre cohort, the conventional IHC markers were no longer statistically significant after adjustment ($P>0.05$), whereas BRAVE remained significant, suggesting that BRAVE captures prognostic information not fully explained by these conventional IHC biomarkers. This independent prognostic value was also reflected by the adjusted survival curves in Figure~\ref{fig:surv}i, which showed clear separation between the BRAVE-defined risk groups after accounting for ER, PR, HER2, and Ki67. These multivariable results indicate that BRAVE captured prognostic information beyond routine clinicopathological IHC biomarkers rather than merely recapitulating their signal.

Overall, these findings indicate that BRAVE captures additional prognostic information from routine H\&E slides beyond standard clinicopathological variables. Its risk score consistently stratified survival, remained significant in multivariable models, and generalized across both pre-operative and post-operative cohorts. These results support the potential utility of BRAVE for pathology-based survival risk stratification across different clinical settings.
\subsection*{Breast-adaptive pretraining improves performance over pan-cancer foundation models}
\begin{figure}[!t]
    \centering
    \includegraphics[width=0.95\linewidth]{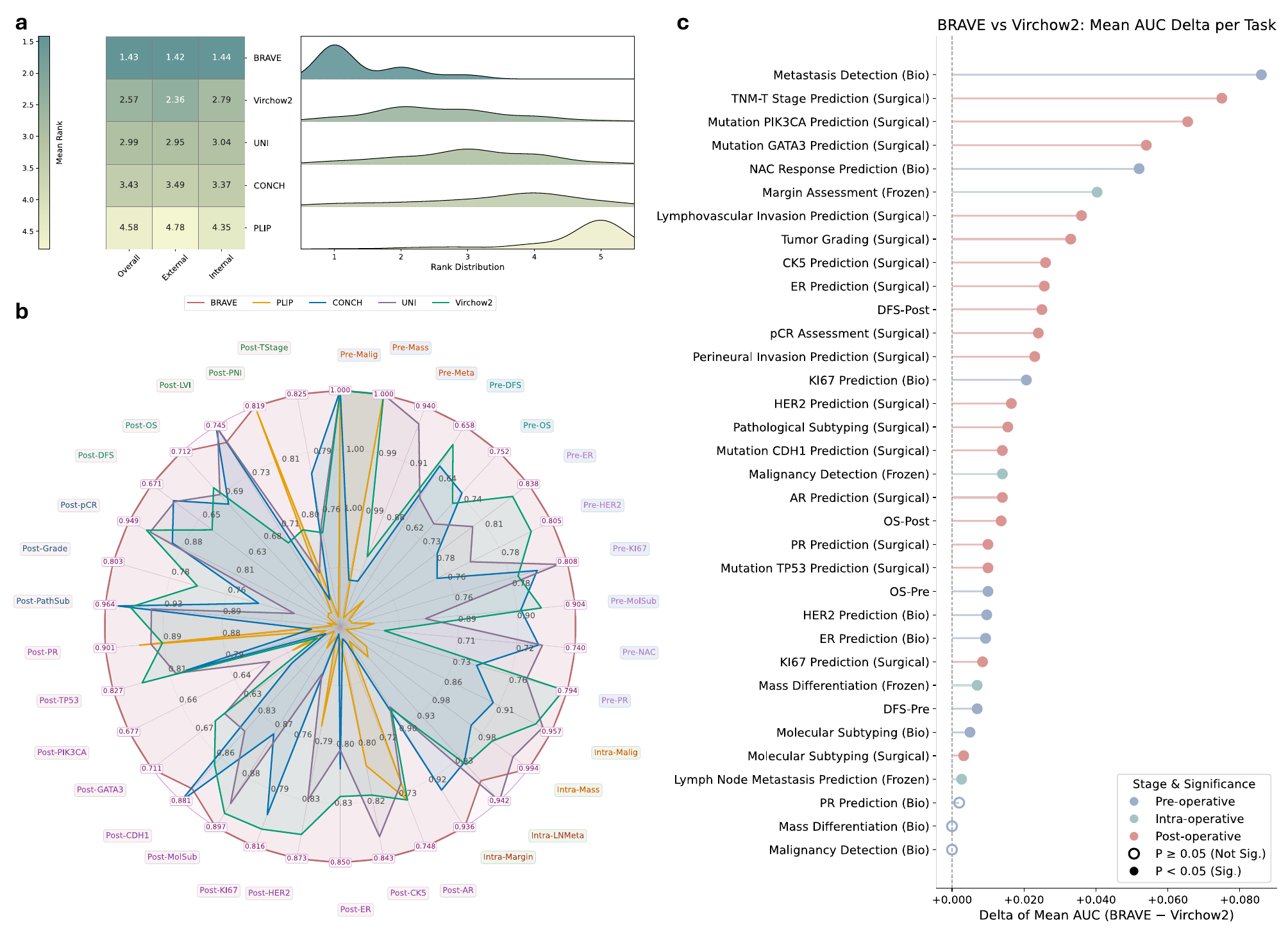}
    \caption{\textbf{Ablation study.} \textbf{a}, Mean rank (lower is better) and rank distributions of 5 pathology foundation models across all 75 retrospective classification cohorts, shown overall as well as on external and internal cohorts. \textbf{b}, Task-level comparison of mean Macro-AUC across all 30 classification tasks for the 5 pathology foundation models. For tasks evaluated in multiple centers, Macro-AUC values were averaged across centers before plotting. \textbf{c}, Task-level improvement in mean Macro-AUC of BRAVE relative to the baseline model Virchow2 across all 30 classification tasks spanning pre-, intra-, and post-operative stages. For tasks evaluated in multiple centers, Macro-AUC values were averaged across centers before computing the task-level difference. Filled and open circles indicate $P\leq 0.05$ and $P>0.05$, respectively, from one-sided Wilcoxon signed-rank tests.}
    \label{fig:ablation}
\end{figure}
\subsubsection*{Breast-adaptive pretraining improved performance}

We benchmarked BRAVE against four state-of-the-art pan-cancer pathology foundation models, including Virchow2~\cite{vorontsov2024foundation}, UNI~\cite{chen2024towards}, CONCH~\cite{lu2024visual}, and PLIP~\cite{huang2023visual}, using all collected retrospective classification datasets spanning 75 cohorts from 30 classification tasks across the pre-, intra-, and post-operative workflow. Because several tasks were evaluated in multiple centers, we summarized performance at both the cohort level (75 cohorts; Figure~\ref{fig:ablation}a) and the task level (30 tasks; Figure~\ref{fig:ablation}b,c).

Across the 75 retrospective cohorts, BRAVE ranked first overall, with the lowest mean rank of 1.43, compared with 2.57 for Virchow2, 2.99 for UNI, 3.43 for CONCH, and 4.58 for PLIP (Figure~\ref{fig:ablation}a). This advantage was consistent across data splits, with mean ranks of 1.42 on external cohorts and 1.44 on internal cohorts, whereas the corresponding mean ranks for Virchow2 were 2.36 and 2.79. These results indicate that breast-adaptive pretraining improved performance broadly rather than only in a particular validation subset.

We then focused on task-level comparisons with Virchow2, the strongest pancancer baseline and the initialization model for BRAVE. Using the mean Macro-AUC averaged across centers when applicable, BRAVE outperformed Virchow2 on almost all 30 classification tasks, with 27 showing statistically significant superiority (Figure~\ref{fig:ablation}b,c).

\subsubsection*{Performance gains were most pronounced in clinically difficult tasks}

The largest improvements over Virchow2 were observed in clinically demanding tasks that require fine-grained breast-specific morphological discrimination. These included metastasis detection in biopsy specimens, TNM-T stage prediction in surgical specimens, PIK3CA and GATA3 mutation prediction, NAC response prediction in biopsy specimens, and frozen-section margin assessment (Figure~\ref{fig:ablation}c). BRAVE also showed positive mean Macro-AUC gains across many post-operative biomarker and invasion-related tasks, including lymphovascular invasion, perineural invasion, tumor grading, and several IHC prediction tasks.

By contrast, only a small number of tasks showed near-zero or non-significant differences relative to Virchow2, indicating that the gain from breast-adaptive pretraining was more concentrated in tasks with greater breast-specific morphological complexity. Overall, these ablation results support the value of domain-specific adaptation for improving retrospective performance across a broad spectrum of clinically relevant breast pathology tasks.
\section*{Discussion}
\textbf{From retrospective benchmarking to workflow-aligned validation}. The main contribution of this study is not only that BRAVE achieved strong retrospective performance, but that a breast-adaptive pathology foundation model was evaluated through a workflow-aligned evidence chain spanning retrospective benchmarking, robustness validation in challenging scenarios, scenario-specific clinical simulations, locked-threshold prospective validation, crossover pathologist-AI interaction, and prognostic analysis. This distinction is crucial because the translational bottleneck in computational pathology is no longer limited to showing discrimination in curated retrospective cohorts. Instead, models must be assessed against the specific decision points at which they may be used in practice. By organizing evaluation across pre-operative, intra-operative, and post-operative settings, and by extending validation from retrospective benchmarking to prospective and human-AI evaluation, this study provides a clinically grounded framework for examining how pathology AI may be translated into breast cancer care.

\noindent
\textbf{Stage-specific utility across the breast pathology workflow}. A central message from the results is that clinically useful AI does not take the same form at every stage of breast pathology. In the pre-operative setting, BRAVE supported routine core-biopsy diagnosis, biomarker-related assessment from limited tissue, and AI-assisted second review to rescue initially missed positive cases before treatment planning. In the intra-operative setting, its value was tied more closely to rapid decision support and safety-constrained triage under frozen-section time pressure, where review burden could be reduced while preserving a high rule-out safety profile. In the post-operative setting, the model extended beyond routine morphology into biomarker prioritization and molecular prediction from resection specimens. Validating BRAVE across the breast pathology workflow showed that the value of AI is stage-specific and depends on whether model performance is aligned with the clinical objective, tissue context, and decision consequence at each point of care.

\noindent
\textbf{Prospective and human-AI evidence provide stronger support for clinical translation}. In this study, prospective validation and direct pathologist-AI interaction provided support for clinical translation beyond retrospective AUCs alone. In the prospective setting, BRAVE retained utility under locked thresholds for triage and biomarker second review, which better reflects how the model would operate in real clinical deployment than retrospectively selecting an optimal threshold. In the reader study, AI assistance improved accuracy while also reducing reading time, increasing confidence, and improving inter-rater agreement across pre-, intra-, and post-operative tasks. This pattern supports an assistive role for BRAVE as a standardizing and efficiency-enhancing tool rather than as a replacement for pathologist judgment. In particular, these gains were particularly marked among junior pathologists and in some settings narrowed the gap between less- and more- experienced readers.

\noindent
\textbf{Breast-adaptive pretraining showed added value in more challenging tasks}. The ablation analyses suggest that breast-specific adaptive pretraining provided additional benefit beyond already strong pan-cancer baselines, with the clearer gains emerging in tasks that require more complex and breast-specific morphological information. BRAVE ranked first overall across the retrospective benchmark and showed the larger improvements over Virchow2 in tasks such as metastasis detection, stage prediction, mutation prediction, and frozen-section margin assessment. This pattern suggests that specialization becomes most valuable when the mapping from morphology to clinically relevant labels is subtle, heterogeneous, or highly specific to breast pathology.

\noindent
\textbf{Biological and prognostic independent information beyond standard clinicopathological assessment from routine H\&E}. Beyond diagnostic assistance, the results also suggest that a breast-adaptive model can extract clinically relevant information from routine histology that is not fully captured by standard clinicopathological assessment. Evidence for additional biological information was more apparent in the post-operative analyses, where the H\&E-derived AI score was associated with mutation count, tumor mutational burden, aneuploidy score, and fraction genome altered, and remained independently informative in multivariable analysis beyond clinicopathologic variables and genomic burden. This is a more convincing evidence of additional biological association than retrospective benchmarking alone, because it suggests that the AI-derived score is not merely recapitulating standard morphologic or clinical correlates, but may also capture additional biological variation encoded in routine histology. In the prognostic setting, BRAVE-derived risk scores stratified both disease-free and overall survival and remained significant after adjustment for routine variables. These findings suggest that BRAVE captures additional prognostic independent information from routine H\&E that is not fully reflected by standard clinicopathological variables. However, this should not be interpreted as evidence that morphology alone can replace established ancillary testing. Instead, it suggests that AI can complementarily capture underused information in routine histology for risk stratification.

\noindent
\textbf{Limitations and next steps}. Several limitations should be considered when interpreting these findings. First, although prospective validation in this study was already conducted across multiple centers, the evidence remained observational rather than interventional. Therefore, intervention-level clinical benefit has not yet been established. Second, prospective validation has not yet been extended to a broader range of centers and workflow scenarios. Third, the prognostic analyses remain based on retrospective cohorts, because breast cancer prognosis is generally favorable, and sufficient survival events may not accumulate within short follow-up periods. Meaningful prospective assessment will therefore require broader cohorts and longer-term outcome collection. At the same time, these limitations should be interpreted in context. Compared with many pathology AI studies that remain confined to retrospective discrimination analyses, the present work already assembles a substantially more complete translational evaluation pathway by integrating large-scale retrospective benchmarking, workflow-specific operating-point analysis, locked-threshold prospective validation, pathologist-AI reader study, and prognostic assessment within a single breast-pathology framework. Future studies should therefore build on this foundation by moving from observational prospective validation to independent clinical trial evaluation, alongside broader multi-center reader studies and intervention-level assessments of clinical and health-economic impact, while also exploring multimodal extensions that combine histology with radiology, clinical variables, and molecular data.

In summary, BRAVE supports a practical framework for translating pathology foundation models into breast cancer care. Its value lies not only in strong retrospective performance, but in demonstrating how a breast-adaptive model can be evaluated across diagnostic assistance, workflow support, and prognostic assessment under conditions that move meaningfully closer to real clinical use.
\section*{Methods}
\subsection*{Ethic Declarations}
This study was reviewed and approved by Human and Artefacts Research Ethics Committee (HAREC) of Hong Kong University of Science and Technology (protocol no. HREP-2024-0212 and HREP-2025-0349), the Medical Ethics Committee of Nanfang Hospital of Southern Medical University (approval numbers: NFEC-2025-419 and NFEC-2025-403), and the Clinical Research Ethics Committee of the First Affiliated Hospital, Zhejiang University School of Medicine (approval No. 1034). The study was carried out in compliance with the Declaration of Helsinki. The retrospective study was registered in the U.S. Clinical Trial Registry (NCT07239297). The prospective validation studies were prospectively registered in the U.S. Clinical Trial Registry for pre-operative and post-operative cohorts (NCT07157618), and the intra-operative cohorts were prospectively registered in the Chinese Clinical Trial Registry (ChiCTR2500106350). The requirement for informed consent was waived by the institutional review board for retrospective cohorts involving analysis of deidentified, routinely collected clinical data.
\subsection*{Large-scale Breast Dataset Curation}
We collected 101,638 H\&E-stained breast tissue WSIs from 45,132 patients drawn from 32 geographically diverse sources worldwide, encompassing 21 in-house medical centers and 11 publicly available datasets. The resulting dataset covers the various range of pathological specimens involved in routine clinical practice, including pre-operative needle biopsies, intra-operative frozen sections, and post-operative surgical resection specimens. This multi-stage collection allows for a thorough and continuous characterization of breast histopathological morphology throughout the diagnostic and therapeutic continuum. The curated data were allocated for four distinct usages: Pretraining, Retrospective internal training and testing, and Prospective validation. Details (e.g., slide counts) per center are presented in Extended Data Table ~\ref{tab:data_src}.

\subsubsection*{Pretraining Cohorts for Model Development}
To construct a robust and generalizable foundation model for breast pathology, we curated a massive-scale pretraining dataset comprising 57,271 breast tissue WSIs (over 113 million patch images). These data were aggregated from 15 independent sources, integrating 5 in-house datasets (Centers H1, H2, H3, H7 and H11) with 10 well-established public repositories, including TCGA-BRCA~\cite{cancer2013cancer}, HistAI-Breast~\cite{nechaev2025histai}, BRACS~\cite{brancati2022bracs}, BCNB~\cite{xu2021predicting}, GTEx-Breast~\cite{gtex2020gtex}, ACROBAT2023~\cite{weitz2022acrobat}, BACH~\cite{polonia2019bach}, etc. (Extended Data Table ~\ref{tab:data_pretrain}).

The dataset captures a high degree of histological heterogeneity, encompassing diverse tissue preparations: needle biopsies, frozen sections, and surgical resections. To standardize the data for pretraining, we implemented a uniform sampling protocol. WSIs were processed at $20\times$ magnification, and tissue regions were tiled into non-overlapping patches. We adopted a dynamic sampling cap to balance computational efficiency with data richness: a maximum of 3,000 patches were randomly sampled from each WSI, where all available patches were included for slides yielding fewer than 3,000 valid tissue patches. This strategy ensures comprehensive coverage of small biopsy specimens while maintaining manageable data volumes for large surgical slides.

\subsubsection*{\hypertarget{methods_downstream_eval_cohorts}{Downstream Evaluation Cohorts for Clinical Validation}}\label{sec:methods_downstream_eval_cohorts}
\noindent
\textbf{Data Eligibility.} Across all study components, eligible data comprised adult breast pathology cases with digitized H\&E-stained whole-slide images derived from pre-operative needle biopsies, intra-operative frozen sections, or post-operative surgical resections. All included cases required reference labels derived from routine pathology reports, supplemented by ancillary testing or clinical follow-up when necessary. Cases were excluded if they were deemed non-evaluable for analysis, either due to technical issues (such as failed digitization, severe image corruption, or absence of interpretable tissue) or due to clinical ambiguities where a final definitive diagnostic conclusion could not be established, resulting in a missing reference label for the target task. This eligibility framework was designed to reflect real-world clinical practice rather than an idealized test setting, and therefore retained the diversity of routine specimens, including diagnostically challenging cases.

\noindent
\textbf{Inclusion for Retrospective and Public Cohorts.} Retrospective cohorts were assembled from deidentified archival cases contributed by participating institutions together with eligible public datasets. These cohorts were included when specimen type, H\&E whole-slide images, and task-specific reference labels were available for model development or retrospective evaluation.

\noindent
\textbf{Characteristics of Retrospective Cohorts.} 
To evaluate model performance across diverse retrospective clinical settings, we assembled a large-scale retrospective dataset comprising 44,367 unique WSIs, organized into 82 cohorts for 34 downstream tasks across the breast pathology workflow. These tasks included 11 pre-operative tasks, 4 intra-operative tasks, and 19 post-operative tasks. The dataset was aggregated from 18 independent centers and 2 public repositories (TCGA-BRCA and CPTAC-TCGA\cite{edwards2015cptac}), capturing substantial variability in tissue preparation and patient populations. Details for each cohort are provided in Extended Data Table\ref{tab:center_distr_pre}, \ref{tab:center_distr_intra} and \ref{tab:center_distr_post}.

Internal cohorts were partitioned at the patient level into training, validation and test sets in a 7:1:2 ratio to prevent data leakage, with the validation set used for hyperparameter tuning and the test set reserved for internal evaluation. To assess external generalization, we defined 37 independent retrospective external validation cohorts that were held out from model development and used exclusively for retrospective external evaluation.

\noindent
\textbf{Inclusion for Prospective Cohorts.} Prospective cohorts were enrolled under institution-specific prospective protocols at participating centers. Eligible participants were adult patients undergoing routine biopsy, intra-operative frozen-section assessment, or surgical resection for suspected or confirmed breast lesions, with diagnostic H\&E slides available for digitization and evaluation. All prospective specimens were processed and reported according to each institution's standard clinical workflow, and BRAVE was evaluated in a silent-run manner without influencing diagnostic interpretation, ancillary testing, or patient management.

\noindent
\textbf{Characteristics of Prospective Cohorts.} To evaluate clinical utility and robustness under prospective real-world workflows, we conducted a multi-center prospective validation study comprising 1,558 patients and 4,318 WSIs from three medical centers (H2, H4 and H6), spanning pre-, intra- and post-operative settings. Center H2 contributed an intra-operative cohort of 390 patients (1,707 WSIs) collected between November 2024 and July 2025. Center H4 contributed dual-stage cohorts collected between August 2025 and February 2026 (NFEC-2025-403), including 880 patients (1,696 WSIs) with pre-operative needle biopsies and 114 patients (527 WSIs) with post-operative surgical resections. Center H6 contributed 174 post-operative cases (388 WSIs) collected between August 2025 and October 2025. Collectively, these prospective cohorts comprised 10 tasks across 11 evaluation cohorts (Extended Data Table~\ref{tab:pros_auc}), covering all three stages of the breast pathology workflow.

\label{sec:method_curation}
\subsection*{Downstream Pathological Tasks Across Clinical Stages}
We organized downstream evaluation based on the breast pathology workflow, spanning pre-operative biopsy, intra-operative frozen section, and post-operative surgical resection settings. Across these stages, BRAVE was evaluated within a total of 82 cohorts for routine diagnostic tasks, biomarker and molecular prediction, treatment-response assessment, tumor-behavior inference, and survival risk stratification. The positivity thresholds for each IHC marker were determined according to standard clinical criteria~\cite{cheang2008basal,allison2020estrogen,coates2015tailoring}, as illustrated in Extended Data Table~\ref{tab:threshold_ihc}.

\noindent
\textbf{\hypertarget{methods_reference_standard}{Reference standard and label assignment}.}\label{sec:methods_reference_standard} The reference standard and label assignment closely paralleled clinical practice, varying by workflow stage, task type, and study setting. 

For pre-operative diagnostic tasks, routine diagnostic endpoints (including Malignancy Detection, Mass Differentiation and Metastasis Detection) were defined from the clinical biopsy diagnosis. We used the clinical biopsy diagnosis as the reference standard for these pre-operative diagnostic tasks because their intended role is to support diagnosis at the biopsy stage~\cite{gradishar2024breast}. As a specific exception, for difficult cases where a definitive diagnostic conclusion could not be reached at the initial biopsy (for example, reports indicating only a "suspicious" tendency), the definitive diagnosis established from the final clinical diagnosis, which is based on a comprehensive assessment of factors such as subsequent repeat biopsies, radiological imaging, and ancillary IHC testing, was used as the reference standard.

For pre-operative molecular and biomarker prediction tasks, establishing a reliable ground truth required different strategies depending on the patient's treatment path. Unlike routine biopsy diagnosis, biomarker and molecular assessment may show clinically relevant discordance between biopsy and surgical specimens, and is generally anchored to standardized pathology and ancillary testing criteria~\cite{allison2020estrogen,wolff2023her2,li2025her2discordance,kalvala2022ki67concordance}. Our internal cohorts were therefore deliberately restricted to direct-surgery patients, allowing reference labels to be derived from the final post-operative surgical pathology together with ancillary test results.

Retrospective external validation cohorts, however, included both direct-surgery and neoadjuvant chemotherapy (NAC)-treated patients. For the direct-surgery external cases (specifically, the ER, PR, and HER2 cohorts from H10, and the ER cohort from H21), labels were again assigned based on the definitive post-operative resection findings. By contrast, for the remaining external cases where patients received NAC before surgery, labels were assigned based on the pre-treatment biopsy pathology assessment together with ancillary biomarker test results where available. We applied this policy because NAC can alter tumor morphology and biomarker expression, meaning that the surgical specimen obtained after treatment no longer directly reflects the tumor state present at the time of pre-operative prediction. Thus, for NAC-treated patients, the pre-treatment biopsy provides the closest available reference standard for assessing pre-operative predictions. As a result, we deliberately use different ground truth definitions depending on whether the patient received NAC or direct surgery. Finally, for the NAC response prediction task itself, labels were defined from the final post-operative pathological response assessment.

For intra-operative tasks, all reference labels were defined from the final post-operative diagnosis, because intra-operative frozen-section diagnoses are preliminary and may be revised on final pathology. For post-operative tasks, all reference labels were derived from the diagnosis established on the surgical resection specimen itself, together with ancillary test results where required. For prospective cohorts, intra-operative and post-operative tasks followed the same reference-standard logic. For pre-operative prospective rule-out and rule-in analyses (Figure~\ref{fig:pros}c) and for end-of-study evaluation of reference performance (Figure~\ref{fig:pros}i), labels were defined from the biopsy diagnosis because obtaining the corresponding post-operative reference standard would require a clinically impractical delay. As a specific exception, cases included in the prospective second-review analysis were restricted to direct-surgery patients with available post-operative reference labels.

\subsubsection*{\hypertarget{methods_preop_eval}{Pre-operative Evaluation from Biopsy Specimens}}\label{sec:methods_preop_eval}
In the pre-operative setting, BRAVE was evaluated on H\&E-stained core-needle biopsy specimens for malignancy detection, mass differentiation, metastasis detection, biomarker prediction, molecular subtyping, NAC response prediction, and survival risk stratification.

Diagnostic biopsy tasks included \textbf{Malignancy Detection}, developed on an internal cohort of 838 cases from Center H4 (705 non-malignant and 133 malignant) and prospectively validated on 882 cases from H4 (670 non-malignant and 212 malignant); \textbf{Mass Differentiation}, developed on 838 cases from H4 (415 fibroadenoma, 290 UDH and 133 IDC) and prospectively validated on 646 cases from H4 (413 fibroadenoma, 73 UDH and 160 IDC); and \textbf{Metastasis Detection}, evaluated on 257 lymph node samples from H5 (224 N0+ and 33 N0). Details are provided in Extended Data Tables~\ref{tab:center_distr_pre} and~\ref{tab:cls_distr_pre}.

Biomarker and subtype prediction tasks included \textbf{ER}, \textbf{PR}, \textbf{HER2}, and \textbf{KI67} prediction together with \textbf{Molecular Subtyping}. The biomarker models were trained and internally validated on internal datasets from H5 ranging from 1,331 to 1,485 cases, and underwent retrospective external validation across Centers H4, H8, H10, H17, H19, H20, H21 and H7. For example, the ER prediction task was evaluated on 4,091 external cases from 8 centers. \textbf{Molecular Subtyping} was developed on 1,400 internal cases and externally validated at H4. We also evaluated \textbf{NAC Response Prediction} using 1,534 cases from H5 (1,181 non-pCR and 353 pCR), with labels defined from the final post-operative pathological response assessment. Details are provided in Extended Data Tables~\ref{tab:center_distr_pre} and~\ref{tab:cls_distr_pre}.

For pre-operative survival prediction, an \textbf{Overall Survival (OS)} model was developed using 1,503 cases from H5, and a \textbf{Disease-Free Survival (DFS)} model was trained and internally validated on 303 cases aggregated from H5 and H8. Details are provided in Extended Data Table~\ref{tab:task_dist_surv_data}.

\subsubsection*{Intra-operative Evaluation from Frozen Sections}
In the intra-operative setting, BRAVE was evaluated directly on the original digitized rapid-frozen H\&E slides (rather than reprocessed permanent FFPE sections derived from the frozen tissue remnants) for malignancy detection, margin assessment, mass differentiation, and lymph node metastasis prediction under time-critical decision-making conditions. Data details are provided in Extended Data Tables~\ref{tab:center_distr_intra} and~\ref{tab:cls_distr_intra}.

These tasks included \textbf{Malignancy Detection}, developed on an internal set of 502 cases from H2 (344 non-malignant and 158 malignant), retrospectively validated on 1,994 external cases from H4, H15 and H12+H16, and prospectively evaluated on 107 cases from H2; \textbf{Margin Assessment}, conducted on 478 internal cases from H2 (346 negative and 132 positive), retrospectively validated on 478 external cases from H4 and H12, and prospectively evaluated on 128 cases from H2; \textbf{Mass Differentiation}, developed using 1,256 frozen-section cases from H4, covering fibroadenoma (571), invasive carcinoma (413) and benign lesions (272); and \textbf{Lymph Node Metastasis Prediction}, developed on 499 cases from H2 (298 negative and 201 positive) and retrospectively validated on 493 external cases from H4 and H12. 

Notably, Malignancy Detection and Margin Assessment serve distinct clinical purposes characterized by differences in operational timing, target anatomical site, and expected lesion size. Malignancy Detection is performed in the early intra-operative stage to evaluate the primary breast mass and diagnose the main tumor. In contrast, Margin Assessment occurs in the late intra-operative stage, specifically examining the limited tissue area at the surgical resection edges after the main tumor is excised to detect any minute residual cancer cells, thus ensuring the tumor was completely removed.

\subsubsection*{Post-operative Evaluation from Surgical Resection Specimens}
In the post-operative setting, BRAVE was evaluated on surgical resection specimens for pathological subtyping, tumor grading, pCR assessment, biomarker and molecular prediction, tumor-behavior inference, mutation prediction, TNM-T stage prediction, and prognostic stratification.

Morphological and response-assessment tasks included \textbf{Pathological Subtyping}, developed on 1,053 TCGA cases (842 IDC and 211 ILC) and further validated on 420 retrospective cases from H4 and 156 prospective cases from H6; \textbf{Tumor Grading}, developed on 393 internal cases from H4 (294 G1+G2 and 99 G3) and prospectively validated on 77 cases from H4 and 161 cases from H6; and \textbf{pCR Assessment}, evaluated on 127 cases from H4 (28 pCR and 99 non-pCR). Details are provided in Extended Data Tables~\ref{tab:center_distr_post} and~\ref{tab:cls_distr_post}.

Prediction tasks included \textbf{ER}, \textbf{PR}, \textbf{HER2}, \textbf{KI67}, \textbf{AR}, and \textbf{CK5} prediction; \textbf{Molecular Subtyping}; \textbf{Gene Mutation Prediction}; \textbf{Lymphovascular Invasion (LVI)} prediction; \textbf{Perineural Invasion (PNI)} prediction; and \textbf{TNM-T Stage Prediction}. The biomarker models were developed using internal cohorts from H6 and externally evaluated in retrospective cohorts. For example, the ER model was tested on 1,563 retrospective cases from H4, H20 and H18. \textbf{Molecular Subtyping} was developed on 2,045 internal cases from H6 and retrospectively validated on 1,386 external cases from H13, H9 and H4+H14. \textbf{Gene Mutation} prediction for CDH1, GATA3, PIK3CA and TP53 was developed using TCGA data and externally validated on CPTAC data. \textbf{LVI} and \textbf{PNI} prediction models were developed on internal cohorts from H4 (480 and 478 cases, respectively). A \textbf{TNM-T Stage Prediction} model was established using 467 cases from H6. Details are provided in Extended Data Tables~\ref{tab:center_distr_post} and~\ref{tab:cls_distr_post}.

For post-operative survival prediction, \textbf{DFS} and \textbf{OS} models were trained and internally validated on 451 cases from H6. External validation was performed on retrospective cohorts from H13 and H9, including 79 DFS cases from H13 and 219 OS cases from H13 and H9. Details are provided in Extended Data Table~\ref{tab:task_dist_surv_data}.
\subsection*{Identification of Challenging Scenarios}
\hypertarget{methods_challenge}{}To evaluate robustness in difficult real-world settings, we predefined challenging subgroups separately for the pre-operative, intra-operative, and post-operative stages according to the data source available at each stage. These subgroup candidates were first screened by keyword matching in the corresponding stage-specific pathology reports, followed by manual review to confirm that each selected case was consistent with the intended challenge definition. Cases meeting more than one definition were allowed to contribute to multiple subgroup analyses when appropriate.

We emphasize that all cases included in these subgroups were fully evaluable. Completely non-diagnostic specimens that precluded any definitive outcome were excluded based on our general eligibility criteria (as described in section \hyperlink{methods_downstream_eval_cohorts}{Methods - Data Eligibility}). Specifically, for pre-operative analyses, these challenging scenarios reflect diagnostic difficulties encountered at the initial biopsy. Notably, every included case ultimately achieved a definitive clinical diagnosis based on a comprehensive assessment of various follow-up factors, such as subsequent repeat biopsies, radiological imaging, and ancillary IHC testing. This final conclusion served as the reliable ground truth. Similarly, all challenging intra-operative frozen-section cases possessed a definitive final diagnosis established from subsequent permanent paraffin sections to act as the reference standard.

For pre-operative biopsy analyses, the \textbf{Insufficient Tissue} subgroup was identified using keywords corresponding to ``scant tissue'', ``focal involvement'', ``complete excision'', ``limited sampling'', and ``limited''. The \textbf{Need Extra Info} subgroup was identified using keywords corresponding to ``additional immunohistochemistry'', ``immunohistochemistry required'', ``clinical correlation required'', and ``imaging correlation required''.

For intra-operative frozen-section analyses, the \textbf{NeedIHC} subgroup was identified using keywords corresponding to ``additional immunohistochemistry'' and ``immunohistochemistry required''. The \textbf{Deformation} subgroup was identified using keywords corresponding to ``crush artifact'', ``deformation'', ``cautery artifact'', and ``fragmentation''. The \textbf{NAC} subgroup was identified using keywords corresponding to ``neoadjuvant''. The \textbf{Defer} subgroup was identified using keywords corresponding to ``defer to paraffin section''.

For post-operative analyses, challenging subgroups were not derived from report keywords but from publicly available TCGA clinical and molecular annotations. Specifically, subgroup analyses were performed within clinically relevant subsets defined by hormone receptor and HER2 status or menopausal status, including HR+HER2-, HR-HER2+, premenopausal, and postmenopausal groups, depending on the downstream task.

\subsection*{Retrospective Clinical Impact Simulation}
\subsubsection*{Pre-operative AI-assisted biopsy second review}
We retrospectively simulated a second-review workflow for pre-operative biomarker assessment to test whether BRAVE could help recover biopsy-stage false-negative interpretations before treatment planning. This analysis was restricted to retrospective direct-surgery cohorts with paired pre-operative biopsy slides and corresponding post-operative surgical biomarker reference labels, because the simulation required biopsy-stage pathologist assessments and a definitive surgical reference for the same patient. In the retrospective datasets used for this analysis, ER was evaluated in the H10 and H21 cohorts, whereas PR and HER2 were evaluated in H10.

The simulated workflow followed the scenario illustrated in Figure~\ref{fig:preop}c. Among cases initially assessed as biomarker-negative by pathologists, BRAVE generated a continuous prediction score from the H\&E biopsy slide. Cases with scores above the operating threshold were flagged for second review, whereas cases below threshold remained classified as negative in the simulated workflow. Surgical biomarker status served as the reference standard for determining whether a biopsy-stage negative call was truly negative or represented a missed positive case.

For each biomarker, we scanned candidate thresholds and quantified the trade-off between rescue benefit and additional review workload. The three core operating metrics were defined as follows:

$$
\text{Rescue Rate} = \frac{\text{Rescued Missed Positives}}{\text{False Negatives by Pathologists}}
$$

$$
\text{Review Burden} = \frac{\text{Review Cases}}{\text{Cases Initially Called Negative by Pathologists}}
$$

$$
\text{Number Needed to Review} = \frac{\text{Review Cases}}{\text{Rescued Missed Positives}}
$$

\textbf{False alarm reviews} were cases that were negative by both the initial biopsy assessment and the surgical reference but were nevertheless flagged by BRAVE for second review.

Finally, thresholds were selected under predefined operating constraints requiring a minimum rescue rate of 0.4 and a maximum review burden of 0.4. The selected thresholds were then used to summarize second-review performance for ER, PR, and HER2, and they would serve as the operating points for subsequent prospective evaluation (Extended Data Tables~\ref{tab:rescue_analysis} and~\ref{tab:threshold_er_pr}).
\subsubsection*{Intra-operative safety-constrained rule-out triage}
We retrospectively simulated an AI-assisted rule-out workflow for intra-operative frozen-section assessment to test whether BRAVE could safely exclude low-risk cases from immediate review under time-critical conditions. The primary analysis focused on frozen-section malignancy detection, for which an internal test cohort from H2 and retrospective external cohorts from H4, H15 and H12+H16 were available. The same framework was additionally explored for lymph node metastasis prediction, margin assessment and mass differentiation using their corresponding internal and external frozen-section cohorts.

The simulated workflow followed the triage scenario shown in Figure~\ref{fig:intraop}a,e--h. For each task, BRAVE generated a continuous prediction score from the frozen-section slide, and cases with scores below a candidate rule-out threshold $T_{\mathrm{out}}$ were classified as low risk and hypothetically excluded from immediate review in the simulation. Cases with scores above threshold remained in the routine review pathway. Threshold selection was performed in the internal test cohort and the selected operating threshold was then applied unchanged to external retrospective cohorts to assess transferability.

The main operating metrics were defined as follows:

$$
\text{Rule-out NPV} = \frac{\text{True Rule-out Negatives}}{\text{All Rule-out Cases}}
$$

$$
\text{Rule-out Coverage} = \frac{\text{Rule-out Cases}}{\text{Total Cases}}
$$

For malignancy detection, the operating threshold was selected in the internal test cohort as the threshold achieving a rule-out NPV of 1.0 while maximizing rule-out coverage, and the resulting threshold was then used for external retrospective validation. For the broader task-level analyses, task-specific rule-out thresholds were selected in the same internal-to-external manner and summarized by rule-out NPV, rule-out coverage, rule-out case counts and total case counts (Extended Data Figure~\ref{fig:intra_all_rule_out} and Extended Data Table~\ref{fig:intra_rule_out_results}). In analyses restricted to cases with deferral annotations, we applied the same locked threshold to the deferred cases and recorded whether each case would have been ruled out by BRAVE or would still have remained in the routine review pathway. Deferred cases that were ruled out by BRAVE but were positive by the final reference standard were counted as unsafe rule-out errors, whereas deferred cases that were ruled out by BRAVE and were negative by the final reference standard were counted as safe rescues.

\subsubsection*{Post-operative prioritization of genomic testing}
We retrospectively simulated an AI-guided prioritization workflow for post-operative genomic testing to assess whether BRAVE-derived mutation scores could enrich testing toward higher-risk surgical cases while reducing unnecessary testing in lower-risk patients. This analysis focused on TP53 mutation prediction as an example, using TCGA as the internal cohort for strategy development and CPTAC as the retrospective external validation cohort.

The internal triage analysis compared three prioritization strategies: \textbf{Clinical}, based on routinely available clinicopathologic variables; \textbf{BRAVE only}, based on the H\&E-derived TP53 score alone; and \textbf{Clinical + BRAVE}, based on the integrated use of clinicopathologic variables together with the H\&E-derived TP53 score. The clinicopathologic variables included diagnosis age, clinical tumor stage, histologic subtype, molecular subtype and menopause status. For each strategy, cases were ranked from highest to lowest predicted likelihood of TP53 alteration, and simulated genomic testing was then restricted to the top proportion of cases under a series of intended testing rates.

The main operating metrics were defined as follows:

$$
\text{Sensitivity} = \frac{\text{Mutated Cases Selected for Testing}}{\text{All Mutated Cases}}
$$

$$
\text{PPV} = \frac{\text{Mutated Cases Selected for Testing}}{\text{All Selected Cases}}
$$

$$
\text{Enrichment vs Prevalence} = \frac{\text{Selected Subgroup Mutation Rate}}{\text{Overall Cohort Mutation Rate}}
$$

$$
\text{Tests per Mutation Found} = \frac{\text{All Selected Cases}}{\text{Mutated Cases Selected for Testing}}
$$

Here, enrichment vs prevalence represents how much higher the mutation rate was in the selected-for-testing subgroup than in the full cohort. Performance was evaluated across multiple intended genomic testing rates, defined as the top percentage of the ranked cohort selected for testing. Internal simulations compared the three strategies at matched testing rates to quantify how much mutation yield could be improved by BRAVE alone or by the integrated clinical + BRAVE approach (Extended Data Table~\ref{tab:triage_internal}). For retrospective external validation, fixed score thresholds were first determined from the BRAVE-only ranking in the internal TCGA cohort at prespecified intended testing rates and then applied unchanged to the CPTAC cohort. External performance was summarized by the realized testing rate together with sensitivity, PPV, enrichment over baseline prevalence and tests required per mutation found (Extended Data Table~\ref{tab:triage_ext}).

\subsection*{Prospective Observational Validation}\label{sec:method_pros}

\noindent
\textbf{Model deployment and locked-threshold inference.} To preserve the integrity of the prospective observational validation, BRAVE was deployed with strictly frozen model parameters. No model retraining, hyperparameter tuning, recalibration, or threshold re-optimization was performed after the start of prospective enrollment. All predictions were generated through a fully automated inference pipeline without manual intervention in model execution. Task-specific operating thresholds were determined retrospectively before prospective evaluation and then locked for silent-run validation on consecutive prospective cases.

\noindent
\textbf{Prospective observational design.} We prospectively evaluated whether workflow strategies defined in retrospective analyses remained clinically actionable when transferred to real-world prospective cohorts. The prospective validation was organized across pre-operative, intra-operative, and post-operative settings, with representative tasks selected according to the intended clinical role of AI at each stage. In the pre-operative setting, we evaluated locked triage workflows for malignancy detection and HER2 prediction together with AI-assisted second-review workflows for biomarker assessment. In the intra-operative setting, we evaluated safety-constrained rule-out triage for frozen-section malignancy assessment. In the post-operative setting, we evaluated rule-out triage for pathological subtyping. During the prospective study period, BRAVE predictions were generated in silent-run mode and did not alter the routine diagnostic workflow.

\noindent
\textbf{\hypertarget{methods_pros_triage}{Prospective validation of locked triage workflows.}}\label{sec:methods_pros_triage} For triage tasks, low and high operating thresholds were first defined in retrospective internal cohorts and were then applied unchanged to the prospective cohorts. For rule-out workflows, cases with prediction scores below the locked low threshold were considered eligible for exclusion from routine review in the simulated workflow, and performance was summarized by rule-out coverage together with the negative predictive value at the rule-out threshold. For rule-in workflows, cases with prediction scores above the locked high threshold were considered eligible for prioritization in the simulated workflow, and performance was summarized by rule-in coverage together with the positive predictive value at the rule-in threshold.

For pre-operative malignancy detection, the low threshold $T_{\mathrm{low}}$ was selected under the constraint $\mathrm{NPV} \geq 1.0$ in the retrospective internal cohort, and among thresholds satisfying this condition we selected the largest threshold so as to maximize rule-out coverage. The high threshold $T_{\mathrm{high}}$ was selected under the constraint $\mathrm{PPV} \geq 1.0$, and among thresholds satisfying this condition we again selected the largest threshold so as to keep the rule-in subset as restrictive as possible and thereby minimize unnecessary resource use.

For pre-operative HER2 prediction, the low threshold $T_{\mathrm{low}}$ was defined using the same rule-out principle, namely $\mathrm{NPV} \geq 1.0$ with selection of the largest threshold among eligible candidates. By contrast, the high threshold $T_{\mathrm{high}}$ was selected under the constraint $\mathrm{PPV} \geq 0.90$, and among thresholds satisfying this condition we selected the smallest threshold so as to maximize rule-in coverage while maintaining high clinical certainty. This choice was intended to ensure that, among patients already diagnosed with breast cancer, as many HER2-positive candidates as possible were brought to additional attention for treatment-related assessment.

For post-operative pathological subtyping, the rule-out threshold $T_{\mathrm{low}}$ was selected under the constraint $\mathrm{NPV} \geq 1.0$, with the largest eligible threshold chosen to maximize rule-out coverage. The same locked thresholds were carried forward across these clinical settings without prospective retuning, and their prospective triage performance was summarized in Extended Data Table~\ref{tab:pros_triage_threshold_perform}.

\noindent
\textbf{Prospective validation of AI-assisted biomarker second review.} \label{sec:methods_pros_second_review} We further prospectively evaluated whether locked second-review thresholds could rescue biopsy-stage missed biomarker-positive cases among specimens initially assessed as negative in routine practice. In this analysis, routine biopsy-based biomarker labels were compared with the corresponding post-operative surgical biomarker labels after they became available. As in the retrospective simulation, only cases initially called negative by pathologists entered the AI-assisted second-review workflow. Cases with BRAVE scores above the locked threshold were flagged for second review, whereas cases below threshold remained negative in the simulated workflow. Rescue rate, review burden, and number needed to review were defined identically to the retrospective analysis.

For ER and PR, the second-review thresholds were carried forward from retrospective second-review analyses, in which thresholds were selected under the joint constraints of minimum rescue rate $\geq 0.4$ and maximum review burden $\leq 0.4$. For Ki67, given prior reports of substantial discordance between core needle biopsy and surgical specimens in breast cancer\cite{kalvala2022ki67concordance}, the threshold was selected retrospectively in the internal cohort before prospective evaluation under a conservative sensitivity constraint of $\mathrm{Sensitivity} \geq 0.98$ to minimize the risk of missing truly positive cases.

\noindent
\textbf{Reference performance after prospective ground-truth collection.} After completion of the prospective study period and subsequent availability of the corresponding ground-truth labels, we additionally summarized task-level predictive performance in the prospective cohorts for reference. These analyses were not used for workflow definition or threshold selection, because all operating thresholds had already been locked before prospective data collection. Instead, they served as post-study contextual evaluations of whether BRAVE retained informative prospective performance across representative tasks from the three clinical stages. The resulting case numbers, class distributions, and Macro-AUC values are reported in Extended Data Table~\ref{tab:pros_auc}.
\subsection*{Human-AI Collaboration: Crossover Reader Study}
\textbf{Study design and participants}. To quantify the clinical utility of BRAVE in direct pathologist-AI interaction, we conducted a randomized crossover multi-reader study spanning pre-operative, intra-operative and post-operative diagnostic tasks. Eight board-certified pathologists participated. Readers were stratified by experience level into junior and senior groups, and the intervention sequence was balanced by experience: two junior and two senior pathologists completed the AI-assisted condition first, whereas the remaining two junior and two senior pathologists completed the unassisted condition first. After the first reading phase, a mandatory 4-week washout period was imposed before crossover to the alternate condition. During washout, readers did not review the study cases or discuss them with colleagues. Pathologists were blinded to the study hypotheses and to the reference diagnoses.

\noindent
\textbf{\hypertarget{methods_reader_tasks}{Reading tasks and reference standard}.}\label{sec:methods_reader_tasks} The reader study included three representative tasks across the clinical workflow: pre-operative malignancy detection on biopsy specimens ($N=100$ cases, 186 WSIs), intra-operative malignancy detection on frozen sections ($N=100$ cases, 148 WSIs), and post-operative pCR assessment on surgical specimens ($N=39$ cases, 542 WSIs). For each task, reader-study cases were selected by class-balanced random sampling from the corresponding collected test cases to ensure a balanced case composition for reader evaluation. In both reading conditions, readers reviewed the digitized H\&E slides and rendered a case-level diagnostic judgment for the assigned task. Reference labels were taken from the final reports generated through the standard routine clinical workflow for the corresponding cases.

Particularly, for the intra-operative task, to mimic the time-critical nature of frozen-section consultation, the reading interface enforced a 10-minute time limit per case. If a final diagnosis was not submitted within this window, the case was automatically closed by the system. For such timed-out cases, reading time was recorded up to automatic timeout, the case was counted as an incorrect response in the accuracy analysis, and automatic timeout was treated as a separate class in the inter-rater agreement analysis.

\noindent
\textbf{AI-assisted and unassisted reading conditions}. In the unassisted condition, pathologists interpreted the H\&E slides without model support. In the AI-assisted condition, pathologists reviewed the same slides together with BRAVE outputs, including the predicted category with probabilities, and an explanatory heatmap. Each reader completed both conditions under the crossover design for within-reader comparison of AI-assisted and unassisted performance.

\noindent
\textbf{Outcome measures}. The primary endpoint was diagnostic accuracy, defined as concordance between the reader's case-level diagnosis and the reference label. Secondary endpoints included diagnostic confidence, recorded on a 10-point Likert scale, and reading time, recorded from case opening to result submission. Inter-rater agreement across readers was additionally quantified using Fleiss' $\kappa$ to assess whether AI assistance improved consistency of interpretation across the reader cohort. For the intra-operative task, specific analytical rules were applied to handle cases that reached the automatic timeout. For these timed-out cases, the case was counted as an incorrect response in the accuracy analysis, and reading time was recorded up to the automatic timeout. For the confidence analysis, if a reader-case pair timed out in either reading condition, that pair was excluded from confidence comparison in both conditions to preserve paired comparability. Finally, automatic timeout was treated as a separate class in the inter-rater agreement analysis.

\subsection*{Model Development}
\subsubsection*{Data Preprocessing}
Each digitized whole-slide image (WSI) was processed as follows: (1) Tissue segmentation: A downsampled version of the slide was converted to the HSV color space, and binary thresholding was applied to the saturation channel to generate a tissue mask. This mask was used to isolate histologically relevant tissue regions. (2) Patching: The WSI was tiled at 20$\times$ magnification ($\sim 0.5 \mu m/\text{px}$)  into non-overlapping $256\times 256$ pixel patches, which were then resized to $224\times 224$ pixels for further training.
\subsubsection*{Network Architecture}
BRAVE employs the vision transformer~\cite{dosovitskiyimage} (ViT) `Huge' (ViT-H/14) architecture consisting of 632 million parameters. The ViT adapts the transformer architecture for image analysis by treating an image as a sequence of embedded patches, which are then processed through a transformer encoder utilizing self-attention mechanisms. The model processes each $224\times 224$-pixel image patch, wherein the patch is tokenized into a sequence of $14\times 14$-pixel tokens. The processed tokens are then passed through 32 transformer layers, each with a hidden size of 1,280 and 16 attention heads. The final output is a concatenation of a 1,280-dimensional feature vector derived from the [CLS] token and the mean embeddings for all patch tokens, resulting in an embedding size of 2,560.

Building on Virchow2~\cite{vorontsov2024foundation}, a state-of-the-art pan-cancer pathology foundation model pretrained on 3.1 million histology slides with diverse stains and cancer types, BRAVE further refines this model to enhance its performance on breast-specific pathology while preserving its broad generalizability. To this end, Low-Rank Adaptation (LoRA)~\cite{hu2022lora} is employed for adaptive pretraining. LoRA is a parameter-efficient fine-tuning method that operates by injecting trainable low-rank matrices into the self-attention layers of a transformer, while keeping the original weights frozen to preserve the previously learned pan-cancer knowledge. Specifically, each query or value projection was parameterized as $W' = W + \frac{\alpha}{r}BA$, where $W$ denotes the original frozen pretrained projection matrix, $W'$ denotes the adapted projection matrix after adding the LoRA update, $A$ and $B$ are trainable low-rank matrices, $r$ is the LoRA rank, and $\alpha$ is the scaling factor. This design significantly reduces trainable parameters, computational cost, and the risk of catastrophic forgetting. In this study, we integrated LoRA modules into the query and value projection matrices of each attention layer and set up the modules with the rank of 8 and a scaling factor of 16, resulting in only 2 million trainable parameters. Details of the network architecture are provided in the Extended Data Figure~\ref{fig:method}.

\subsubsection*{Adaptive Pretraining on Breast Tissue Slides}
To specialize the general-purpose pathology foundation model for breast pathology, BRAVE underwent breast-focused adaptive pretraining using 57,271 whole-slide images of breast tissue (more than 113 million image patches) collected from 15 geographically diverse data sources spanning North America, Europe, and Asia (Extended Data Table~\ref{tab:data_pretrain}). BRAVE employed DINO~\cite{caron2021emerging} (self-DIstillation with NO labels) to facilitate robust representation learning for pathological images, as shown in Extended Data Figure~\ref{fig:method}. During this process, whole-slide images were segmented and patched to generate tissue tiles, from which paired global and local views were sampled for self-supervised training. DINO~\cite{caron2021emerging} leverages a teacher-student knowledge distillation paradigm, in which the student network is optimized by back-propagation to match the output of a momentum-updated teacher network, while the teacher is updated by exponential moving average with a momentum coefficient set to 0.9995 and does not receive direct gradient updates. By distilling local cellular details into global tissue contexts across augmented views of tissue patches, this paradigm effectively captures intrinsic morphological phenotypes, enabling the model to learn generalizable features without the need for manual annotation. The model was optimized with AdamW ($\beta_1=0.9$, $\beta_2=0.999$) using mixed-precision (FP16) arithmetic to expedite training and lower memory usage. Pretraining was performed on eight 80GB NVIDIA H800 GPUs with a batch size of 384 over 882,000 iterations. Throughout adaptation, the pretrained Virchow2 backbone weights remained frozen and only the LoRA adaptation weights were updated. After adaptive pretraining, the frozen pretrained backbone together with the learned LoRA weights formed the final BRAVE encoder used for patch-level feature extraction in downstream tasks.

\subsubsection*{Aggregation Models for Downstream Tasks}
After data processing, following the typical paradigm of Multiple instance learning (MIL) in computational pathology, every case undergoes two steps, including (1) feature extraction from patch images and (2) feature aggregation for case-level decision-making. For patch feature extraction, with pretrained BRAVE after adaptive learning on breast tissue data, each patch image was processed through the weight-frozen BRAVE ViT encoder to obtain a 2,560-dimensional embedding. To obtain the case-level representation, we employ ABMIL~\cite{ilse2018attention}, an attention-based MIL algorithm, which treats all WSIs of a case as a ``bag'' composed of patch-level instances. In this framework, the case-level label is assigned to the entire bag, while the labels of individual patches remain unknown. Finally, the case representation aggregated from all patch images was fed into a head consisting of a single fully connected classification layer with a dropout rate of 0.25, to generate the final prediction for each case. Notably, for cases comprising multiple WSIs, all available slides were concatenated and collectively inputted into the model to generate a case-level prediction. To optimize the learning of aggregation models, Cross-entropy loss was applied for classification tasks, while Negative Log-Likelihood (NLL) loss~\cite{zadeh2020bias} was employed for survival prediction. To preserve the knowledge from pretraining on breast tissues, only the ABMIL aggregator and the head layer were trained from scratch, while the BRAVE encoder was kept frozen. Optimization was performed with the Adam optimizer at a learning rate of $2\times 10^{-4}$ and a weight decay of $1\times 10^{-5}$. To avoid overfitting and halt training when validation performance plateaued, early stopping based on validation loss was applied. In all downstream task comparisons, the same training and evaluation protocols were used across all baseline models to ensure a fair and consistent performance assessment.
\subsubsection*{Statistical Analysis}
\textbf{Performance metrics and comparative benchmarking}. For retrospective and prospective classification tasks, model discrimination was summarized using the area under the receiver operating characteristic curve (AUC). Binary tasks were further described by sensitivity and specificity at the reported operating points, whereas multi-class tasks were summarized by macro-AUC computed with a one-versus-rest strategy and averaged across classes. For workflow-oriented triage and second-review analyses, performance was additionally summarized using task-specific operating metrics defined in the corresponding Methods subsections, including negative predictive value (NPV), positive predictive value (PPV), rule-out or rule-in coverage, rescue rate, review burden, and number needed to review. In the foundation-model comparison, task-level differences between BRAVE and Virchow2 were assessed using one-sided Wilcoxon signed-rank tests~\cite{woolson2007wilcoxon} on mean macro-AUC values aggregated at the task level.

\noindent
\textbf{Robustness, paired-agreement, and post-operative association analyses}. In challenging-scenario analyses, predictive score distributions between subgroup cohorts and their corresponding overall cohorts were compared using the Kolmogorov-Smirnov test, with subgroup AUCs reported together with 95\% confidence intervals. In the prospective paired biomarker analysis, agreement between pre-operative biopsy labels and post-operative surgical labels was summarized by concordance and Cohen's kappa, and directional asymmetry among discordant paired classifications was evaluated using McNemar testing. For post-operative molecular analyses, associations between H\&E-derived TP53 scores and continuous genomic-instability-related variables were quantified using Spearman correlation, whereas concordance between binned AI-score trends and observed mutation rates was assessed using Pearson correlation. Independent predictive value beyond clinicopathologic variables was evaluated using multivariable logistic regression, with odds ratios and 95\% confidence intervals reported. Incremental value after adding the AI-derived score was assessed by likelihood-ratio testing and summarized by changes in AUROC, AUPRC, and Brier score.

\noindent
\textbf{Survival analysis}. For prognostic tasks, discrimination was measured using the concordance index (C-index). In the internal setting, survival models were evaluated within a 5-fold cross-validation framework. For external cohorts, predictions from the five fold-specific models were averaged to obtain the final risk score for each patient. Prognostic effect sizes were summarized as hazard ratios (HRs) using univariable and multivariable Cox proportional hazards regression~\cite{harrell2001cox}, where HRs greater than 1 indicate higher event risk in the higher-risk group or per unit increase in the corresponding predictor. A confidence interval that excluded 1 was taken as indicating a statistically significant association. Kaplan-Meier curves were used to visualize risk-group separation, and differences between survival curves were assessed with the log-rank test~\cite{ziegler2007survival}. To visualize adjusted survival separation independent of routine biomarkers, adjusted survival curves were generated from the multivariable Cox models using direct adjustment with Breslow baseline hazard estimation.

\noindent
\textbf{Statistical analysis of human-AI reader study}. In the crossover multi-reader study, reader performance was descriptively summarized by balanced accuracy, diagnostic confidence, and reading time. To account for repeated measurements clustered within readers, we used generalized estimating equations (GEE) with an exchangeable correlation structure and pathologist as the clustering unit~\cite{hardin2002generalized}. Models were adjusted for reading condition, period, task, and reader experience level, with the unassisted condition as the reference. Diagnostic accuracy was reported as an odds ratio (OR), where values greater than 1 indicate higher odds of correct diagnosis under AI assistance. Reading time was reported as a time ratio (TR), where values less than 1 indicate shorter reading time with AI assistance. Confidence was reported as a mean difference, where positive values indicate higher confidence in the AI-assisted condition. A confidence interval that excluded the null value (1 for ratio-based measures, 0 for mean differences) was taken as indicating a statistically significant difference. Sequence effects in the crossover design were examined using additional GEE analyses performed overall and separately for each task. Inter-rater agreement was quantified using Fleiss' kappa ($\kappa$)~\cite{falotico2015fleiss}, where higher values indicate stronger agreement among readers.

For the intra-operative task, specific analytical rules were applied to handle cases that reached the automatic timeout. Timed-out cases were treated as incorrect responses in the accuracy analysis. Reading time for timed-out cases was recorded as the observed time to automatic timeout. Confidence analyses were performed on paired completed reads only, such that if a reader-case pair timed out in either condition, that reader-case pair was excluded from the confidence comparison in both conditions. Accordingly, the estimated effect of AI assistance on confidence should be interpreted conditionally on reader-case pairs that were completed under both reading conditions. Finally, automatic timeout was treated as an additional response category in the kappa calculation so that all reader-case observations remained analyzable within each reading condition.

\noindent
\textbf{Uncertainty quantification and significance testing}. Unless otherwise specified, 95\% confidence intervals for classification metrics were estimated using non-parametric bootstrapping with 1,000 resamples. For survival analyses, 1,000 bootstrap replicates were generated within each fold of the 5-fold scheme, yielding 5,000 total replicates per dataset. In the reader study, the significance of the difference in Fleiss' kappa between AI-assisted and unassisted readings was evaluated using 1,000 non-parametric bootstrap replicates of $\Delta\kappa$. Reported P values for the remaining analyses were obtained from the corresponding statistical procedures described above, including GEE, Cox regression, log-rank testing, Kolmogorov-Smirnov testing, McNemar testing, correlation analysis, and likelihood-ratio testing. A P-value of less than 0.05 was considered statistically significant.
\section*{Data Availability}
This study incorporates a total of 82 datasets from 32 data source including 21 in-house medical institutions and 11 public repositories. The source of public datasets are summarized in Extended Data Table~\ref{tab:data_src} and~\ref{tab:link_public}. For private cohorts (H1-H21), these datasets are not publicly available due to patient privacy obligations, institutional review board requirements, and data use agreements. However, researchers interested in accessing deidentified data may submit a request directly to the corresponding author, subject to obtaining the necessary ethical approvals and complying with institutional policies.
\section*{Code Availability}
The code and weights of BRAVE will be released upon acceptance.
\section*{Author contributions}
Y.X., H.C. and Z.Z. conceived the study and designed the stage-specific tasks. Y.X. prepared the manuscript and figures, designed and performed the experiments, preprocessed data, conducted model pretraining and statistical analysis. Z.Z. and X.Z. coordinated the prospective study and enrolled patients, provided pathological expertise, and curated WSI dataset. M.X. collected WSI data and provided pathological expertise. F.Z. assisted in model pretraining and preprocessed intra-operative data. Y.W. extracted WSI features and coordinated data collection. J.M. coordinated data collection and preprocessed pretraining data. Y.X. preprocessed biopsy data. D.L., Z.C., and Y.T. participated as pathologists in the reader study. Z.C., Y.T. and R.M collected WSI data. C.L. set up the platform for the reader study. Q.Y. coordinated the external validation. Z.G., Y.Z., J.C., F.L., Q.X., Y.D., and H.T. collected and curated WSI datasets. J.C. developed the platform for the reader study. H.Z, Z.G., L.L and H.W. coordinated data collection. Y.C., and X.W. participated in discussions on experiments design. Z.L., R.C.K.C, N.M., Y.C. Z.W., L.L., and H.C oversee data collection and quality control processes, and supervised the study.
\section*{Acknowledgements}
We sincerely appreciate the pathologists, Dr. Zhen Wang, Dr. Qi Xie, Dr. Rui Mao, Dr. Weihao Qiu and Dr. Feifei Wang, as well as additional clinicians and pathologists who participated in the execution of the reader study but are not named here. This work was supported by the National Natural Science Foundation of China (Project No. 62202403), Research Grants Council of the Hong Kong Special Administrative Region, China (Project R6003-22 and C4024-22GF), National Key R\&D Program of China (Project No. 2023YFE0204000), Hong Kong Innovation and Technology Commission (Project No. MHP/002/22 and ITCPD/17-9), HKUST-HKUST(GZ) Cross-Campus Collaborative Research Scheme (Project No. C036) and  Guangdong Provincial Department of Science and Technology's `1+1+1' Joint Funding Program for Guangdong-Hong Kong Universities.
\section*{Competing interests}
The authors declare no competing interests.

\bibliography{ref}

@article{allison2020estrogen,
  title={Estrogen and progesterone receptor testing in breast cancer: ASCO/CAP guideline update},
  author={Allison, Kimberly H and Hammond, M Elizabeth H and Dowsett, Mitchell and McKernin, Shannon E and Carey, Lisa A and Fitzgibbons, Patrick L and Hayes, Daniel F and Lakhani, Sunil R and Chavez-MacGregor, Mariana and Perlmutter, Jane and others},
  journal={Journal of Clinical Oncology},
  volume={38},
  number={12},
  pages={1346--1366},
  year={2020},
  publisher={American Society of Clinical Oncology}
}

@article{wolff2023her2,
  title={Human Epidermal Growth Factor Receptor 2 Testing in Breast Cancer: ASCO-College of American Pathologists Guideline Update},
  author={Wolff, Antonio C and Somerfield, Mark R and Dowsett, Mitchell and Hammond, M Elizabeth H and Hayes, Daniel F and McShane, Lisa M and Saphner, Thomas J and Spears, Patricia A and Allison, Kimberly H},
  journal={Journal of Clinical Oncology},
  volume={41},
  number={22},
  pages={3867--3872},
  year={2023},
  doi={10.1200/JCO.22.02864}
}

@article{li2025her2discordance,
  title={Illuminating the clinicopathological and genomic landscape of HER2-null, ultralow, and low breast cancers: insights into diagnostic discordance between biopsy and surgical excision},
  author={Li, Ming and Cai, Meng-Yuan and Lv, Hong and Zhou, Shu-Ling and Zhu, Yi-Fei and Shui, Ruo-Hong and Yang, Wen-Tao},
  journal={npj Breast Cancer},
  year={2025},
  publisher={Nature Publishing Group UK London}
}

@article{cheang2008basal,
  title={Basal-like breast cancer defined by five biomarkers has superior prognostic value than triple-negative phenotype},
  author={Cheang, Maggie CU and Voduc, David and Bajdik, Chris and Leung, Samuel and McKinney, Steven and Chia, Stephen K and Perou, Charles M and Nielsen, Torsten O},
  journal={Clinical cancer research},
  volume={14},
  number={5},
  pages={1368--1376},
  year={2008},
  publisher={American Association for Cancer Research}
}

@article{kalvala2022ki67concordance,
  title={Concordance between core needle biopsy and surgical excision specimens for Ki-67 in breast cancer - a systematic review of the literature},
  author={Kalvala, Jahnavi and Parks, Ruth M and Green, Andrew R and Cheung, Kwok-Leung},
  journal={Histopathology},
  volume={80},
  number={3},
  pages={468--484},
  year={2022},
  doi={10.1111/his.14555}
}

@article{ma2025generalizable,
  title={A generalizable pathology foundation model using a unified knowledge distillation pretraining framework},
  author={Ma, Jiabo and Guo, Zhengrui and Zhou, Fengtao and Wang, Yihui and Xu, Yingxue and Li, Jinbang and Yan, Fang and Cai, Yu and Zhu, Zhengjie and Jin, Cheng and others},
  journal={Nature Biomedical Engineering},
  pages={1--20},
  year={2025},
  publisher={Nature Publishing Group UK London}
}

@article{xu2025multimodal,
  title={A multimodal knowledge-enhanced whole-slide pathology foundation model},
  author={Xu, Yingxue and Wang, Yihui and Zhou, Fengtao and Ma, Jiabo and Jin, Cheng and Yang, Shu and Li, Jinbang and Zhang, Zhengyu and Zhao, Chenglong and Zhou, Huajun and others},
  journal={Nature Communications},
  year={2025},
  publisher={Nature Publishing Group UK London}
}

@article{wang2024pathology,
  title={A pathology foundation model for cancer diagnosis and prognosis prediction},
  author={Wang, Xiyue and Zhao, Junhan and Marostica, Eliana and Yuan, Wei and Jin, Jietian and Zhang, Jiayu and Li, Ruijiang and Tang, Hongping and Wang, Kanran and Li, Yu and others},
  journal={Nature},
  volume={634},
  number={8035},
  pages={970--978},
  year={2024},
  publisher={Nature Publishing Group UK London}
}

@article{xu2024whole,
  title={A whole-slide foundation model for digital pathology from real-world data},
  author={Xu, Hanwen and Usuyama, Naoto and Bagga, Jaspreet and Zhang, Sheng and Rao, Rajesh and Naumann, Tristan and Wong, Cliff and Gero, Zelalem and Gonz{\'a}lez, Javier and Gu, Yu and others},
  journal={Nature},
  volume={630},
  number={8015},
  pages={181--188},
  year={2024},
  publisher={Nature Publishing Group UK London}
}

@article{vorontsov2024foundation,
  title={A foundation model for clinical-grade computational pathology and rare cancers detection},
  author={Vorontsov, Eugene and Bozkurt, Alican and Casson, Adam and Shaikovski, George and Zelechowski, Michal and Severson, Kristen and Zimmermann, Eric and Hall, James and Tenenholtz, Neil and Fusi, Nicolo and others},
  journal={Nature medicine},
  volume={30},
  number={10},
  pages={2924--2935},
  year={2024},
  publisher={Nature Publishing Group US New York}
}

@inproceedings{caron2021emerging,
  title={Emerging properties in self-supervised vision transformers},
  author={Caron, Mathilde and Touvron, Hugo and Misra, Ishan and J{\'e}gou, Herv{\'e} and Mairal, Julien and Bojanowski, Piotr and Joulin, Armand},
  booktitle={Proceedings of the IEEE/CVF international conference on computer vision},
  pages={9650--9660},
  year={2021}
}

@article{gradishar2024breast,
  title={Breast cancer, version 3.2024, NCCN clinical practice guidelines in oncology},
  author={Gradishar, William J and Moran, Meena S and Abraham, Jame and Abramson, Vandana and Aft, Rebecca and Agnese, Doreen and Allison, Kimberly H and Anderson, Bethany and Bailey, Janet and Burstein, Harold J and others},
  journal={Journal of the National Comprehensive Cancer Network},
  volume={22},
  number={5},
  pages={331--357},
  year={2024},
  publisher={National Comprehensive Cancer Network}
}

@article{chen2024towards,
  title={Towards a general-purpose foundation model for computational pathology},
  author={Chen, Richard J and Ding, Tong and Lu, Ming Y and Williamson, Drew FK and Jaume, Guillaume and Song, Andrew H and Chen, Bowen and Zhang, Andrew and Shao, Daniel and Shaban, Muhammad and others},
  journal={Nature medicine},
  volume={30},
  number={3},
  pages={850--862},
  year={2024},
  publisher={Nature Publishing Group US New York}
}

@article{lu2024visual,
  title={A visual-language foundation model for computational pathology},
  author={Lu, Ming Y and Chen, Bowen and Williamson, Drew FK and Chen, Richard J and Liang, Ivy and Ding, Tong and Jaume, Guillaume and Odintsov, Igor and Le, Long Phi and Gerber, Georg and others},
  journal={Nature medicine},
  volume={30},
  number={3},
  pages={863--874},
  year={2024},
  publisher={Nature Publishing Group US New York}
}

@article{zhao2025clinical,
  title={A clinical-grade universal foundation model for intraoperative pathology},
  author={Zhao, Zihan and Zhou, Fengtao and Li, Ronggang and Chu, Bing and Zhang, Xinke and Zheng, Xueyi and Zheng, Ke and Wen, Xiaobo and Ma, Jiabo and Wang, Yihui and others},
  journal={arXiv preprint arXiv:2510.04861},
  year={2025}
}

@article{yan2025pathorchestra,
  title={Pathorchestra: A comprehensive foundation model for computational pathology with over 100 diverse clinical-grade tasks},
  author={Yan, Fang and Wu, Jianfeng and Li, Jiawen and Wang, Wei and Chen, Yirong and Wei, Linda and Lu, Jiaxuan and Chen, Wen and Gao, Zizhao and Li, Jianan and others},
  journal={npj Digital Medicine},
  volume={8},
  number={1},
  pages={695},
  year={2025},
  publisher={Nature Publishing Group UK London}
}

@article{huang2023visual,
  title={A visual--language foundation model for pathology image analysis using medical twitter},
  author={Huang, Zhi and Bianchi, Federico and Yuksekgonul, Mert and Montine, Thomas J and Zou, James},
  journal={Nature medicine},
  volume={29},
  number={9},
  pages={2307--2316},
  year={2023},
  publisher={Nature Publishing Group US New York}
}

@article{campanella2025clinical,
  title={A clinical benchmark of public self-supervised pathology foundation models},
  author={Campanella, Gabriele and Chen, Shengjia and Singh, Manbir and Verma, Ruchika and Muehlstedt, Silke and Zeng, Jennifer and Stock, Aryeh and Croken, Matt and Veremis, Brandon and Elmas, Abdulkadir and others},
  journal={Nature Communications},
  volume={16},
  number={1},
  pages={3640},
  year={2025},
  publisher={Nature Publishing Group UK London}
}

@article{neidlinger2025benchmarking,
  title={Benchmarking foundation models as feature extractors for weakly supervised computational pathology},
  author={Neidlinger, Peter and El Nahhas, Omar SM and Muti, Hannah Sophie and Lenz, Tim and Hoffmeister, Michael and Brenner, Hermann and van Treeck, Marko and Langer, Rupert and Dislich, Bastian and Behrens, Hans Michael and others},
  journal={Nature biomedical engineering},
  pages={1--11},
  year={2025},
  publisher={Nature Publishing Group UK London}
}

@article{campanella2025real,
  title={Real-world deployment of a fine-tuned pathology foundation model for lung cancer biomarker detection},
  author={Campanella, Gabriele and Kumar, Neeraj and Nanda, Swaraj and Singi, Siddharth and Fluder, Eugene and Kwan, Ricky and Muehlstedt, Silke and Pfarr, Nicole and Sch{\"u}ffler, Peter J and H{\"a}ggstr{\"o}m, Ida and others},
  journal={Nature Medicine},
  volume={31},
  number={9},
  pages={3002--3010},
  year={2025},
  publisher={Nature Publishing Group US New York}
}

@article{huang2025pathologist,
  title={A pathologist--AI collaboration framework for enhancing diagnostic accuracies and efficiencies},
  author={Huang, Zhi and Yang, Eric and Shen, Jeanne and Gratzinger, Dita and Eyerer, Frederick and Liang, Brooke and Nirschl, Jeffrey and Bingham, David and Dussaq, Alex M and Kunder, Christian and others},
  journal={Nature Biomedical Engineering},
  volume={9},
  number={4},
  pages={455--470},
  year={2025},
  publisher={Nature Publishing Group UK London}
}

@article{hu2022lora,
  title={Lora: Low-rank adaptation of large language models.},
  author={Hu, Edward J and Shen, Yelong and Wallis, Phillip and Allen-Zhu, Zeyuan and Li, Yuanzhi and Wang, Shean and Wang, Lu and Chen, Weizhu and others},
  journal={ICLR},
  volume={1},
  number={2},
  pages={3},
  year={2022}
}

@article{de2023perspectives,
  title={Perspectives on validation of clinical predictive algorithms},
  author={de Hond, Anne AH and Shah, Vaibhavi B and Kant, Ilse MJ and Van Calster, Ben and Steyerberg, Ewout W and Hernandez-Boussard, Tina},
  journal={NPJ digital medicine},
  volume={6},
  number={1},
  pages={86},
  year={2023},
  publisher={Nature Publishing Group UK London}
}

@article{you2025clinical,
  title={Clinical trials informed framework for real world clinical implementation and deployment of artificial intelligence applications},
  author={You, Jacqueline G and Hernandez-Boussard, Tina and Pfeffer, Michael A and Landman, Adam and Mishuris, Rebecca G},
  journal={NPJ Digital Medicine},
  volume={8},
  number={1},
  pages={107},
  year={2025},
  publisher={Nature Publishing Group UK London}
}

@book{hardin2002generalized,
  title={Generalized estimating equations},
  author={Hardin, James W and Hilbe, Joseph M},
  year={2002},
  publisher={chapman and hall/CRC}
}

@article{falotico2015fleiss,
  title={Fleiss’ kappa statistic without paradoxes},
  author={Falotico, Rosa and Quatto, Piero},
  journal={Quality \& Quantity},
  volume={49},
  number={2},
  pages={463--470},
  year={2015},
  publisher={Springer}
}

@incollection{harrell2001cox,
  title={Cox proportional hazards regression model},
  author={Harrell Jr, Frank E},
  booktitle={Regression Modeling Strategies: With Applications to Linear Models, Logistic Regression, and Survival Analysis},
  pages={465--507},
  year={2001},
  publisher={Springer}
}

@article{nechaev2025histai,
  title={HISTAI: An Open-Source, Large-Scale Whole Slide Image Dataset for Computational Pathology},
  author={Nechaev, Dmitry and Pchelnikov, Alexey and Ivanova, Ekaterina},
  journal={arXiv preprint arXiv:2505.12120},
  year={2025}
}

@article{cancer2013cancer,
  title={The cancer genome atlas pan-cancer analysis project},
  author={Cancer Genome Atlas Research Network, JN and others},
  journal={Nat. Genet},
  volume={45},
  number={10},
  pages={1113--1120},
  year={2013}
}

@article{tcga2012breast,
  title={Comprehensive molecular portraits of human breast tumours},
  author={Cancer Genome Atlas Network and others},
  journal={Nature},
  volume={490},
  number={7418},
  pages={61--70},
  year={2012}
}

@article{griffith2018erpositive,
  title={The prognostic effects of somatic mutations in ER-positive breast cancer},
  author={Griffith, Olivia L and Spies, Nicholas C and Anurag, Meenakshi and Griffith, Malachi and Luo, Jun and Tu, Dongsheng and Yeo, Bernice and Kunisaki, Joyce and Miller, Christopher A and Krysiak, Katherine and Hundal, Jathin and Ainscough, Benjamin J and Skidmore, Zachary L and Campbell, Kelsey and Kumar, Rohit and Fronick, Catrina and Cook, Leia and Snider, Julie E and Davies, Simon and Kavuri, Srikanth M and Chang, Eric C and Magrini, Vincent and Larson, David E and Fulton, Robert S and Liu, Shuang and Leung, Samuel and Voduc, David and Bose, Ron and Dowsett, Mitch and Wilson, Richard K and Nielsen, Torsten O and Mardis, Elaine R and Ellis, Matthew J},
  journal={Nature Communications},
  volume={9},
  number={1},
  pages={3476},
  year={2018},
  doi={10.1038/s41467-018-05914-x}
}

@article{kalinsky2009pik3ca,
  title={PIK3CA mutation associates with improved outcome in breast cancer},
  author={Kalinsky, Kevin and Jacks, Lindsay M and Heguy, Adriana and Patil, Smita and Drobnjak, Maja and Bhanot, Usha K and Hedvat, Craig V and Traina, Tiffany A and Solit, David B and Gerald, William and Moynahan, Mary Ellen},
  journal={Clinical Cancer Research},
  volume={15},
  number={16},
  pages={5049--5059},
  year={2009},
  doi={10.1158/1078-0432.CCR-09-0632}
}

@article{chollet2016breast,
  title={Breast cancer biologic and etiologic heterogeneity by young age and menopausal status in the Carolina Breast Cancer Study: a case-control study},
  author={Chollet-Hinton, Lynn and Anders, Carey K and Tse, Chiu-Kit and Bell, Mary Beth and Yang, Yang Claire and Carey, Lisa A and Olshan, Andrew F and Troester, Melissa A},
  journal={Breast Cancer Research},
  volume={18},
  number={1},
  pages={79},
  year={2016},
  publisher={Springer},
  doi={10.1186/s13058-016-0736-y}
}

@article{hertel2025p53mitotic,
  title={The Role of p53 Mutations in Early and Late Response to Mitotic Aberrations},
  author={Hertel, A and Storchov{\'a}, Z},
  journal={Biomolecules},
  number={2},
  pages={244},
  year={2025},
  doi={10.3390/biom15020244}
}

@article{levine2020p53,
  title={p53: 800 million years of evolution and 40 years of discovery},
  author={Levine, Arnold J},
  journal={Nature Reviews Cancer},
  volume={20},
  number={8},
  pages={471--480},
  year={2020},
  doi={10.1038/s41568-020-0262-1}
}

@article{brancati2022bracs,
  title={Bracs: A dataset for breast carcinoma subtyping in h\&e histology images},
  author={Brancati, Nadia and Anniciello, Anna Maria and Pati, Pushpak and Riccio, Daniel and Scognamiglio, Giosu{\`e} and Jaume, Guillaume and De Pietro, Giuseppe and Di Bonito, Maurizio and Foncubierta, Antonio and Botti, Gerardo and others},
  journal={Database},
  volume={2022},
  pages={baac093},
  year={2022},
  publisher={Oxford University Press UK}
}

@article{polonia2019bach,
  title={BACH dataset: Grand challenge on breast cancer histology images},
  author={Pol{\'o}nia, Ant{\'o}nio and Eloy, Catarina and Aguiar, Paulo},
  journal={Med. Image Anal},
  volume={2019},
  pages={563},
  year={2019}
}

@article{weitz2022acrobat,
  title={ACROBAT--a multi-stain breast cancer histological whole-slide-image data set from routine diagnostics for computational pathology},
  author={Weitz, Philippe and Valkonen, Masi and Solorzano, Leslie and Carr, Circe and Kartasalo, Kimmo and Boissin, Constance and Koivukoski, Sonja and Kuusela, Aino and Rasic, Dusan and Feng, Yanbo and others},
  journal={arXiv preprint arXiv:2211.13621},
  year={2022}
}

@article{xu2021predicting,
    title={Predicting Axillary Lymph Node Metastasis in Early Breast Cancer Using Deep Learning on Primary Tumor Biopsy Slides},
    author={Xu, Feng and Zhu, Chuang and Tang, Wenqi and Wang, Ying and Zhang, Yu and Li, Jie and Jiang, Hongchuan and Shi, Zhongyue and Liu, Jun and Jin, Mulan},
    journal={Frontiers in Oncology},
    pages={4133},
    year={2021},
    publisher={Frontiers}
}

@article{gtex2020gtex,
  title={The GTEx Consortium atlas of genetic regulatory effects across human tissues},
  author={GTEx Consortium},
  journal={Science},
  volume={369},
  number={6509},
  pages={1318--1330},
  year={2020},
  publisher={American Association for the Advancement of Science}
}

@article{edwards2015cptac,
  title={The CPTAC data portal: a resource for cancer proteomics research},
  author={Edwards, Nathan J and Oberti, Mauricio and Thangudu, Ratna R and Cai, Shuang and McGarvey, Peter B and Jacob, Shine and Madhavan, Subha and Ketchum, Karen A},
  journal={Journal of proteome research},
  volume={14},
  number={6},
  pages={2707--2713},
  year={2015},
  publisher={ACS Publications}
}

@article{coates2015tailoring,
  title={Tailoring therapies—improving the management of early breast cancer: St Gallen International Expert Consensus on the Primary Therapy of Early Breast Cancer 2015},
  author={Coates, Alan S and Winer, Eric P and Goldhirsch, Aron and Gelber, Richard D and Gnant, Michael and Piccart-Gebhart, M and Th{\"u}rlimann, Beat and Senn, H-J and Members, Panel and Andr{\'e}, Fabrice and others},
  journal={Annals of oncology},
  volume={26},
  number={8},
  pages={1533--1546},
  year={2015},
  publisher={Elsevier}
}

@inproceedings{dosovitskiyimage,
  title={An Image is Worth 16x16 Words: Transformers for Image Recognition at Scale},
  author={Dosovitskiy, Alexey and Beyer, Lucas and Kolesnikov, Alexander and Weissenborn, Dirk and Zhai, Xiaohua and Unterthiner, Thomas and Dehghani, Mostafa and Minderer, Matthias and Heigold, Georg and Gelly, Sylvain and others},
  booktitle={International Conference on Learning Representations}
}

@inproceedings{ilse2018attention,
  title={Attention-based deep multiple instance learning},
  author={Ilse, Maximilian and Tomczak, Jakub and Welling, Max},
  booktitle={International conference on machine learning},
  pages={2127--2136},
  year={2018},
  organization={PMLR}
}

@article{zadeh2020bias,
  title={Bias in cross-entropy-based training of deep survival networks},
  author={Zadeh, Shekoufeh Gorgi and Schmid, Matthias},
  journal={IEEE transactions on pattern analysis and machine intelligence},
  volume={43},
  number={9},
  pages={3126--3137},
  year={2020},
  publisher={IEEE}
}

@article{ziegler2007survival,
  title={Survival analysis: log rank test},
  author={Ziegler, A and Lange, S and Bender, R},
  journal={Dtsch Med Wochenschr},
  volume={132},
  number={Suppl 1},
  pages={e39--41},
  year={2007}
}

@article{woolson2007wilcoxon,
  title={Wilcoxon signed-rank test},
  author={Woolson, Robert F},
  journal={Wiley encyclopedia of clinical trials},
  pages={1--3},
  year={2007},
  publisher={Wiley Online Library}
}
    
\clearpage
\setcounter{figure}{0}
\setcounter{table}{0}
\section*{Supplementary information}

\begin{figure}
  \centering
  \captionsetup{name=Extended Data Figure}
  \includegraphics[width=1.0\linewidth]{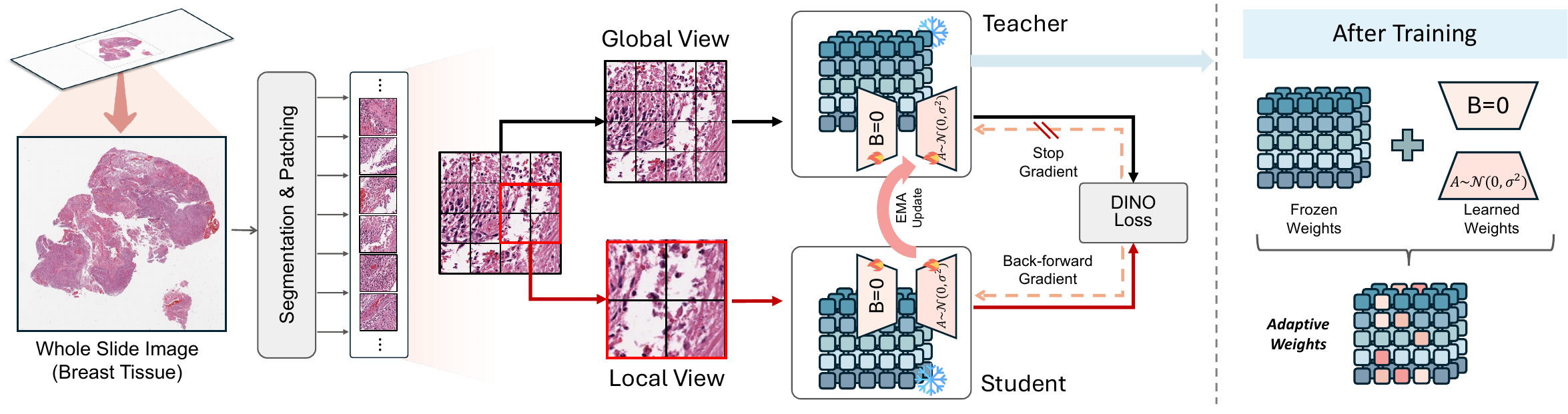}
  \caption{\textbf{Methodological details of BRAVE.} \textbf{a}, Schematic of the breast-adaptive pretraining process. Whole-slide images were segmented and patched to generate breast tissue tiles, from which paired global and local views were sampled for self-supervised training. A pan-cancer foundation model (Virchow2) served as the initialization for both the teacher and student networks, and breast adaptation was performed with a DINO-style objective in which the student was optimized by back-propagation while the teacher was updated by exponential moving average (EMA). During adaptation, the backbone weights remained frozen and only lightweight learnable adaptation weights were updated. After training, the frozen pretrained weights were retained and combined with the learned adaptation weights to form the final breast-adaptive BRAVE parameters.}
  \label{fig:method}
\end{figure}
\begin{table}[htbp]
  \centering
  \captionsetup{name=Extended Data Table}
  \caption{\textbf{Data Details from 32 sources} (21 in-house medical institutions and 11 public sources), including the number of slides and cases, data source, their usage in this work, and tissue type.}
    \begin{tabular}{l|r|r|r|r|r}
    \toprule
    Center & \multicolumn{1}{l|}{Slides} & \multicolumn{1}{l|}{Case} & \multicolumn{1}{l|}{Usage} & \multicolumn{1}{l|}{Tissue Type} & \multicolumn{1}{l}{Geographic} \\
    \midrule
    H1    & 26469 & 2976  & \multicolumn{1}{l|}{Pretrain} & \multicolumn{1}{l|}{Surgical, Biopsy} & \multicolumn{1}{l}{Asia} \\
    H2    & 16939 & 6449  & \multicolumn{1}{l|}{Pretrain,Downstream} & \multicolumn{1}{l|}{Frozen} & \multicolumn{1}{l}{Asia} \\
    H4    & 15629 & 4055  & \multicolumn{1}{l|}{Downstream} & \multicolumn{1}{l|}{Surgical,Frozen,Biopsy} & \multicolumn{1}{l}{Asia} \\
    H3    & 7303  & 7303  & \multicolumn{1}{l|}{Pretrain} & \multicolumn{1}{l|}{Surgical} & \multicolumn{1}{l}{Asia} \\
    H5    & 6305  & 2411  & \multicolumn{1}{l|}{Downstream} & \multicolumn{1}{l|}{Biopsy} & \multicolumn{1}{l}{Asia} \\
    H10   & 4725  & 4546  & \multicolumn{1}{l|}{Downstream} & \multicolumn{1}{l|}{Biopsy,Surgical} & \multicolumn{1}{l}{Asia} \\
    H6    & 4646  & 2219  & \multicolumn{1}{l|}{Downstream} & \multicolumn{1}{l|}{Surgical} & \multicolumn{1}{l}{Asia} \\
    TCGA-BRCA & 2975  & 1399  & \multicolumn{1}{l|}{Pretrain,Downstream} & \multicolumn{1}{l|}{Frozen,Surgical} & \multicolumn{1}{l}{North America} \\
    H7    & 2406  & 2368  & \multicolumn{1}{l|}{Pretrain,Downstream} & \multicolumn{1}{l|}{Biopsy} & \multicolumn{1}{l}{Asia} \\
    HistAI-Breast & 1876  & 1692  & \multicolumn{1}{l|}{Pretrain} & \multicolumn{1}{l|}{Surgical} & \multicolumn{1}{l}{North America} \\
    H20   & 2403  & 1814  & \multicolumn{1}{l|}{Downstream} & \multicolumn{1}{l|}{Biopsy,Surgical} & \multicolumn{1}{l}{Asia} \\
    H17   & 1654  & 904   & \multicolumn{1}{l|}{Downstream} & \multicolumn{1}{l|}{Biopsy} & \multicolumn{1}{l}{Asia} \\
    H18   & 1397  & 1209  & \multicolumn{1}{l|}{Downstream} & \multicolumn{1}{l|}{Surgical} & \multicolumn{1}{l}{Asia} \\
    BCNB  & 1036  & 1036  & \multicolumn{1}{l|}{Pretrain} & \multicolumn{1}{l|}{Biopsy} & \multicolumn{1}{l}{Asia} \\
    GTEx-Breast & 820   & 812   & \multicolumn{1}{l|}{Pretrain} & \multicolumn{1}{l|}{Surgical, Frozen} & \multicolumn{1}{l}{North America} \\
    H9    & 791   & 788   & \multicolumn{1}{l|}{Downstream} & \multicolumn{1}{l|}{Surgical} & \multicolumn{1}{l}{Asia} \\
    H19   & 585   & 461   & \multicolumn{1}{l|}{Downstream} & \multicolumn{1}{l|}{Biopsy} & \multicolumn{1}{l}{Asia} \\
    BRACS & 545   & 189   & \multicolumn{1}{l|}{Pretrain} & \multicolumn{1}{l|}{Biopsy,Surgical} & \multicolumn{1}{l}{Europe} \\
    H11   & 529   & 511   & \multicolumn{1}{l|}{Pretrain} & \multicolumn{1}{l|}{Surgical} & \multicolumn{1}{l}{Asia} \\
    H12   & 475   & 177   & \multicolumn{1}{l|}{Downstream} & \multicolumn{1}{l|}{Frozen} & \multicolumn{1}{l}{Asia} \\
    H13   & 421   & 418   & \multicolumn{1}{l|}{Downstream} & \multicolumn{1}{l|}{Surgical} & \multicolumn{1}{l}{Asia} \\
    H21   & 341   & 328   & \multicolumn{1}{l|}{Downstream} & \multicolumn{1}{l|}{Biopsy} & \multicolumn{1}{l}{Asia} \\
    H8    & 211   & 203   & \multicolumn{1}{l|}{Downstream} & \multicolumn{1}{l|}{Biopsy} & \multicolumn{1}{l}{Asia} \\
    H14   & 205   & 60    & \multicolumn{1}{l|}{Downstream} & \multicolumn{1}{l|}{Surgical} & \multicolumn{1}{l}{Asia} \\
    MIDOG\_2021 & 193   & 193   & \multicolumn{1}{l|}{Pretrain} & \multicolumn{1}{l|}{Surgical} & \multicolumn{1}{l}{Europe} \\
    H15   & 178   & 171   & \multicolumn{1}{l|}{Downstream} & \multicolumn{1}{l|}{Frozen} & \multicolumn{1}{l}{Asia} \\
    ACROBAT2023 & 154   & 55    & \multicolumn{1}{l|}{Pretrain} & \multicolumn{1}{l|}{Surgical} & \multicolumn{1}{l}{Europe} \\
    SLN-Breast & 129   & 129   & \multicolumn{1}{l|}{Pretrain} & \multicolumn{1}{l|}{Surgical} & \multicolumn{1}{l}{North America} \\
    CPTAC-BRCA & 99    & 99    & \multicolumn{1}{l|}{Downstream} & \multicolumn{1}{l|}{Surgical} & \multicolumn{1}{l}{North America} \\
    Post-NAT-BRCA & 96    & 54    & \multicolumn{1}{l|}{Pretrain} & \multicolumn{1}{l|}{Surgical} & \multicolumn{1}{l}{North America} \\
    H16   & 63    & 63    & \multicolumn{1}{l|}{Downstream} & \multicolumn{1}{l|}{Frozen} & \multicolumn{1}{l}{Asia} \\
    BACH  & 40    & 40    & \multicolumn{1}{l|}{Pretrain} & \multicolumn{1}{l|}{Surgical} & \multicolumn{1}{l}{Europe} \\
    \midrule
    Sum   & 101638 & 45132 &       &       &  \\
    \bottomrule
    \end{tabular}%
  \label{tab:data_src}%
\end{table}%

\begin{table}
\captionsetup{name=Extended Data Table}
\caption{\textbf{The public datasets used in this study.} Please note that some datasets may need permission before downloading.}
    \begin{tabular}{l | l}
        \toprule
         Dataset& Link or Source\\
         \midrule
         1. TCGA &\url{https://portal.gdc.cancer.gov/} \\
         2. CPTAC &\url{https://proteomic.datacommons.cancer.gov/pdc/}\\
         3. BCNB &\url{https://bcnb.grand-challenge.org/}\\
         4. HistAI-Breast &\url{https://huggingface.co/datasets/histai/HISTAI-breast} \\
         5. BRACS &\url{https://www.bracs.icar.cnr.it/download/} \\
         6. MIDOG2021 &\url{https://imig.science/midog2021/download-dataset/}\\
         7. ACROBAT2023 &\url{https://acrobat.grand-challenge.org/}\\
         8. BACH &\url{https://zenodo.org/records/3632035}\\
         9. Post-NAT-BRCA &\url{https://www.cancerimagingarchive.net/collection/post-nat-brca/}\\
          \bottomrule
    \end{tabular}

    \label{tab:link_public}
\end{table}%

\begin{table}[htbp]
  \centering
  \captionsetup{name=Extended Data Table}
  \caption{\textbf{Details of Pretraining Data from 15 sources}, including data source, the number of slides and sampled patches, and their tissue type.}
    \begin{tabular}{l|r|r|l|l}
    \toprule
    Center & \# Slides & \# Patches & Tissue Type & Geographic Sources\\
    \midrule
    H1    &            26,469  &               70,693,634  & Surgical, Biopsy & Asia\\
    H2    &            13,800  &                  3,653,172  & Frozen & Asia\\
    H3    &               7,303  &               19,558,368  & Surgical & Asia\\
    TCGA-BRCA &               2,622  &                  5,525,942  & Frozen,Surgical & North America\\
    HistAI-Breast &               1,876  &                  4,321,147  & Surgical & North America\\
    H7    &               1,659  &                  4,730,669  & Biopsy & Asia\\
    BCNB  &               1,036  &                       261,365  & Biopsy & Asia\\
    GTEx-Breast &                    820  &                       582,303  & Surgical, Frozen & North America\\
    BRACS &                    545  &                  1,593,759  & Biopsy,Surgical &  Europe\\
    H11   &                    529  &                  1,540,629  & Surgical & Asia\\
    MIDOG\_2021 &                    193  &                          24,025  & Surgical & Europe\\
    ACROBAT2023 &                    154  &                          77,079  & Surgical & Europe\\
    SLN-Breast &                    129  &                       132,556  & Surgical & North America\\
    Post-NAT-BRCA &                       96  &                       212,680  & Surgical & North America\\
    BACH  &                       40  &                       108,952  & Surgical & Europe\\
    \midrule
    Sum   &            57,271  &            113,016,280  &   &\\
    \bottomrule
    \end{tabular}%
  \label{tab:data_pretrain}%
\end{table}%

\begin{sidewaystable}
  \centering
  \captionsetup{name=Extended Data Table}
  \caption{\textbf{Averaged Performance of Different Models Across Various Stages and Tasks.}}
    \begin{tabular}{l|l|l|l|l|l|l|l}
    \toprule
    \multicolumn{1}{c|}{\textbf{Stage}} & \multicolumn{1}{c|}{\textbf{Type}} & \multicolumn{1}{c|}{\textbf{ShortName}} & \multicolumn{1}{c|}{\textbf{PLIP}} & \multicolumn{1}{c|}{\textbf{CONCH}} & \multicolumn{1}{c|}{\textbf{UNI}} & \multicolumn{1}{c|}{\textbf{Virchow2}} & \multicolumn{1}{c}{\textbf{BRAVE}} \\
    \midrule
    Pre-operative & Differential Diagnosis & Pre-\newline{}Diagnosis & 0.940 $\pm$ 0.085 & 0.940 $\pm$ 0.071 & 0.977 $\pm$ 0.033 & 0.951 $\pm$ 0.069 & 0.980 $\pm$ 0.028 \\
    Pre-operative & Treatment Decision & Pre-\newline{}Treatment & 0.726 $\pm$ 0.082 & 0.781 $\pm$ 0.059 & 0.797 $\pm$ 0.059 & 0.811 $\pm$ 0.052 & 0.822 $\pm$ 0.047 \\
    Pre-operative & Survival Prediction & Pre-\newline{}Survival & 0.655 $\pm$ 0.068 & 0.694 $\pm$ 0.050 & 0.684 $\pm$ 0.053 & 0.697 $\pm$ 0.045 & 0.705 $\pm$ 0.047 \\
    \midrule
    Intra-operative & Differential Diagnosis & Intra-\newline{}Diagnosis & 0.813 $\pm$ 0.129 & 0.921 $\pm$ 0.073 & 0.958 $\pm$ 0.031 & 0.952 $\pm$ 0.037 & 0.964 $\pm$ 0.036 \\
    Intra-operative & Surgical Decision & Intra-\newline{}SurgDec & 0.896 $\pm$ 0.049 & 0.929 $\pm$ 0.025 & 0.918 $\pm$ 0.047 & 0.916 $\pm$ 0.045 & 0.937 $\pm$ 0.027 \\
    \midrule
    Post-operative & Morphology Assessment & Post-\newline{}Morphology & 0.759 $\pm$ 0.088 & 0.887 $\pm$ 0.086 & 0.881 $\pm$ 0.089 & 0.898 $\pm$ 0.078 & 0.920 $\pm$ 0.069 \\
    Post-operative & Tumor Behavior & Post-\newline{}Behavior & 0.733 $\pm$ 0.061 & 0.770 $\pm$ 0.018 & 0.762 $\pm$ 0.019 & 0.750 $\pm$ 0.038 & 0.794 $\pm$ 0.039 \\
    Post-operative & Molecular Inference & Post-\newline{}Molecular & 0.761 $\pm$ 0.090 & 0.779 $\pm$ 0.099 & 0.796 $\pm$ 0.091 & 0.809 $\pm$ 0.091 & 0.830 $\pm$ 0.075 \\
    Post-operative & Survival Prediction & Post-\newline{}Survival & 0.629 $\pm$ 0.077 & 0.680 $\pm$ 0.067 & 0.682 $\pm$ 0.069 & 0.677 $\pm$ 0.073 & 0.696 $\pm$ 0.072 \\
    \bottomrule
    \end{tabular}%
  \label{tab:performance_radar}%
\end{sidewaystable}%

\clearpage
\begin{sidewaystable}
  \centering
  \captionsetup{name=Extended Data Table}
  \caption{\textbf{Center Distribution of Tasks at Pre-operative Stage}.}
    \begin{tabular}{l|l|l}
    \toprule
    \multicolumn{1}{c|}{\textbf{Stage}} & \multicolumn{1}{c|}{\textbf{Task}} & \multicolumn{1}{c}{\textbf{Center Distribution (Case)}} \\
    \midrule
    Pre-operative & Malignancy Detection (biopsy) & Train: 586; Val: 84; Test: 168 \\
    \midrule
    Pre-operative & Mass Differentiation (biopsy) & Train: 586; Val: 84; Test: 168 \\
    \midrule
    Pre-operative & Metastasis Detection (biopsy) & Train: 179; Val: 26; Test: 52 \\
    \midrule
    Pre-operative & ER Prediction (biopsy) & Train: 1039; Val: 149; Test: 297; H10: 703; H17: 690; H19: 455; H20: 1159; H21: 180; H4: 124; H7: 569; H8: 211 \\
    \midrule
    Pre-operative & PR Prediction (biopsy) & Train: 1037; Val: 149; Test: 297; H10: 622; H17: 665; H4: 123; H8: 211 \\
    \midrule
    Pre-operative & HER2 Prediction (biopsy) & Train: 931; Val: 133; Test: 267; H10: 622; H4: 91 \\
    \midrule
    Pre-operative & KI67 Prediction (biopsy) & Train: 1036; Val: 148; Test: 297; H17: 865; H4: 123 \\
    \midrule
    Pre-operative & Molecular Subtyping (biopsy) & Train: 980; Val: 140; Test: 280; H4: 52 \\
    \midrule
    Pre-operative & NAC Response Prediction (biopsy) & Train: 1073; Val: 154; Test: 307 \\
    \bottomrule
    \end{tabular}%
  \label{tab:center_distr_pre}%
\end{sidewaystable}%

\begin{table}[htbp]
  \centering
  \captionsetup{name=Extended Data Table}
  \caption{\textbf{Class Distribution on 26 Retrospective Cohorts from 9 Centers for Pre-operative Stage}.}
  \scalebox{0.9}{  
  \begin{tabular}{l|l|l|l|r|l}
    \toprule
    \multicolumn{1}{c|}{\textbf{Stage}} & \multicolumn{1}{c|}{\textbf{Task}} & \multicolumn{1}{c|}{\textbf{Center}} & \multicolumn{1}{c|}{\textbf{Setting}} & \multicolumn{1}{c|}{\textbf{\# Cases}} & \multicolumn{1}{c}{\textbf{Class}} \\
    \midrule
    Pre-operative & Malignancy Detection & H4    & Internal & 838   & non-malignant:705; malignant:133 \\
    \midrule
    Pre-operative & Mass Differentiation & H4    & Internal & 838   & fibro:415; udh:290; idc:133 \\
    \midrule
    Pre-operative & Metastasis Detection & H5    & Internal & 257   & n0+:224; n0:33 \\
    \midrule
    Pre-operative & ER Prediction & H5    & Internal & 1485  & positive:1011; negative:474 \\
    Pre-operative & ER Prediction & H10   & Retrospective & 703   & positive:529; negative:174 \\
    Pre-operative & ER Prediction & H17   & Retrospective & 690   & positive:510; negative:180 \\
    Pre-operative & ER Prediction & H19   & Retrospective & 455   & positive:312; negative:143 \\
    Pre-operative & ER Prediction & H20   & Retrospective & 1159  & positive:858; negative:301 \\
    Pre-operative & ER Prediction & H21   & Retrospective & 180   & positive:141; negative:39 \\
    Pre-operative & ER Prediction & H4    & Retrospective & 124   & positive:89; negative:35 \\
    Pre-operative & ER Prediction & H7    & Retrospective & 569   & positive:448; negative:121 \\
    Pre-operative & ER Prediction & H8    & Retrospective & 211   & positive:119; negative:92 \\
    \midrule
    Pre-operative & PR Prediction & H5    & Internal & 1483  & positive:1119; negative:364 \\
    Pre-operative & PR Prediction & H10   & Retrospective & 622   & positive:417; negative:205 \\
    Pre-operative & PR Prediction & H17   & Retrospective & 665   & positive:425; negative:240 \\
    Pre-operative & PR Prediction & H4    & Retrospective & 123   & positive:70; negative:53 \\
    Pre-operative & PR Prediction & H8    & Retrospective & 211   & negative:129; positive:82 \\
    \midrule
    Pre-operative & HER2 Prediction & H5    & Internal & 1331  & low:857; high:474 \\
    Pre-operative & HER2 Prediction & H10   & Retrospective & 622   & low:396; high:226 \\
    Pre-operative & HER2 Prediction & H4    & Retrospective & 91    & low:70; high:21 \\
    \midrule
    Pre-operative & KI67 Prediction & H5    & Internal & 1481  & high:1171; low:310 \\
    Pre-operative & KI67 Prediction & H17   & Retrospective & 865   & high:756; low:109 \\
    Pre-operative & KI67 Prediction & H4    & Retrospective & 123   & high:109; low:14 \\
    \midrule
    Pre-operative & Molecular Subtyping & H5    & Internal & 1400  & luminal b:972; luminal a:172; her2:131; tnbc:125 \\
    Pre-operative & Molecular Subtyping & H4    & Retrospective & 52    & luminal b:20; tnbc:14; her2:11; luminal a:7 \\
    \midrule
    Pre-operative & NAC Response Prediction & H5    & Internal & 1534  & non-pcr:1181; pcr:353 \\
    \bottomrule
    \end{tabular}%
  }
  \label{tab:cls_distr_pre}%
\end{table}%

\begin{table}[htbp]
  \centering
  \captionsetup{name=Extended Data Table}
  \caption{Thresholds for IHC markers to determine Positive/High Expression.}
    \begin{tabular}{l|l}
    \toprule
    IHC Marker & Threshold for Positive/High Expression \\
    \midrule
    ER    & >10\% \\
    \midrule
    PR    & >10\% \\
    \midrule
    HER2  & 3+ / 2+\&FISH+ \\
    \midrule
    KI67  & >14\% \\
    \midrule
    CK5   & >1\% (any positive staining) \\
    \midrule
    AR    & >1\% (any nuclear staining) \\
    \bottomrule
    \end{tabular}%
  \label{tab:threshold_ihc}%
\end{table}%

\begin{table}
  \centering
  \captionsetup{name=Extended Data Table}
  \caption{\textbf{Performance of BRAVE  for Pre-operative Stage on 26 retrospective cohorts from 9 centers}, where `Internal' represents the result of the internal test set. Sensitivity, Specificity, and NPV are reported at the maximum Youden's Index. 95\% CI is included in parentheses.}
   \scalebox{0.75}{
    \begin{tabular}{l|c|c|c|c|c|c}
    \toprule
    \multicolumn{1}{c|}{\textbf{Task}} & \textbf{Center} & \textbf{Type} & \textbf{Macro-AUC} & \textbf{Sensitivity} & \textbf{Specificity} & \textbf{NPV} \\
    \midrule
    Malignancy Detection (Bio) & H4 (Internal) & Diagnosis & 1.000 (1.000-1.000) & 1.000 (1.000-1.000) & 1.000 (1.000-1.000) & 1.000 (1.000-1.000) \\
    \midrule
    Mass Differentiation (Bio) & H4 (Internal) & Diagnosis & 1.000 (0.998-1.000) & 1.000 (1.000-1.000) & 0.996 (0.979-1.000) & 1.000 (1.000-1.000) \\
    \midrule
    Metastasis Detection (Bio) & H5 (Internal) & Diagnosis & 0.940 (0.820-1.000) & 0.887 (0.636-1.000) & 0.931 (0.714-1.000) & 0.679 (0.300-1.000) \\
    \midrule
    ER Prediction (Bio) & H5 (Internal) & Decision & 0.871 (0.826-0.912) & 0.778 (0.707-0.887) & 0.914 (0.795-0.971) & 0.663 (0.595-0.777) \\
    ER Prediction (Bio) & H10-Retro & Decision & 0.840 (0.805-0.874) & 0.774 (0.719-0.819) & 0.838 (0.777-0.896) & 0.551 (0.498-0.603) \\
    ER Prediction (Bio) & H17-Retro & Decision & 0.829 (0.798-0.862) & 0.715 (0.625-0.779) & 0.824 (0.746-0.899) & 0.507 (0.448-0.563) \\
    ER Prediction (Bio) & H19-Retro & Decision & 0.801 (0.760-0.841) & 0.690 (0.599-0.795) & 0.839 (0.726-0.921) & 0.557 (0.502-0.630) \\
    ER Prediction (Bio) & H20-Retro & Decision & 0.835 (0.810-0.860) & 0.768 (0.667-0.824) & 0.790 (0.717-0.879) & 0.548 (0.474-0.599) \\
    ER Prediction (Bio) & H21-Retro & Decision & 0.840 (0.772-0.900) & 0.731 (0.586-0.855) & 0.873 (0.727-1.000) & 0.481 (0.382-0.610) \\
    ER Prediction (Bio) & H4-Retro & Decision & 0.843 (0.764-0.910) & 0.752 (0.602-0.878) & 0.851 (0.707-0.973) & 0.583 (0.469-0.722) \\
    ER Prediction (Bio) & H7-Retro & Decision & 0.789 (0.749-0.826) & 0.631 (0.532-0.782) & 0.867 (0.701-0.958) & 0.393 (0.343-0.480) \\
    ER Prediction (Bio) & H8-Retro & Decision & 0.894 (0.847-0.935) & 0.808 (0.686-0.932) & 0.861 (0.721-0.965) & 0.783 (0.695-0.898) \\
    \midrule
    PR Prediction (Bio) & H5 (Internal) & Decision & 0.773 (0.709-0.829) & 0.671 (0.554-0.835) & 0.800 (0.610-0.910) & 0.449 (0.381-0.570) \\
    PR Prediction (Bio) & H10-Retro & Decision & 0.823 (0.789-0.854) & 0.715 (0.651-0.795) & 0.825 (0.738-0.889) & 0.589 (0.545-0.648) \\
    PR Prediction (Bio) & H17-Retro & Decision & 0.762 (0.725-0.798) & 0.723 (0.573-0.808) & 0.719 (0.624-0.858) & 0.600 (0.522-0.669) \\
    PR Prediction (Bio) & H4-Retro & Decision & 0.811 (0.735-0.887) & 0.833 (0.632-0.961) & 0.726 (0.542-0.902) & 0.782 (0.636-0.926) \\
    PR Prediction (Bio) & H8-Retro & Decision & 0.802 (0.738-0.859) & 0.846 (0.630-0.954) & 0.666 (0.525-0.879) & 0.877 (0.787-0.954) \\
    \midrule
    HER2 Prediction (Bio) & H5 (Internal) & Decision & 0.824 (0.771-0.872) & 0.743 (0.638-0.876) & 0.831 (0.674-0.901) & 0.855 (0.809-0.913) \\
    HER2 Prediction (Bio) & H10-Retro & Decision & 0.791 (0.755-0.827) & 0.715 (0.606-0.857) & 0.742 (0.591-0.841) & 0.823 (0.779-0.883) \\
    HER2 Prediction (Bio) & H4-Retro & Decision & 0.801 (0.676-0.917) & 0.757 (0.524-1.000) & 0.801 (0.591-0.971) & 0.920 (0.861-1.000) \\
    \midrule
    KI67 Prediction (Bio) & H5 (Internal) & Decision & 0.816 (0.753-0.873) & 0.733 (0.647-0.814) & 0.851 (0.736-0.943) & 0.459 (0.392-0.537) \\
    KI67 Prediction (Bio) & H17-Retro & Decision & 0.765 (0.723-0.807) & 0.628 (0.539-0.743) & 0.821 (0.686-0.912) & 0.243 (0.214-0.289) \\
    KI67 Prediction (Bio) & H4-Retro & Decision & 0.843 (0.739-0.922) & 0.705 (0.505-0.926) & 0.926 (0.714-1.000) & 0.312 (0.206-0.557) \\
    \midrule
    Molecular Subtyping (Bio) & H5 (Internal) & Decision & 0.841 (0.756-0.912) & 0.835 (0.692-0.949) & 0.806 (0.709-0.871) & 0.972 (0.950-0.991) \\
    Molecular Subtyping (Bio) & H4-Retro & Decision & 0.967 (0.910-1.000) & 0.999 (1.000-1.000) & 0.928 (0.833-1.000) & 1.000 (1.000-1.000) \\
    \midrule
    NAC Response Prediction (Bio) & H5 (Internal) & Decision & 0.740 (0.671-0.805) & 0.688 (0.453-0.926) & 0.711 (0.439-0.893) & 0.890 (0.842-0.957) \\
    \bottomrule
    \end{tabular}%
    }
  \label{tab:result_pre}%
\end{table}

\begin{table}[htbp]
  \centering
  \captionsetup{name=Extended Data Table}
  \caption{\textbf{Comparison between Challenging Subgroup and Overall Cohorts at Pre-operative Stage}. KS Statistic is obtained by Kolmogorov-Smirnov (KS) Test. 95\% CI is included in parentheses.}
  \scalebox{0.75}{
    \begin{tabular}{c|c|c|c|l|c|c|c|c}
    \toprule
    \multicolumn{1}{l|}{\textbf{Baseline Cohort}} & \multicolumn{1}{l|}{\textbf{Subgroup}} & \textbf{AUC (baseline)} & \textbf{AUC (subgroup)} & \textbf{Class} & \textbf{Baseline N} & \textbf{Subgroup N} & \textbf{KS Statistic} & \textbf{P-Value} \\
    \midrule
    \multirow{4}[8]{*}{ER-Bio-All (H20)} & \multirow{2}[4]{*}{Need Extra Info (H20)} & \multirow{4}[8]{*}{0.835 (0.810-0.860)} & \multirow{2}[4]{*}{0.79 (0.668-0.888)} & Negative & 301   & 22    & 0.2543  & 0.1171  \\
\cmidrule{5-9}          &       &       &       & Positive & 858   & 42    & 0.1510  & 0.2899  \\
\cmidrule{2-2}\cmidrule{4-9}          & \multirow{2}[4]{*}{Insufficient Tissue (H20)} &       & \multirow{2}[4]{*}{0.841 (0.724-0.933)} & Negative & 301   & 18    & 0.1591  & 0.7226  \\
\cmidrule{5-9}          &       &       &       & Positive & 858   & 43    & 0.1385  & 0.3762  \\
    \midrule
    \multirow{4}[8]{*}{ER-Bio-All (H17)} & \multirow{2}[4]{*}{Need Extra Info (H17)} & \multirow{4}[8]{*}{0.829 (0.798-0.862)} & \multirow{2}[4]{*}{0.828 (0.757-0.893)} & Negative & 180   & 43    & 0.1346  & 0.5058  \\
\cmidrule{5-9}          &       &       &       & Positive & 510   & 121   & 0.0750  & 0.6099  \\
\cmidrule{2-2}\cmidrule{4-9}          & \multirow{2}[4]{*}{Insufficient Tissue (H17)} &       & \multirow{2}[4]{*}{0.833 (0.567-1.000)} & Negative & 180   & 4     & 0.3722  & 0.5483  \\
\cmidrule{5-9}          &       &       &       & Positive & 510   & 9     & 0.2484  & 0.5653  \\
    \bottomrule
    \end{tabular}%
}
  \label{tab:ks_stat_pre}%
\end{table}%

\begin{table}[htbp]
  \centering
  \captionsetup{name=Extended Data Table}
  \caption{\textbf{Performance Comparison (Balanced Accuracy)} between Biopsy (from pathologists) and BRAVE in Predicting Post-operative Surgical IHC Status from Pre-operative Biopsies. Pathologists made decisions based on the standard clinical workflow using both H\&E and IHC biopsies, while BRAVE used H\&E biopsies only. 95\% CI is included in parentheses.}
    \begin{tabular}{ccc|c|c}
    \toprule
    \multicolumn{1}{c|}{\textbf{Task}} & \multicolumn{1}{c|}{\textbf{Center}} & \textbf{N (Case)} & \textbf{Biopsy} & \textbf{BRAVE} \\
    \midrule
    \multicolumn{1}{l|}{ER (Bio)} & \multicolumn{1}{c|}{H21} & 180   & 0.804 (0.729, 0.881) & 0.792 (0.723, 0.853) \\
    \midrule
    \multicolumn{1}{l|}{ER (Bio)} & \multicolumn{1}{c|}{H10} & 703   & 0.789 (0.752, 0.823) & 0.804 (0.771, 0.837) \\
    \midrule
    \multicolumn{1}{l|}{PR (Bio)} & \multicolumn{1}{c|}{H10} & 622   & 0.754 (0.718, 0.787) & 0.767 (0.734, 0.799) \\
    \midrule
    \multicolumn{1}{l|}{HER2 (Bio)} & \multicolumn{1}{c|}{H10} & 622   & 0.524 (0.482, 0.565) & 0.722 (0.685, 0.758) \\
    \midrule
    \multicolumn{3}{c|}{Overall} & 0.718$\pm$0.113 & 0.771$\pm$0.031 \\
    \bottomrule
    \end{tabular}%
  \label{tab:res_cmp_human_ai}%
\end{table}%

\clearpage

\begin{figure}
    \centering
    \captionsetup{name=Extended Data Figure}
    \includegraphics[width=1.0\linewidth]{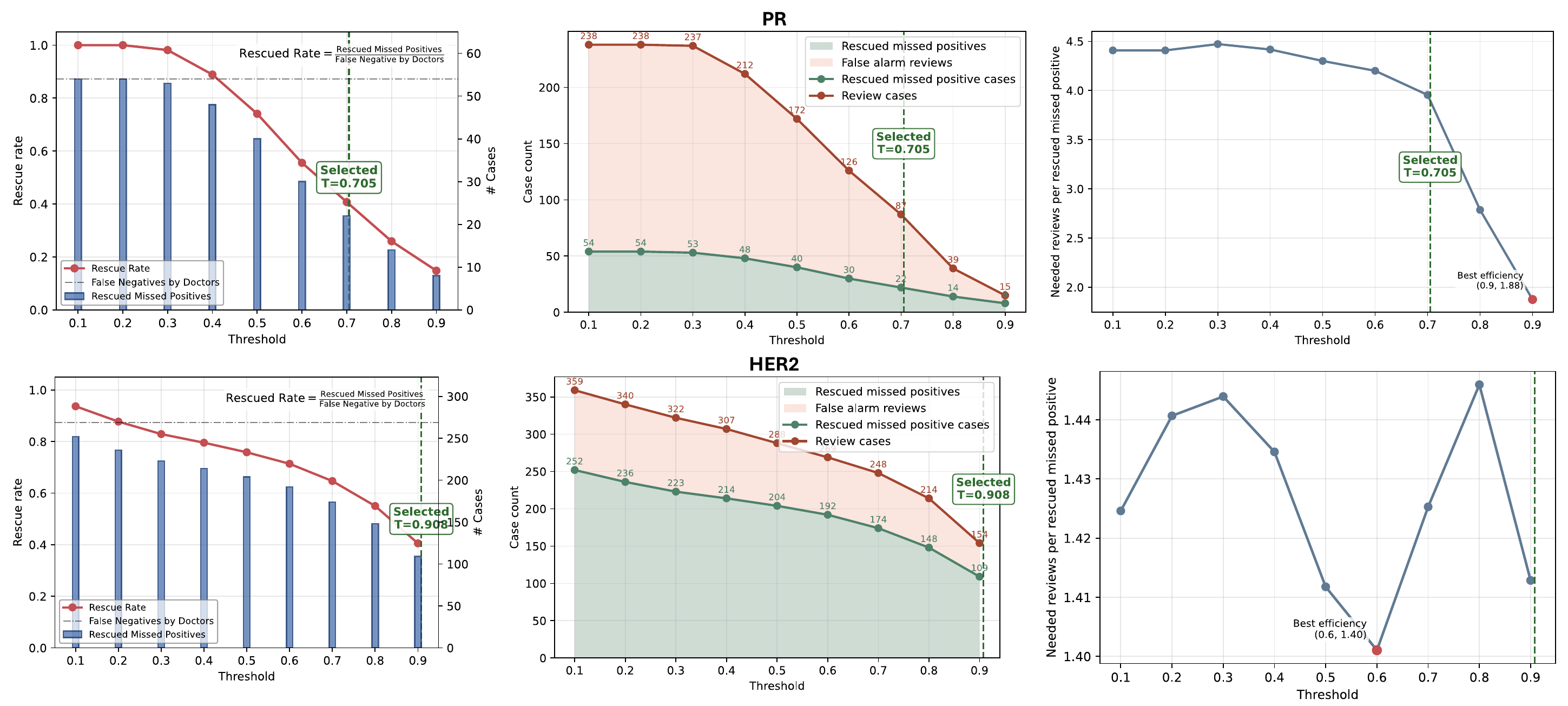}
  \caption{\textbf{Threshold selection and second-review cost analysis of PR and HER2 prediction on pre-operative biopsies.} 
  The three panels show the rescue rate of missed positives under second review (left), 
  the numbers of review cases, false-alarm reviews, and rescued missed positives (middle), 
  and the number of reviews required per rescued missed positive case (right) across candidate thresholds. 
  These analyses were used to balance rescue benefit against second-review workload and to select the operating threshold.}
    \label{fig:pr_her2_prevspost_analysis}
\end{figure}

\begin{table}[htbp]
  \centering
  \captionsetup{name=Extended Data Table}
  \caption{\textbf{Rescue Analysis for IHC (ER) Status Differentiation with AI assistance}, involving two cohorts: H10 and H21.}
  \scalebox{0.8}{
    \begin{tabular}{c|c|c|c|c|c|c}
    \toprule
    \textbf{Total FN by Doctors} & \textbf{Threshold} & \textbf{Rescued FN} & \textbf{Rescue Rate} & \textbf{Review Cases} & \textbf{False Alarm Reviews} & \textbf{Needed Reviews per Rescued FN} \\
    \midrule
    \multirow{9}[18]{*}{50} & 0.1   & 47    & 0.94  & 189   & 142   & 4.02 \\
\cmidrule{2-7}          & 0.2   & 38    & 0.76  & 147   & 109   & 3.87 \\
\cmidrule{2-7}          & 0.3   & 30    & 0.60  & 109   & 79    & 3.63 \\
\cmidrule{2-7}          & 0.4   & 25    & 0.50  & 90    & 65    & 3.60 \\
\cmidrule{2-7}          & 0.5   & 20    & 0.40  & 71    & 51    & 3.55 \\
\cmidrule{2-7}          & 0.6   & 11    & 0.22  & 50    & 39    & 4.55 \\
\cmidrule{2-7}          & 0.7   & 10    & 0.20  & 36    & 26    & 3.60 \\
\cmidrule{2-7}          & 0.8   & 10    & 0.20  & 32    & 22    & 3.20 \\
\cmidrule{2-7}          & 0.9   & 9     & 0.18  & 20    & 11    & 2.22 \\
    \bottomrule
    \end{tabular}%
    }
  \label{tab:rescue_analysis}%
\end{table}%

\begin{table}[htbp]
  \centering
  \captionsetup{name=Extended Data Table}
  \caption{\textbf{Threshold selection for second-review cost analysis of ER, PR and HER2 on pre-operative biopsies.} 
  Rescue rate denotes the proportion of doctor-missed true-positive cases that could be rescued by model-triggered second review. 
  Review burden denotes the proportion of cases initially judged negative by doctors that would be sent for second review. 
  Number needed to review denotes the average number of second-review cases required to rescue one doctor-missed positive case.
  Rescued\_fn denotes the number of doctor-missed positive cases that could be rescued by second review at the selected threshold.
  Review\_cases denotes the number of cases that would be sent for second review at the selected threshold.
  }
    \begin{tabular}{l|c|c|c}
\midrule    \textbf{Task} & \textbf{ER} & \textbf{PR} & \textbf{HER2} \\
    \midrule
    \textbf{Center} & H10+H21 & H10   & H10 \\
    \midrule
    \textbf{selected\_threshold} & 0.479 & 0.705 & 0.908 \\
    \midrule
    \textbf{constraints} & \multicolumn{3}{c}{max\_review\_burden': 0.4, 'min\_rescue\_rate': 0.4} \\
    \midrule
    \textbf{rescue\_rate} & 0.440  & 0.333  & 0.405  \\
    \midrule
    \textbf{review\_burden} & 0.357  & 0.227  & 0.395  \\
    \midrule
    \textbf{number\_needed\_to\_review} & 3.4   & 3.0   & 1.4  \\
    \midrule
    \textbf{rescued\_fn} & 22    & 18    & 109 \\
    \midrule
    \textbf{review\_cases} & 74    & 54    & 153 \\
    \bottomrule
    \end{tabular}%
  \label{tab:threshold_er_pr}%
\end{table}%

\begin{table}[htbp]
  \centering
  \captionsetup{name=Extended Data Table}
  \caption{\textbf{Threshold selection for second-review cost analysis of KI67 on pre-operative biopsies.} Sensitivity is reported at the selected threshold. 95\% CI is included in parentheses.}
    \begin{tabular}{l|c}
    \toprule
    \textbf{Task} & \textbf{Pre-KI67} \\
    \midrule
    \textbf{N} & 297 \\
    \midrule
    \textbf{Constraints} & Sensitivity >= 0.98 \\
    \midrule
    \textbf{Selected\_threshold} & 0.425 \\
    \midrule
    \textbf{Sensitivity} & 0.983  \\
    \bottomrule
    \end{tabular}%
  \label{tab:threshold_ki67}%
\end{table}%

\begin{table}[htbp]
  \captionsetup{name=Extended Data Table}
  \caption{\textbf{Class Distribution on 11 Retrospective Cohorts from 5 Centers for Intra-operative Stage}.}
  \scalebox{0.75}{
    \begin{tabular}{l|l|l|l|r|l}
    \toprule
    \multicolumn{1}{c|}{\textbf{Stage}} & \multicolumn{1}{c|}{\textbf{Task}} & \multicolumn{1}{c|}{\textbf{Center}} & \multicolumn{1}{c|}{\textbf{Setting}} & \multicolumn{1}{c|}{\textbf{\# Cases}} & \multicolumn{1}{c}{\textbf{Class (cases)}} \\
    \midrule
    Intra-operative & Malignancy Detection & H2    & Internal & 502   & non-malignant:344; malignant:158 \\
    Intra-operative & Malignancy Detection & H12+H16 & Retrospective & 181   & non-malignant:156; malignant:25 \\
    Intra-operative & Malignancy Detection & H15   & Retrospective & 173   & non-malignant:130; malignant:43 \\
    Intra-operative & Malignancy Detection & H4    & Retrospective & 1640  & non-malignant:1073; malignant:567 \\
    \midrule
    Intra-operative & Mass Differentiation & H4    & Internal & 1256  & fibroadenoma:571; invasive carcinoma:413; other benign lesion:272 \\
    \midrule
    Intra-operative & Margin Assessment & H2    & Internal & 478   & negative:346; positive:132 \\
    Intra-operative & Margin Assessment & H12   & Retrospective & 194   & negative:168; positive:26 \\
    Intra-operative & Margin Assessment & H4    & Retrospective & 284   & negative:247; positive:37 \\
    \midrule
    Intra-operative & Lymph Node Metastasis Prediction & H2    & Internal & 499   & negative:298; positive:201 \\
    Intra-operative & Lymph Node Metastasis Prediction & H12   & Retrospective & 194   & negative:153; positive:41 \\
    Intra-operative & Lymph Node Metastasis Prediction & H4    & Retrospective & 299   & negative:236; positive:63 \\
    \bottomrule
    \end{tabular}%
  }
  \label{tab:cls_distr_intra}%
\end{table}%

\begin{table}
  \centering
  \captionsetup{name=Extended Data Table}
  \caption{\textbf{Center Distribution of Tasks at Intra-operative Stage}.}
  \scalebox{0.9}{  
    \begin{tabular}{l|l|l}
    \toprule
    \multicolumn{1}{c|}{\textbf{Stage}} & \multicolumn{1}{c|}{\textbf{Task}} & \multicolumn{1}{c}{\textbf{Center Distribution (Case)}} \\
    \midrule
    Intra-operative & Malignancy Detection (frozen) & Train: 350; Val: 51; Test: 101; H4: 1640; H12+H16: 181; H15: 173 \\
    \midrule
    Intra-operative & Mass Differentiation (frozen) & Train: 878; Val: 126; Test: 252 \\
    \midrule
    Intra-operative & Margin Assessment (frozen) & Train: 334; Val: 48; Test: 96; H4: 284; H12: 194 \\
    \midrule
    Intra-operative & Lymph Node Metastasis Prediction (frozen) & Train: 349; Val: 50; Test: 100; H4: 299; H12: 194 \\
    \bottomrule
    \end{tabular}%
  }
  \label{tab:center_distr_intra}%
\end{table}%

\begin{table}[htbp]
  \centering
  \captionsetup{name=Extended Data Table}
  \caption{\textbf{Performance of BRAVE  for Intra-operative Stage on 11 retrospective cohorts from 5 centers}, where `Internal' represents the result of the internal test set. Sensitivity, Specificity, and NPV are reported at the maximum Youden's Index. 95\% CI is included in parentheses.}
   \scalebox{0.7}{
    \begin{tabular}{l|l|l|l|l|l|l}
    \toprule
    \multicolumn{1}{c|}{\textbf{Task}} & \multicolumn{1}{c|}{\textbf{Center}} & \multicolumn{1}{c|}{\textbf{Type}} & \multicolumn{1}{c|}{\textbf{Macro-AUC}} & \multicolumn{1}{c|}{\textbf{Sensitivity}} & \multicolumn{1}{c|}{\textbf{Specificity}} & \multicolumn{1}{c}{\textbf{NPV}} \\
    \midrule
    Malignancy Detection (Frozen) & H2 (Internal) & Diagnosis & 0.999 (0.996-1.000) & 0.989 (0.958-1.000) & 0.999 (0.971-1.000) & 0.978 (0.918-1.000) \\
    Malignancy Detection (Frozen) & H12+H16-Retro & Diagnosis & 0.936 (0.879-0.978) & 0.882 (0.750-0.968) & 0.908 (0.773-1.000) & 0.584 (0.380-0.808) \\
    Malignancy Detection (Frozen) & H15-Retro & Diagnosis & 0.908 (0.842-0.965) & 0.964 (0.902-0.993) & 0.821 (0.696-0.938) & 0.888 (0.740-0.975) \\
    Malignancy Detection (Frozen) & H4-Retro & Diagnosis & 0.985 (0.980-0.990) & 0.937 (0.916-0.962) & 0.967 (0.940-0.986) & 0.891 (0.860-0.930) \\
    \midrule
    Mass Differentiation (Frozen) & H4 (Internal) & Diagnosis & 0.994 (0.986-0.999) & 0.965 (0.911-1.000) & 0.972 (0.921-1.000) & 0.983 (0.958-1.000) \\
    \midrule
    Margin Assessment (Frozen) & H2 (Internal) & Decision & 0.927 (0.862-0.981) & 0.818 (0.643-0.971) & 0.929 (0.767-1.000) & 0.931 (0.873-0.987) \\
    Margin Assessment (Frozen) & H12-Retro & Decision & 0.973 (0.948-0.990) & 0.967 (0.880-1.000) & 0.893 (0.789-0.976) & 0.995 (0.981-1.000) \\
    Margin Assessment (Frozen) & H4-Retro & Decision & 0.909 (0.831-0.969) & 0.812 (0.683-0.932) & 0.989 (0.976-1.000) & 0.972 (0.954-0.989) \\
    \midrule
    Lymph Node Metastasis Prediction (Frozen) & H2 (Internal) & Decision & 0.976 (0.923-1.000) & 0.976 (0.923-1.000) & 1.000 (1.000-1.000) & 0.985 (0.951-1.000) \\
    Lymph Node Metastasis Prediction (Frozen) & H12-Retro & Decision & 0.920 (0.844-0.980) & 0.860 (0.745-0.969) & 0.966 (0.887-0.994) & 0.963 (0.935-0.991) \\
    Lymph Node Metastasis Prediction (Frozen) & H4-Retro & Decision & 0.919 (0.869-0.959) & 0.872 (0.787-0.947) & 0.929 (0.890-0.965) & 0.965 (0.942-0.985) \\
    \bottomrule
    \end{tabular}%
   }
  \label{tab:result_intra}%
\end{table}%

\begin{table}[htbp]
  \centering
  \captionsetup{name=Extended Data Table}
  \caption{\textbf{Comparison between Challenging Subgroup and Overall Cohorts at Intra-operative Stage}. KS Statistic is obtained by Kolmogorov-Smirnov (KS) Test. 95\% CI is included in parentheses.}
  \scalebox{0.65}{
    \begin{tabular}{c|c|c|c|c|l|c|c|c|c}
    \toprule
    \textbf{Task} & \textbf{Baseline Cohort} & \textbf{Subgroup} & \textbf{AUC (baseline)} & \textbf{AUC (subgroup)} & \multicolumn{1}{c|}{\textbf{Class}} & \textbf{Baseline N} & \textbf{Subgroup N} & \textbf{KS Statistic} & \textbf{P-Value} \\
    \midrule
    \multirow{2}[4]{*}{Malignancy (Frozen)} & \multirow{2}[4]{*}{Overall (H2)} & \multirow{2}[4]{*}{NeedIHC (H2)} & \multirow{2}[4]{*}{0.998 (0.995-1.000)} & \multirow{2}[4]{*}{1.000 (1.000-1.000)} & Positive & 48    & 5     & 0.3208  & 0.6321  \\
\cmidrule{6-10}          &       &       &       &       & Negative & 104   & 11    & 0.4047  & 0.0557  \\
    \midrule
    \multirow{3}[6]{*}{MarginAssess (Frozen)} & \multirow{3}[6]{*}{Overall (H12)} & \multirow{2}[4]{*}{NeedIHC (H12)} & \multirow{3}[6]{*}{0.973 (0.948-0.990)} & \multirow{2}[4]{*}{1.000 (1.000-1.000)} & Negative & 168   & 23    & 0.1488  & 0.7016  \\
\cmidrule{6-10}          &       &       &       &       & Positive & 26    & 14    & 0.2582  & 0.4851  \\
\cmidrule{3-3}\cmidrule{5-10}          &       & NAC (H12) &       & 1.000 (1.000-1.000) & Negative & 168   & 11    & 0.3496  & 0.1238  \\
    \midrule
    \multirow{2}[4]{*}{LN\_Metastasis (Frozen)} & \multirow{2}[4]{*}{Overall (H2+H4)} & \multirow{2}[4]{*}{Deformation (H2+H4)} & \multirow{2}[4]{*}{0.955 (0.927-0.977)} & \multirow{2}[4]{*}{0.913 (0.667-1.000)} & Negative & 326   & 5     & 0.5202  & 0.0922  \\
\cmidrule{6-10}          &       &       &       &       & Positive & 123   & 7     & 0.1324  & 0.9985  \\
    \midrule
    \multirow{2}[4]{*}{Malignancy (Frozen)} & \multirow{2}[4]{*}{Overall (H2)} & \multirow{2}[4]{*}{Defer (H2)} & \multirow{2}[4]{*}{0.998 (0.995-1.000)} & \multirow{2}[4]{*}{1.000 (1.000-1.000)} & Positive & 48    & 6     & 0.1875  & 0.9794  \\
\cmidrule{6-10}          &       &       &       &       & Negative & 104   & 11    & 0.4047  & 0.0557  \\
    \midrule
    \multirow{2}[4]{*}{MarginAssess (Frozen)} & \multirow{2}[4]{*}{Overall (H12)} & \multirow{2}[4]{*}{Defer (H12)} & \multirow{2}[4]{*}{0.973 (0.948-0.990)} & \multirow{2}[4]{*}{1.000 (1.000-1.000)} & Negative & 168   & 37    & 0.1488  & 0.4611  \\
\cmidrule{6-10}          &       &       &       &       & Positive & 26    & 16    & 0.1779  & 0.8463  \\
    \midrule
    \multirow{2}[4]{*}{LN\_Metastasis (Frozen)} & \multirow{2}[4]{*}{Overall (H2)} & \multirow{2}[4]{*}{Defer (H2)} & \multirow{2}[4]{*}{0.983 (0.941-1.000)} & \multirow{2}[4]{*}{0.999 (0.997-1.000)} & Negative & 90    & 41    & 0.1182  & 0.7732  \\
\cmidrule{6-10}          &       &       &       &       & Positive & 60    & 43    & 0.0380  & 1.0000  \\
    \midrule
    \multirow{3}[6]{*}{MassDiff (Frozen)} & \multirow{3}[6]{*}{Overall (H4)} & \multirow{3}[6]{*}{Defer (H4)} & \multirow{3}[6]{*}{0.963 (0.948-0.976)} & \multirow{3}[6]{*}{0.959 (0.877-1.000)} & IDC   & 125   & 5     & 0.6480  & 0.0183  \\
\cmidrule{6-10}          &       &       &       &       & Fibro & 171   & 9     & 0.1637  & 0.9530  \\
\cmidrule{6-10}          &       &       &       &       & Others & 82    & 8     & 0.2805  & 0.5262  \\
    \bottomrule
    \end{tabular}%
  }
  \label{tab:ks_stat_intra}%
\end{table}%

\begin{figure}
    \centering
    \captionsetup{name=Extended Data Figure}
    \includegraphics[width=1.0\linewidth]{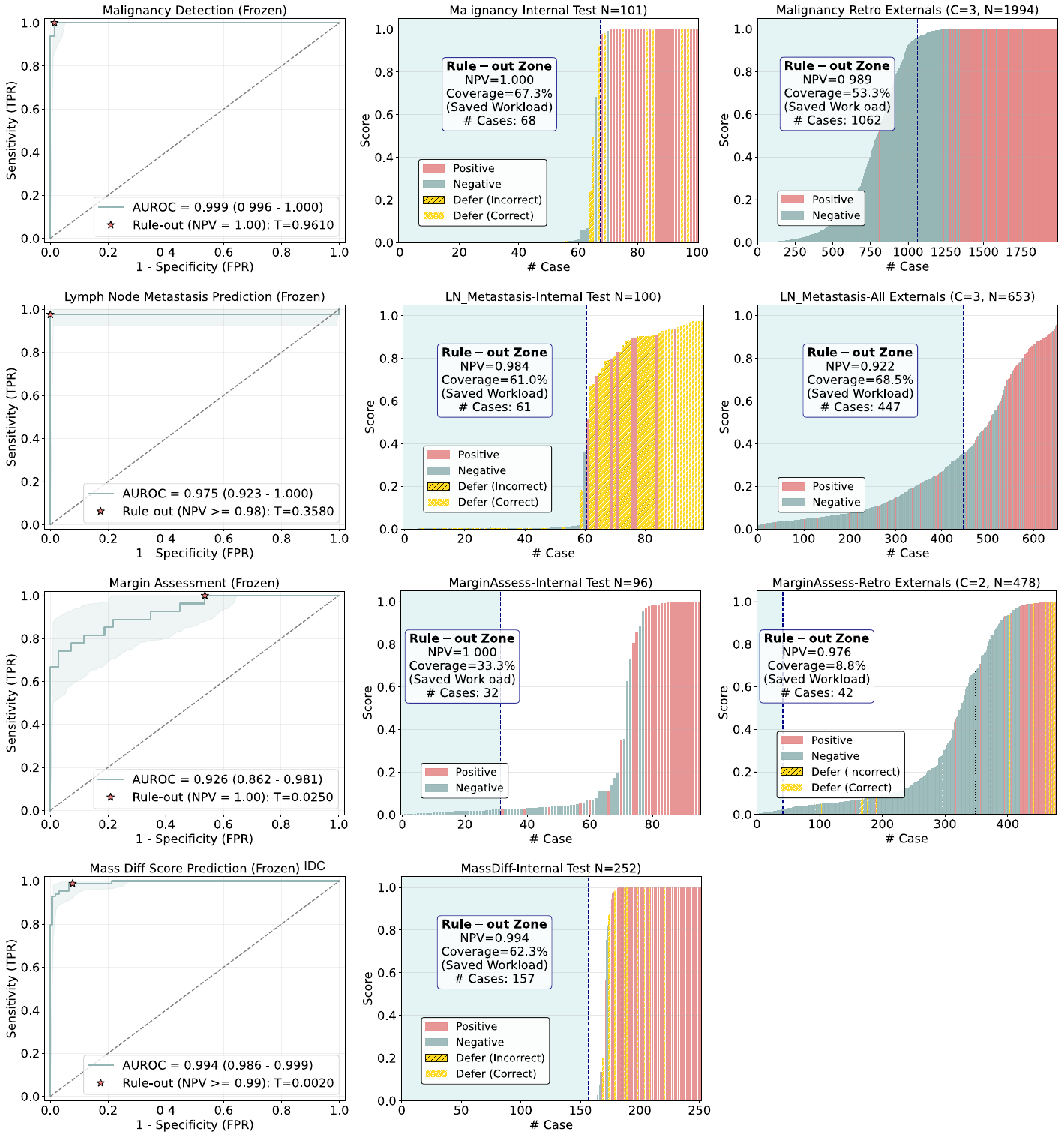}
    \caption{\textbf{Threshold Selection for Safety-constrained Rule-out Prescreening and Results of Rule-out at Intra-operative Stage.}}
    \label{fig:intra_all_rule_out}
\end{figure}

\begin{table}[htbp]
  \centering
  \captionsetup{name=Extended Data Table}
  \caption{\textbf{Safety-constrained Rule-out Prescreening at Intra-operative Stage.} T\_out refers to the threshold for rule-out.}
    \begin{tabular}{l|l|r|r|r|r|r}
    \toprule
    \multicolumn{1}{c|}{\textbf{Task}} & \multicolumn{1}{c|}{\textbf{Center}} & \multicolumn{1}{c|}{\textbf{T\_out}} & \multicolumn{1}{c|}{\textbf{Rule-out NPV}} & \multicolumn{1}{c|}{\textbf{Rule-out Coverage}} & \textbf{Rule-out Cases} & \textbf{Total Cases} \\
    \midrule
    \multicolumn{1}{c|}{\multirow{5}[6]{*}{Malignancy (Frozen)}} & Internal Test & \multicolumn{1}{c|}{\multirow{5}[6]{*}{0.961}} & 1.000  & 67.3\% & 68    & 101 \\
\cmidrule{2-2}\cmidrule{4-7}          & H15   &       & 0.940  & 67.1\% & 116   & 173 \\
          & H4    &       & 0.995  & 52.0\% & 853   & 1640 \\
          & H12+H16 &       & 0.989  & 51.4\% & 93    & 181 \\
\cmidrule{2-2}\cmidrule{4-7}          & All Externals &       & 0.989  & 53.3\% & 1062  & 1994 \\
    \midrule
    \multicolumn{1}{c|}{\multirow{4}[6]{*}{LN\_Metastasis (Frozen)}} & Internal Test & \multicolumn{1}{c|}{\multirow{4}[6]{*}{0.358}} & 0.984  & 61.0\% & 61    & 100 \\
\cmidrule{2-2}\cmidrule{4-7}          & H4    &       & 0.964  & 74.6\% & 223   & 299 \\
          & H12   &       & 0.969  & 49.2\% & 96    & 195 \\
\cmidrule{2-2}\cmidrule{4-7}          & All Externals &       & 0.922  & 68.5\% & 447   & 653 \\
    \midrule
    \multicolumn{1}{c|}{\multirow{4}[6]{*}{MarginAssess (Frozen)}} & Internal Test & \multicolumn{1}{c|}{\multirow{4}[6]{*}{0.025}} & 1.000  & 33.3\% & 32    & 96 \\
\cmidrule{2-2}\cmidrule{4-7}          & H4    &       & 0.974  & 13.7\% & 39    & 284 \\
          & H12   &       & 1.000  & 1.5\% & 3     & 194 \\
\cmidrule{2-2}\cmidrule{4-7}          & All Externals &       & 0.976  & 8.8\% & 42    & 478 \\
    \midrule
    MassDiff (Frozen) & Internal Test & 0.002 & 0.994  & 62.3\% & 157   & 252 \\
    \bottomrule
    \end{tabular}%
    \label{fig:intra_rule_out_results}
\end{table}%

\begin{table}[htbp]
  \centering
  \captionsetup{name=Extended Data Table}
  \caption{\textbf{Class Distribution on 27 Retrospective Cohorts from 10 Centers for Post-operative Stage}.}
  \scalebox{0.8}{
    \begin{tabular}{l|l|l|l|r|l}
    \toprule
    \multicolumn{1}{c|}{\textbf{Stage}} & \multicolumn{1}{c|}{\textbf{Task}} & \multicolumn{1}{c|}{\textbf{Center}} & \multicolumn{1}{c|}{\textbf{Setting}} & \multicolumn{1}{c|}{\textbf{\# Cases}} & \multicolumn{1}{c}{\textbf{Class (cases)}} \\
    \midrule
    Post-operative & Pathological Subtyping & TCGA & Internal & 1053  & IDC:842; ILC:211 \\
    Post-operative & Pathological Subtyping & H4    & Retrospective & 420   & idc:404; ilc:16 \\
    \midrule
    Post-operative & Tumor Grading & H4 & Internal & 393   & i+ii:294; iii:99 \\
    \midrule
    Post-operative & pCR Assessment & H4 & Internal & 127   & non-pcr:99; pcr:28 \\
    \midrule
    Post-operative & TNM-T Stage Prediction & H6 & Internal & 467   & t1:249; t1+:218 \\
    \midrule
    Post-operative & Lymphovascular Invasion Prediction & H4 & Internal & 480   & negative:303; positive:177 \\
    \midrule
    Post-operative & Perineural Invasion Prediction & H4 & Internal & 478   & negative:401; positive:77 \\
    \midrule
    Post-operative & AR Prediction & H6 & Internal & 1140  & positive:677; negative:463 \\
    \midrule
    Post-operative & CK5 Prediction & H6 & Internal & 961   & negative:753; positive:208 \\
    \midrule
    Post-operative & ER Prediction & H6 & Internal & 1548  & positive:781; negative:767 \\
    Post-operative & ER Prediction & H18   & Retrospective & 980   & positive:676; negative:304 \\
    Post-operative & ER Prediction & H20   & Retrospective & 250   & positive:175; negative:75 \\
    \midrule
    Post-operative & PR Prediction & H6 & Internal & 1556  & positive:933; negative:623 \\
    \midrule
    Post-operative & HER2 Prediction & H6 & Internal & 1344  & high:833; low:511 \\
    Post-operative & HER2 Prediction & H18   & Retrospective & 461   & high:238; low:223 \\
    \midrule
    Post-operative & KI67 Prediction & H4 & Internal & 309   & high:250; low:59 \\
    Post-operative & KI67 Prediction & H18   & Retrospective & 1079  & high:712; low:367 \\
    \midrule
    Post-operative & Molecular Subtyping & H6 & Internal & 2045  & lumb1:614; tnbc:589; luma:307; her2:292; lumb2:243 \\
    Post-operative & Molecular Subtyping & H13   & Retrospective & 418   & luma:102; her2:102; tnbc:101; lumb1:89; lumb2:24 \\
    Post-operative & Molecular Subtyping & H4+H14 & Retrospective & 180   & tnbc:89; her2:36; lumb2:22; luma:19; lumb1:14 \\
    Post-operative & Molecular Subtyping & H9    & Retrospective & 788   & tnbc:293; her2:149; luma:142; lumb1:122; lumb2:82 \\
    \midrule
    Post-operative & Mutation CDH1 Prediction & TCGA & Internal & 965   & no alteration:848; Mutated:117 \\
    \midrule
    Post-operative & Mutation GATA3 Prediction & TCGA & Internal & 1030  & no alteration:907; Mutated:123 \\
    \midrule
    Post-operative & Mutation PIK3CA Prediction & TCGA & Internal & 1030  & no alteration:692; Mutated:338 \\
    Post-operative & Mutation PIK3CA Prediction & CPTAC & Retrospective & 99    & no alteration:65; Mutated:34 \\
    \midrule
    Post-operative & Mutation TP53 Prediction & TCGA & Internal & 1030  & no alteration:690; Mutated:340 \\
    Post-operative & Mutation TP53 Prediction & CPTAC & Retrospective & 99    & no alteration:63; Mutated:36 \\
    \bottomrule
    \end{tabular}%
  }
  \label{tab:cls_distr_post}%
\end{table}%

\begin{table}[htbp]
  \centering
  \captionsetup{name=Extended Data Table}
  \caption{\textbf{Center Distribution of Tasks at Post-operative Stage}.}
  \scalebox{0.9}{
    \begin{tabular}{l|l|l}
    \toprule
    \multicolumn{1}{c|}{\textbf{Stage}} & \multicolumn{1}{c|}{\textbf{Task}} & \multicolumn{1}{c}{\textbf{Center Distribution (Case)}} \\
    \midrule
    Post-operative & Pathological Subtyping (surgical) & Train: 712; Val: 99; Test: 242; H4: 420 \\
    Post-operative & Tumor Grading (surgical) & Train: 274; Val: 40; Test: 79 \\
    Post-operative & pCR Assessment (surgical) & Train: 88; Val: 13; Test: 26 \\
    \midrule
    Post-operative & TNM-T Stage Prediction (surgical) & Train: 326; Val: 47; Test: 94 \\
    Post-operative & Lymphovascular Invasion Prediction (surgical) & Train: 336; Val: 48; Test: 96 \\
    Post-operative & Perineural Invasion Prediction (surgical) & Train: 334; Val: 48; Test: 96 \\
    \midrule
    Post-operative & AR Prediction (surgical) & Train: 798; Val: 114; Test: 228 \\
    Post-operative & CK5 Prediction (surgical) & Train: 672; Val: 96; Test: 193 \\
    Post-operative & ER Prediction (surgical) & Train: 1083; Val: 155; Test: 310; H18: 980; H20: 250 \\
    Post-operative & PR Prediction (surgical) & Train: 1088; Val: 156; Test: 312 \\
    Post-operative & HER2 Prediction (surgical) & Train: 940; Val: 135; Test: 269; H18: 461 \\
    Post-operative & KI67 Prediction (surgical) & Train: 216; Val: 31; Test: 62; H18: 1079 \\
    Post-operative & Molecular Subtyping (surgical) & Train: 1431; Val: 205; Test: 409; H13: 418; H4+H14: 180; H9: 788 \\
    Post-operative & Mutation CDH1 Prediction (surgical) & Train: 652; Val: 96; Test: 217 \\
    Post-operative & Mutation GATA3 Prediction (surgical) & Train: 697; Val: 98; Test: 235 \\
    Post-operative & Mutation PIK3CA Prediction (surgical) & Train: 697; Val: 98; Test: 235; CPTAC: 99 \\
    Post-operative & Mutation TP53 Prediction (surgical) & Train: 697; Val: 98; Test: 235; CPTAC: 99 \\
    \bottomrule
    \end{tabular}%
  }
  \label{tab:center_distr_post}%
\end{table}%

\begin{table}[htbp]
  \captionsetup{name=Extended Data Table}
  \caption{\textbf{Performance of BRAVE  for Post-operative Stage on 27 retrospective cohorts from 9 centers}, where `Internal' represents the result of the internal test set. Sensitivity, Specificity, and NPV are reported at the maximum Youden's Index. 95\% CI is included in parentheses.}
   \scalebox{0.6}{
    \begin{tabular}{l|l|l|l|l|l|l|l}
    \toprule
    \multicolumn{1}{c|}{\textbf{Task}} & \multicolumn{1}{c|}{\textbf{Center}} & \multicolumn{1}{c|}{\textbf{Type}} & \multicolumn{1}{c|}{\textbf{Setting}} & \multicolumn{1}{c|}{\textbf{Macro-AUC}} & \multicolumn{1}{c|}{\textbf{Sensitivity}} & \multicolumn{1}{c|}{\textbf{Specificity}} & \multicolumn{1}{c}{\textbf{NPV}} \\
    \midrule
    Pathological Subtyping (Surgical) & TCGA (Internal) & Morphology Assessment & Retrospective & 0.980 (0.962-0.993) & 0.956 (0.868-1.000) & 0.920 (0.821-0.985) & 0.990 (0.972-1.000) \\
    Pathological Subtyping (Surgical) & H4-Retro & Morphology Assessment & Retrospective & 0.949 (0.907-0.984) & 0.959 (0.800-1.000) & 0.834 (0.721-0.998) & 0.998 (0.992-1.000) \\
    \midrule
    Tumor Grading (Surgical) & H4 (Internal) & Morphology Assessment & Retrospective & 0.803 (0.662-0.918) & 0.780 (0.556-0.950) & 0.839 (0.733-0.968) & 0.920 (0.851-0.980) \\
    \midrule
    pCR Assessment (Surgical) & H4 (Internal) & Morphology Assessment & Retrospective & 0.949 (0.841-1.000) & 0.998 (1.000-1.000) & 0.921 (0.778-1.000) & 0.999 (1.000-1.000) \\
    \midrule
    TNM-T Stage Prediction (Surgical) & H6 (Internal) & Tumor Behavior & Retrospective & 0.825 (0.738-0.901) & 0.787 (0.609-0.923) & 0.797 (0.630-0.943) & 0.814 (0.719-0.918) \\
    \midrule
    Lymphovascular Invasion Prediction (Surgical) & H4 (Internal) & Tumor Behavior & Retrospective & 0.739 (0.615-0.848) & 0.698 (0.452-0.897) & 0.770 (0.574-0.934) & 0.821 (0.732-0.919) \\
    \midrule
    Perineural Invasion Prediction (Surgical) & H4 (Internal) & Tumor Behavior & Retrospective & 0.819 (0.719-0.903) & 0.962 (0.727-1.000) & 0.657 (0.500-0.902) & 0.992 (0.944-1.000) \\
    \midrule
    AR Prediction (Surgical) & H6 (Internal) & Molecular Inference & Retrospective & 0.748 (0.679-0.812) & 0.846 (0.656-0.930) & 0.596 (0.462-0.774) & 0.735 (0.596-0.846) \\
    \midrule
    CK5 Prediction (Surgical) & H6 (Internal) & Molecular Inference & Retrospective & 0.843 (0.775-0.901) & 0.826 (0.656-0.962) & 0.774 (0.614-0.916) & 0.943 (0.900-0.985) \\
    \midrule
    ER Prediction (Surgical) & H6 (Internal) & Molecular Inference & Retrospective & 0.908 (0.875-0.938) & 0.882 (0.796-0.942) & 0.833 (0.760-0.914) & 0.876 (0.810-0.932) \\
    ER Prediction (Surgical) & H18-Retro & Molecular Inference & Retrospective & 0.834 (0.807-0.860) & 0.786 (0.655-0.880) & 0.739 (0.635-0.864) & 0.617 (0.523-0.707) \\
    ER Prediction (Surgical) & H20-Retro & Molecular Inference & Retrospective & 0.807 (0.746-0.863) & 0.737 (0.567-0.917) & 0.773 (0.562-0.915) & 0.567 (0.464-0.744) \\
    \midrule
    PR Prediction (Surgical) & H6 (Internal) & Molecular Inference & Retrospective & 0.901 (0.866-0.936) & 0.860 (0.737-0.929) & 0.822 (0.738-0.931) & 0.801 (0.699-0.879) \\
    \midrule
    HER2 Prediction (Surgical) & H6 (Internal) & Molecular Inference & Retrospective & 0.850 (0.801-0.891) & 0.738 (0.606-0.929) & 0.832 (0.618-0.941) & 0.674 (0.582-0.846) \\
    HER2 Prediction (Surgical) & H18-Retro & Molecular Inference & Retrospective & 0.895 (0.866-0.922) & 0.808 (0.728-0.906) & 0.847 (0.730-0.911) & 0.807 (0.751-0.883) \\
    \midrule
    KI67 Prediction (Surgical) & H4 (Internal) & Molecular Inference & Retrospective & 0.843 (0.710-0.952) & 0.809 (0.460-0.979) & 0.840 (0.583-1.000) & 0.576 (0.308-0.890) \\
    KI67 Prediction (Surgical) & H18-Retro & Molecular Inference & Retrospective & 0.789 (0.761-0.817) & 0.656 (0.552-0.750) & 0.793 (0.690-0.887) & 0.546 (0.499-0.601) \\
    \midrule
    Molecular Subtyping (Surgical) & H6 (Internal) & Molecular Inference & Retrospective & 0.904 (0.867-0.935) & 0.861 (0.750-0.983) & 0.814 (0.655-0.909) & 0.972 (0.952-0.996) \\
    Molecular Subtyping (Surgical) & H13-Retro & Molecular Inference & Retrospective & 0.903 (0.870-0.931) & 0.893 (0.800-0.981) & 0.797 (0.681-0.875) & 0.960 (0.929-0.992) \\
    Molecular Subtyping (Surgical) & H4+H14-Retro & Molecular Inference & Retrospective & 0.937 (0.888-0.974) & 0.944 (0.808-1.000) & 0.857 (0.714-0.964) & 0.993 (0.976-1.000) \\
    Molecular Subtyping (Surgical) & H9-Retro & Molecular Inference & Retrospective & 0.842 (0.801-0.878) & 0.818 (0.717-0.907) & 0.765 (0.675-0.851) & 0.951 (0.929-0.972) \\
    \midrule
    Mutation CDH1 Prediction (Surgical) & TCGA (Internal) & Molecular Inference & Retrospective & 0.874 (0.745-0.953) & 0.931 (0.789-1.000) & 0.820 (0.733-0.901) & 0.992 (0.976-1.000) \\
    \midrule
    Mutation GATA3 Prediction (Surgical) & TCGA (Internal) & Molecular Inference & Retrospective & 0.711 (0.619-0.796) & 0.823 (0.609-1.000) & 0.592 (0.302-0.790) & 0.970 (0.941-1.000) \\
    \midrule
    Mutation PIK3CA Prediction (Surgical) & TCGA (Internal) & Molecular Inference & Retrospective & 0.641 (0.565-0.712) & 0.621 (0.279-0.859) & 0.658 (0.396-0.945) & 0.803 (0.744-0.882) \\
    Mutation PIK3CA Prediction (Surgical) & CPTAC-Retro & Molecular Inference & Retrospective & 0.713 (0.595-0.814) & 0.671 (0.429-0.900) & 0.743 (0.483-0.923) & 0.818 (0.740-0.921) \\
    \midrule
    Mutation TP53 Prediction (Surgical) & TCGA (Internal) & Molecular Inference & Retrospective & 0.829 (0.770-0.884) & 0.798 (0.648-0.917) & 0.771 (0.628-0.895) & 0.873 (0.809-0.934) \\
    Mutation TP53 Prediction (Surgical) & CPTAC-Retro & Molecular Inference & Retrospective & 0.825 (0.733-0.903) & 0.917 (0.719-1.000) & 0.677 (0.523-0.859) & 0.938 (0.840-1.000) \\
    \bottomrule
    \end{tabular}%
   }
  \label{tab:result_post}%
\end{table}%

\begin{table}[htbp]
  \centering
  \captionsetup{name=Extended Data Table}
  \caption{\textbf{Comparison between Challenging Subgroup Cohorts at Post-operative Stage}. KS Statistic is obtained by Kolmogorov-Smirnov (KS) Test. 95\% CI is included in parentheses.}
  \scalebox{0.75}{
    \begin{tabular}{c|c|c|c|c|c|c|c|c|c}
    \toprule
    \textbf{Task} & \textbf{Baseline Cohort} & \textbf{Subgroup} & \textbf{AUC (baseline)} & \textbf{AUC (subgroup)} & \textbf{Class} & \textbf{Baseline N} & \textbf{Subgroup N} & \textbf{KS Statistic} & \textbf{P-Value} \\
    \midrule
    \multirow{4}[8]{*}{PIK3CA} & \multirow{2}[4]{*}{Overall (TCGA)} & \multirow{2}[4]{*}{HR+HER2-} & \multirow{2}[4]{*}{0.644 (0.565-0.712)} & \multirow{2}[4]{*}{0.619 (0.507-0.730)} & Negative & 163   & 80    & 0.1621 & 0.1035  \\
\cmidrule{6-10}          &       &       &       &       & Positive & 72    & 44    & 0.1263 & 0.7211  \\
\cmidrule{2-10}          & \multirow{2}[4]{*}{PreMeno} & \multirow{2}[4]{*}{PostMeno} & \multirow{2}[4]{*}{0.594 (0.430-0.760)} & \multirow{2}[4]{*}{0.641 (0.536-0.736)} & Negative & 32    & 106   & 0.1621 & 0.4791  \\
\cmidrule{6-10}          &       &       &       &       & Positive & 21    & 43    & 0.1229 & 0.9603  \\
    \midrule
    \multirow{2}[4]{*}{TP53} & \multirow{2}[4]{*}{Overall (TCGA)} & \multirow{2}[4]{*}{HR+HER2-} & \multirow{2}[4]{*}{0.830 (0.770-0.884)} & \multirow{2}[4]{*}{0.835 (0.745-0.916)} & Negative & 150   & 94    & 0.1339 & 0.2251  \\
\cmidrule{6-10}          &       &       &       &       & Positive & 85    & 30    & 0.1608 & 0.5560  \\
    \midrule
    GATA3 & Overall (TCGA) & HR-HER2+ & 0.711 (0.619-0.796) & NaN   & Negative & 210   & 11    & 0.2325 & 0.5454  \\
    \bottomrule
    \end{tabular}%
  }
  \label{tab:ks_stat_post}%
\end{table}%

\begin{table}[htbp]
  \centering
  \captionsetup{name=Extended Data Table}
  \caption{\textbf{Association of H\&E-derived TP53 Scores from BRAVE with Genomic Instability at Post-operative Stage}. 
  Spearman correlation $\rho$ measures the association between the H\&E-derived TP53 score and each genomic variable ($0.3 < \rho < 0.5$ indicates moderate positive association), 
  and the corresponding Spearman $p$ value indicates the statistical significance of this association. 
  Pearson $R$ quantifies the linear correlation between bin-level predicted scores and ground-truth mutation rates (GT rate) with values close to 1 indicating strong consistency, 
  and the corresponding Pearson $p$ value indicates the statistical significance of this consistency.}
    \begin{tabular}{l|c|c|c|c|c}
    \toprule
    \textbf{Variable} & \textbf{N\_samples} & \textbf{Spearman $\rho$} & \textbf{Spearman $p$} & \textbf{Pearson $R$} & \textbf{Pearson $p$} \\
    \midrule
    log\_Mutation\_Count & 921   & 0.3387 & 3.76E-26 & 0.9842 & 9.80E-06 \\
    \midrule
    log\_TMB & 921   & 0.3283 & 1.39E-24 & 0.9901 & 2.43E-06 \\
    \midrule
    Fraction Genome Altered & 1009  & 0.4239 & 2.81E-45 & 0.9864 & 6.22E-06 \\
    \midrule
    Aneuploidy Score & 982   & 0.4252 & 2.23E-44 & 0.9889 & 3.37E-06 \\
    \bottomrule
    \end{tabular}%
  \label{tab:trend_tp53}%
\end{table}%

\begin{table}[htbp]
  \centering
  \captionsetup{name=Extended Data Table}
  \caption{\textbf{Independent predictive value of H\&E-derived AI scores for genetic mutations in the multivariate analysis}. 
  Odds Ratios (OR) for all continuous variables (Age, TMB, FGA, and AI Score) represent the relative risk per 1 standard deviation (SD) increase, where `N/A' indicates the unstable estimate. 
  95\% CI is included in parentheses. 
  The incremental predictive value of the AI score was assessed using a likelihood ratio (LR) test comparing the full model (Clinical + Genomic Burden + AI score) with a baseline model (Clinical + Genomic Burden). }
  \begin{tabular}{c|l|l|l}
    \toprule
    \textbf{Gene} & \textbf{Variable} & \textbf{OR (95\% CI)} & \textbf{P-value} \\
    \midrule
    \multirow{14}[28]{*}{TP53} & Age (per 1 SD) & 0.91 (0.74--1.13) & p=0.39 \\
  \cmidrule{2-4}          & log(TMB) (per 1 SD) & 1.17 (0.94--1.44) & p=0.16 \\
  \cmidrule{2-4}          & FGA (per 1 SD) & 1.56 (1.25--1.95) & p<0.001 \\
  \cmidrule{2-4}          & Stage II vs Stage I & 0.57 (0.32--1.02) & p=0.06 \\
  \cmidrule{2-4}          & Stage III vs Stage I & 1.05 (0.54--2.02) & p=0.89 \\
  \cmidrule{2-4}          & Stage IV vs Stage I & 0.73 (0.12--4.46) & p=0.73 \\
  \cmidrule{2-4}          & ILC vs IDC & 0.35 (0.14--0.84) & p=0.02 \\
  \cmidrule{2-4}          & Mixed vs IDC & 1.25 (0.39--4.06) & p=0.70 \\
  \cmidrule{2-4}          & HR+/HER2- vs LumA & 0.82 (0.48--1.41) & p=0.47 \\
  \cmidrule{2-4}          & HR-/HER2+ vs LumA & 2.02 (0.71--5.77) & p=0.19 \\
  \cmidrule{2-4}          & TNBC vs LumA & 1.29 (0.61--2.72) & p=0.50 \\
  \cmidrule{2-4}          & H\&E-derived TP53 AI Score (per 1 SD) & 4.19 (3.12--5.61) & p<0.001 \\
\cmidrule{2-4}          & LR Test vs. Baseline (Cli+Burden) & $\chi^2$=112.07 & \multirow{2}{*}{p<0.001} \\
\cmidrule{2-3}          & Delta AUROC / Delta AUPRC & +4.8\% / +8.5\% &  \\
    \bottomrule
    \end{tabular}%
  \label{tab:multivariate_analysis}%
\end{table}%

\begin{table}[htbp]
  \centering
  \captionsetup{name=Extended Data Table}
  \caption{\textbf{Internal Triage Simulation of different triage strategies for prioritizing genomic testing based on AI-derived scores and clinical variables}: 
  (1) Clinical only, the current standard utilizing clinicopathologic variables; (2) BRAVE only, using solely the H\&E-derived AI score; and (3) Clinical + BRAVE, the integrated model combining both clinical and AI features. 
  Performance is evaluated across multiple Genomic Testing Rates, representing the top percentage of the population prioritized for testing. 
  Enrichment vs. Prevalence reflects the precision and the fold-increase in mutation detection compared to the population baseline. 
  Tests per Mutation Found represent the number of patients required to undergo testing to identify one true mutation carrier.}
  \scalebox{0.9}{
  \begin{tabular}{c|l|l|l|l|l}
    \toprule
    \textbf{Genomic Testing Rate (\%)} & \multicolumn{1}{c|}{\textbf{Strategy}} & \multicolumn{1}{c|}{\textbf{Sensitivity}} & \multicolumn{1}{c|}{\textbf{PPV}} & \multicolumn{1}{c|}{\textbf{Enrichment vs Prev.}} & \multicolumn{1}{c}{\textbf{Tests per Mutation}} \\
    \midrule
    \multirow{3}[6]{*}{$\bm{Selected: 10\%}$} & Clinical (current standard) & 20.0\% & 70.8\% & 1.96  & 1.4 \\
\cmidrule{2-6}          & BRAVE only & 23.5\% & 83.3\% & 2.30  & 1.2 \\
\cmidrule{2-6}          & $\bm{Clinical + BRAVE}$ & 21.2\% & 75.0\% & 2.07  & 1.3 \\
    \midrule
    \multirow{3}[6]{*}{$\bm{Selected: 20\%}$} & Clinical (current standard) & 37.6\% & 68.1\% & 1.88  & 1.5 \\
\cmidrule{2-6}          & BRAVE only & 42.4\% & 76.6\% & 2.12  & 1.3 \\
\cmidrule{2-6}          & $\bm{Clinical + BRAVE}$ & 43.5\% & 78.7\% & 2.18  & 1.3 \\
    \midrule
    \multirow{3}[6]{*}{$\bm{Selected: 30\%}$} & Clinical (current standard) & 48.2\% & 57.7\% & 1.60  & 1.7 \\
\cmidrule{2-6}          & BRAVE only & 58.8\% & 70.4\% & 1.95  & 1.4 \\
\cmidrule{2-6}          & $\bm{Clinical + BRAVE}$ & 60.0\% & 71.8\% & 1.99  & 1.4 \\
    \midrule
    \multirow{3}[6]{*}{$\bm{Selected: 40\%}$} & Clinical (current standard) & 56.5\% & 51.1\% & 1.41  & 2.0 \\
\cmidrule{2-6}          & BRAVE only & 75.3\% & 68.1\% & 1.88  & 1.5 \\
\cmidrule{2-6}          & $\bm{Clinical + BRAVE}$ & 72.9\% & 66.0\% & 1.82  & 1.5 \\
    \midrule
    \multirow{3}[6]{*}{$\bm{Selected: 50\%}$} & Clinical (current standard) & 68.2\% & 49.2\% & 1.36  & 2.0 \\
\cmidrule{2-6}          & BRAVE only & 83.5\% & 60.2\% & 1.66  & 1.7 \\
\cmidrule{2-6}          & $\bm{Clinical + BRAVE}$ & 83.5\% & 60.2\% & 1.66  & 1.7 \\
    \midrule
    \multirow{3}[6]{*}{$\bm{Selected: 100\%}$} & Clinical (current standard) & 100.0\% & 36.2\% & 1.00  & 2.8 \\
\cmidrule{2-6}          & BRAVE only & 100.0\% & 36.2\% & 1.00  & 2.8 \\
\cmidrule{2-6}          & $\bm{Clinical + BRAVE}$ & 100.0\% & 36.2\% & 1.00  & 2.8 \\
    \bottomrule
    \end{tabular}%
  }
  \label{tab:triage_internal}%
\end{table}%

\begin{table}[htbp]
  \centering
  \captionsetup{name=Extended Data Table}
  \caption{\textbf{External validation (CPTAC) of the AI-derived clinical triage policy established on the TCGA cohort}. 
  For each Intended Rate, a strict cut-off (Threshold) was predetermined based on the internal TCGA score distribution. 
  The Actual Rate indicates the true proportion of CPTAC cases that fell above the threshold.}
  \scalebox{0.8}{
    \begin{tabular}{c|c|c|c|c|c|c}
    \toprule
    \textbf{Intended Rate (TCGA)} & \textbf{Threshold (TCGA)} & \textbf{Actual Rate (CPTAC)} & \textbf{Sensitivity} & \textbf{PPV} & \textbf{Enrichment vs Prev.} & \textbf{Tests per Mutation} \\
    \midrule
    10\%  & 0.831 & 2.8\% & 7.7\% & 100.0\% & 2.77  & 1.0 \\
    \midrule
    20\%  & 0.656 & 20.4\% & 35.9\% & 63.6\% & 1.76  & 1.6 \\
    \midrule
    30\%  & 0.445 & 56.5\% & 92.3\% & 59.0\% & 1.63  & 1.7 \\
    \midrule
    40\%  & 0.286 & 75.0\% & 100.0\% & 48.1\% & 1.33  & 2.1 \\
    \midrule
    50\%  & 0.202 & 84.3\% & 100.0\% & 42.9\% & 1.19  & 2.3 \\
    \midrule
    60\%  & 0.136 & 95.4\% & 100.0\% & 37.9\% & 1.05  & 2.6 \\
    \midrule
    70\%  & 0.095 & 97.2\% & 100.0\% & 37.1\% & 1.03  & 2.7 \\
    \midrule
    80\%  & 0.068 & 98.1\% & 100.0\% & 36.8\% & 1.02  & 2.7 \\
    \midrule
    90\%  & 0.044 & 98.1\% & 100.0\% & 36.8\% & 1.02  & 2.7 \\
    \bottomrule
    \end{tabular}%
  }
  \label{tab:triage_ext}%
\end{table}%

\begin{table}[htbp]
  \centering
  \captionsetup{name=Extended Data Table}
  \caption{\textbf{Triage performance in prospective observational validation using thresholds selected in the internal cohort.} 
  For each task, the low and high thresholds were first determined in the internal cohort and then applied to the prospective observational cohort for triage. 
  The table reports the resulting rule-out and rule-in coverage, together with NPV at the rule-out threshold and PPV at the rule-in threshold.}
  \scalebox{0.8}{  
  \begin{tabular}{l|l|c|c|c|c|c|c|c}
    \toprule
    \multicolumn{1}{c|}{\textbf{Task}} & \multicolumn{1}{c|}{\textbf{Center}} & \textbf{N} & \textbf{T\_low} & \textbf{T\_high} & \textbf{RuleOut\_Coverage} & \textbf{NPV@RuleOut} & \textbf{RuleIn\_Coverage} & \textbf{PPV@RuleIn} \\
    \midrule
    Pre-Malignancy & H4    & 882   & \multirow{2}[2]{*}{0.011 } & \multirow{2}[2]{*}{0.999} & 76.9\% & 0.953 (0.937-0.969) & 0.7\% & 1.000 (1.000-1.000) \\
    Pre-Malignancy & Internal-Test & 168   &       &       & 83.3\% & 1.000 (1.000-1.000) & 13.1\% & 1.000 (1.000-1.000) \\
    \midrule
    Pre-HER2 & H4    & 112   & \multirow{2}[2]{*}{0.016 } & \multirow{2}[2]{*}{0.986} & 20.5\% & 0.957 (0.873-1.000) & 6.3\% & 1.000 (1.000-1.000) \\
    Pre-HER2 & Internal-Test & 267   &       &       & 10.1\% & 1.000 (1.000-1.000) & 4.1\% & 0.909 (0.739-1.000) \\
    \midrule
    Intra-Malignancy & H2    & 107   & \multirow{2}[2]{*}{0.961 } & \multirow{2}[2]{*}{N/A} & 70.1\% & 0.973 (0.937-1.000) & N/A   & N/A \\
    Intra-Malignancy & Internal-Test & 101   &       &       & 67.3\% & 1.000 (1.000-1.000) & N/A   & N/A \\
    \midrule
    Post-PathSubtype & H6    & 156   & \multirow{2}[2]{*}{0.020 } & \multirow{2}[2]{*}{N/A} & 78.8\% & 1.000 (1.000-1.000) & N/A   & N/A \\
    Post-PathSubtype & Internal-Test & 240   &       &       & 68.3\% & 1.000 (1.000-1.000) & N/A   & N/A \\
    \bottomrule
    \end{tabular}%
  }
  \label{tab:pros_triage_threshold_perform}%
\end{table}%

\begin{table}[htbp]
  \centering
  \captionsetup{name=Extended Data Table}
  \caption{\textbf{Paired distribution of biomarker status (labels from pathologist) between pre-operative biopsy and post-operative surgical specimens, based on pathologist labels.} PP: both pre and post positive; PN: pre positive but post negative; NP: pre negative but post positive; NN: both pre and post negative.}
    \begin{tabular}{c|c|c|c|c|c|c|c|c|c}
    \toprule
    \textbf{Biomarker} & \textbf{Total Paired (N)} & \textbf{Pre\_Positive} & \textbf{Pre\_Negative} & \textbf{Post\_Positive} & \textbf{Post\_Negative} & \textbf{PP} & \textbf{PN} & \textbf{NP} & \textbf{NN} \\
    \midrule
    ER    & 77    & 58    & 19    & 58    & 19    & 57    & 1     & 1     & 18 \\
    \midrule
    HER2  & 28    & 12    & 16    & 12    & 16    & 12    & 0     & 0     & 16 \\
    \midrule
    KI67  & 81    & 65    & 16    & 71    & 10    & 63    & 2     & 8     & 8 \\
    \midrule
    PR    & 65    & 38    & 27    & 37    & 28    & 33    & 5     & 4     & 23 \\
    \bottomrule
    \end{tabular}%
  \label{tab:pros_biomarker_paired_distribution}%
\end{table}%

\begin{table}[htbp]
  \centering
  \captionsetup{name=Extended Data Table}
  \caption{\textbf{Concordance of biomarker status (labels from pathologists) between pre-operative biopsy and post-operative surgical specimens.} 
  Concordance quantifies the overall paired agreement, Cohen's Kappa measures agreement beyond chance, and the McNemar test assesses whether discordant paired classifications are directionally asymmetric. 
  Higher concordance and Kappa values indicate stronger agreement, whereas McNemar $p>0.05$ value indicates no significant directional asymmetry between discordant paired classifications.}
    \begin{tabular}{l|c|c|c|c}
    \toprule
    \multicolumn{1}{c|}{\textbf{Biomarker}} & \textbf{Total Paired (N)} & \textbf{Concordance (95\% CI)} & \textbf{Cohen's Kappa} & \textbf{McNemar (p-value)} \\
    \midrule
    ER    & 77    & 97.4 (91.0-99.3) & 0.930 & 1.000 \\
    \midrule
    HER2  & 28    & 100.0 (87.9-100.0) & 1.000 & - \\
    \midrule
    KI67  & 81    & 87.7 (78.7-93.2) & 0.546 & 0.109 \\
    \midrule
    PR    & 65    & 86.2 (75.7-92.5) & 0.716 & 1.000 \\
    \bottomrule
    \end{tabular}%
  \label{tab:pros_biomarker_concordance_stats}%
\end{table}%

\begin{table}[htbp]
  \centering
  \captionsetup{name=Extended Data Table}
  \caption{\textbf{Reference results from prospective observational validation cohorts, including case numbers, class distributions, and Macro-AUC (95\% confidence intervals).} Although decision thresholds were fixed in advance, these cases were collected consecutively over a subsequent time period and their ground-truth labels became available after follow-up. 
  Therefore, the Macro-AUC results are provided for reference only and mainly to facilitate comparison with the retrospective cohorts.}
    \begin{tabular}{l|l|r|l|l}
    \toprule
    \multicolumn{1}{c|}{\textbf{Task}} & \multicolumn{1}{c|}{\textbf{Center}} & \multicolumn{1}{c|}{\textbf{N (Case)}} & \multicolumn{1}{c|}{\textbf{Class Distribution (Case)}} & \multicolumn{1}{c}{\textbf{Macro-AUC}} \\
    \midrule
    Pre-Malignancy & H4    & 882   & malignant: 212; non-malignant: 670 & 0.9749 (0.9645-0.9839) \\
    Pre-Malignancy & Internal Test & 168   & malignant: 133; non-malignant: 705 & 0.9997 (0.9984-1.0000) \\
    \midrule
    Pre-MassDiff & H4    & 646   & fibro: 413; idc: 160; udh: 73 & 0.9217 (0.8874-0.9548) \\
    Pre-MassDiff & Internal Test & 168   & fibro: 415; idc: 133; udh: 290 & 0.9712 (0.9041-1.0000) \\
    \midrule
    Pre-ER & H4    & 197   & negative: 48; positive: 149 & 0.8851 (0.8349-0.9308) \\
    Pre-ER & Internal Test & 297   & negative: 474; positive: 1011 & 0.8714 (0.8262-0.9122) \\
    \midrule
    Pre-HER2 & H4    & 112   & negative: 76; positive: 36 & 0.9123 (0.8473-0.9658) \\
    Pre-HER2 & Internal Test & 267   & negative: 857; positive: 474 & 0.8245 (0.7706-0.8705) \\
    \midrule
    Pre-KI67 & H4    & 223   & negative: 64; positive: 159 & 0.8400 (0.7865-0.8858) \\
    Pre-KI67 & Internal Test & 297   & negative: 310; positive: 1171 & 0.8157 (0.7529-0.8718) \\
    \midrule
    Pre-PR & H4    & 161   & negative: 61; positive: 100 & 0.8591 (0.7962-0.9149) \\
    Pre-PR & Internal Test & 297   & negative: 364; positive: 1119 & 0.7729 (0.7090-0.8281) \\
    \midrule
    Intra-Malignancy & H2    & 107   & malignant: 25; non-malignant: 82 & 0.9698 (0.9330-0.9942) \\
    Intra-Malignancy & Internal Test & 101   & malignant: 158; non-malignant: 344 & 0.9991 (0.9958-1.0000) \\
    \midrule
    Intra-Margin & H2    & 128   & negative: 94; positive: 34 & 0.8829 (0.8122-0.9416) \\
    Intra-Margin & Internal Test & 96    & negative: 346; positive: 132 & 0.9269 (0.8623-0.9810) \\
    \midrule
    Post-Grading & H4    & 77    & i+ii: 59; iii: 18 & 0.9331 (0.8671-0.9816) \\
    Post-Grading & H6    & 161   & i+ii: 97; iii: 64 & 0.8969 (0.8431-0.9438) \\
    Post-Grading & Internal Test & 79    & i+ii: 294; iii: 99 & 0.8034 (0.6621-0.9179) \\
    \midrule
    Post-PathSubtype (PSub) & H6    & 156   & idc: 153; ilc: 3 & 0.9873 (0.9548-1.0000) \\
    Post-PathSubtype (PSub) & Internal Test & 240   & idc: 842; ilc: 211 & 0.9802 (0.9624-0.9931) \\
    \bottomrule
    \end{tabular}%
  \label{tab:pros_auc}%
\end{table}%

\begin{table}[htbp]
  \centering
  \captionsetup{name=Extended Data Table}
  \caption{Balanced Accuracy, Time (s), Confidence Score and their 95\% CI from 8 pathologists on two conditions (with/without AI) for Malignancy Detection (Biopsy) using H\&E slides only, and Balanced Accuracy from BRAVE for referecne.}
  \scalebox{0.7}{
    \begin{tabular}{c|c|c|l|c|c|c|c}
    \toprule
    \textbf{Experience} & \textbf{Pathologist} & \textbf{Task} & \multicolumn{1}{c|}{\textbf{Group}} & \textbf{Balanced\_Accuracy} & \textbf{Time (s)} & \textbf{Confidence Score} & \textbf{BAccuracy\_Model} \\
    \midrule
    \multicolumn{1}{l|}{Junior} & \multicolumn{1}{l|}{P1} & \multicolumn{1}{l|}{Malignancy Detection (Biopsy)} & WithAI & 0.901 (0.833-0.957) & 105.641 (102.729-108.931) & 8.786 (8.580-8.980) & \multirow{18}[18]{*}{1.000 (1.000-1.000)} \\
    \multicolumn{1}{l|}{Junior} & \multicolumn{1}{l|}{P1} & \multicolumn{1}{l|}{Malignancy Detection (Biopsy)} & WithoutAI & 0.731 (0.622-0.827) & 119.318 (116.550-121.990) & 7.545 (7.300-7.800) &  \\
\cmidrule{1-7}    \multicolumn{1}{l|}{Junior} & \multicolumn{1}{l|}{P2} & \multicolumn{1}{l|}{Malignancy Detection (Biopsy)} & WithAI & 0.943 (0.875-0.988) & 153.397 (129.729-178.452) & 8.102 (7.810-8.370) &  \\
    \multicolumn{1}{l|}{Junior} & \multicolumn{1}{l|}{P2} & \multicolumn{1}{l|}{Malignancy Detection (Biopsy)} & WithoutAI & 0.800 (0.708-0.875) & 166.887 (142.748-192.901) & 6.972 (6.720-7.190) &  \\
\cmidrule{1-7}    \multicolumn{1}{l|}{Junior} & \multicolumn{1}{l|}{P3} & \multicolumn{1}{l|}{Malignancy Detection (Biopsy)} & WithAI & 1.000 (1.000-1.000) & 119.389 (110.358-128.451) & 9.107 (8.980-9.240) &  \\
    \multicolumn{1}{l|}{Junior} & \multicolumn{1}{l|}{P3} & \multicolumn{1}{l|}{Malignancy Detection (Biopsy)} & WithoutAI & 0.881 (0.799-0.947) & 134.270 (124.529-144.030) & 7.857 (7.720-8.000) &  \\
\cmidrule{1-7}    \multicolumn{1}{l|}{Junior} & \multicolumn{1}{l|}{P4} & \multicolumn{1}{l|}{Malignancy Detection (Biopsy)} & WithAI & 0.979 (0.955-1.000) & 91.232 (86.039-96.452) & 9.086 (8.910-9.250) &  \\
    \multicolumn{1}{l|}{Junior} & \multicolumn{1}{l|}{P4} & \multicolumn{1}{l|}{Malignancy Detection (Biopsy)} & WithoutAI & 0.891 (0.816-0.954) & 106.386 (100.579-111.903) & 8.125 (7.950-8.300) &  \\
\cmidrule{1-7}    \multicolumn{1}{l|}{Senior} & \multicolumn{1}{l|}{P5} & \multicolumn{1}{l|}{Malignancy Detection (Biopsy)} & WithAI & 1.000 (1.000-1.000) & 41.937 (37.808-46.233) & 9.878 (9.808-9.939) &  \\
    \multicolumn{1}{l|}{Senior} & \multicolumn{1}{l|}{P5} & \multicolumn{1}{l|}{Malignancy Detection (Biopsy)} & WithoutAI & 0.944 (0.875-1.000) & 48.023 (43.394-52.921) & 8.939 (8.889-8.980) &  \\
\cmidrule{1-7}    \multicolumn{1}{l|}{Senior} & \multicolumn{1}{l|}{P6} & \multicolumn{1}{l|}{Malignancy Detection (Biopsy)} & WithAI & 0.981 (0.935-1.000) & 87.862 (80.569-95.650) & 9.610 (9.500-9.720) &  \\
    \multicolumn{1}{l|}{Senior} & \multicolumn{1}{l|}{P6} & \multicolumn{1}{l|}{Malignancy Detection (Biopsy)} & WithoutAI & 0.962 (0.900-1.000) & 97.588 (88.289-107.316) & 9.128 (8.970-9.270) &  \\
\cmidrule{1-7}    \multicolumn{1}{l|}{Senior} & \multicolumn{1}{l|}{P7} & \multicolumn{1}{l|}{Malignancy Detection (Biopsy)} & WithAI & 0.993 (0.977-1.000) & 57.710 (55.469-60.031) & 10.000 (10.000-10.000) &  \\
    \multicolumn{1}{l|}{Senior} & \multicolumn{1}{l|}{P7} & \multicolumn{1}{l|}{Malignancy Detection (Biopsy)} & WithoutAI & 0.916 (0.873-0.956) & 67.442 (65.224-69.715) & 8.574 (8.418-8.724) &  \\
\cmidrule{1-7}    \multicolumn{1}{l|}{Senior} & \multicolumn{1}{l|}{P8} & \multicolumn{1}{l|}{Malignancy Detection (Biopsy)} & WithAI & 1.000 (1.000-1.000) & 50.726 (47.440-53.900) & 9.721 (9.610-9.830) &  \\
    \multicolumn{1}{l|}{Senior} & \multicolumn{1}{l|}{P8} & \multicolumn{1}{l|}{Malignancy Detection (Biopsy)} & WithoutAI & 0.944 (0.885-0.993) & 68.968 (66.300-71.781) & 9.060 (8.890-9.220) &  \\
\cmidrule{1-7}    \multicolumn{3}{c|}{\multirow{2}[2]{*}{Mean}} & WithAI & 0.975$\pm$0.033 & 88.487$\pm$35.374 & 9.286$\pm$0.600 &  \\
    \multicolumn{3}{c|}{} & WithoutAI & 0.884$\pm$0.075 & 101.110$\pm$36.713 & 8.275$\pm$0.733 &  \\
    \bottomrule
    \end{tabular}%
    }
  \label{tab:all_reader_results_biopsy}%
\end{table}%

\begin{table}[htbp]
  \centering
  \captionsetup{name=Extended Data Table}
  \caption{Balanced Accuracy, Time (s), Confidence Score and their 95\% CI from 8 pathologists on two conditions (with/without AI) for Malignancy Detection (Frozen) using H\&E slides only, and Balanced Accuracy from BRAVE for referecne.}
  \scalebox{0.7}{
    \begin{tabular}{c|c|c|l|c|c|c|c}
    \toprule
    \textbf{Experience} & \textbf{Pathologist} & \textbf{Task} & \multicolumn{1}{c|}{\textbf{Group}} & \textbf{Balanced\_Accuracy} & \textbf{Time (s)} & \textbf{Confidence Score} & \textbf{BAccuracy\_Model} \\
    \midrule
    \multicolumn{1}{l|}{Junior} & \multicolumn{1}{l|}{P1} & \multicolumn{1}{l|}{Malignancy Detection (Frozen)} & WithAI & 0.925 (0.872-0.971) & 54.890 (51.528-58.410) & 8.679 (8.430-8.920) & \multirow{18}[4]{*}{0.955 (0.941-0.967)} \\
    \multicolumn{1}{l|}{Junior} & \multicolumn{1}{l|}{P1} & \multicolumn{1}{l|}{Malignancy Detection (Frozen)} & WithoutAI & 0.769 (0.676-0.852) & 72.452 (70.040-74.940) & 7.626 (7.430-7.790) &  \\
    \multicolumn{1}{l|}{Junior} & \multicolumn{1}{l|}{P2} & \multicolumn{1}{l|}{Malignancy Detection (Frozen)} & WithAI & 0.932 (0.889-0.970) & 89.857 (78.569-103.674) & 8.024 (7.750-8.300) &  \\
    \multicolumn{1}{l|}{Junior} & \multicolumn{1}{l|}{P2} & \multicolumn{1}{l|}{Malignancy Detection (Frozen)} & WithoutAI & 0.856 (0.800-0.907) & 99.969 (88.149-114.874) & 6.016 (5.760-6.260) &  \\
    \multicolumn{1}{l|}{Junior} & \multicolumn{1}{l|}{P3} & \multicolumn{1}{l|}{Malignancy Detection (Frozen)} & WithAI & 0.955 (0.918-0.985) & 118.845 (112.189-126.352) & 9.133 (8.990-9.270) &  \\
    \multicolumn{1}{l|}{Junior} & \multicolumn{1}{l|}{P3} & \multicolumn{1}{l|}{Malignancy Detection (Frozen)} & WithoutAI & 0.886 (0.829-0.936) & 141.630 (135.198-149.310) & 8.032 (7.900-8.180) &  \\
    \multicolumn{1}{l|}{Junior} & \multicolumn{1}{l|}{P4} & \multicolumn{1}{l|}{Malignancy Detection (Frozen)} & WithAI & 0.961 (0.925-0.992) & 54.146 (50.340-57.870) & 8.324 (8.160-8.510) &  \\
    \multicolumn{1}{l|}{Junior} & \multicolumn{1}{l|}{P4} & \multicolumn{1}{l|}{Malignancy Detection (Frozen)} & WithoutAI & 0.857 (0.782-0.924) & 69.724 (66.480-73.520) & 7.275 (7.090-7.480) &  \\
    \multicolumn{1}{l|}{Senior} & \multicolumn{1}{l|}{P5} & \multicolumn{1}{l|}{Malignancy Detection (Frozen)} & WithAI & 0.985 (0.961-1.000) & 43.389 (39.809-47.301) & 9.689 (9.560-9.810) &  \\
    \multicolumn{1}{l|}{Senior} & \multicolumn{1}{l|}{P5} & \multicolumn{1}{l|}{Malignancy Detection (Frozen)} & WithoutAI & 0.985 (0.961-1.000) & 50.608 (46.170-55.500) & 8.172 (8.050-8.290) &  \\
    \multicolumn{1}{l|}{Senior} & \multicolumn{1}{l|}{P6} & \multicolumn{1}{l|}{Malignancy Detection (Frozen)} & WithAI & 0.978 (0.939-1.000) & 54.049 (49.838-59.178) & 9.451 (9.320-9.570) &  \\
    \multicolumn{1}{l|}{Senior} & \multicolumn{1}{l|}{P6} & \multicolumn{1}{l|}{Malignancy Detection (Frozen)} & WithoutAI & 0.948 (0.902-0.984) & 59.847 (54.859-66.013) & 8.344 (8.140-8.540) &  \\
    \multicolumn{1}{l|}{Senior} & \multicolumn{1}{l|}{P7} & \multicolumn{1}{l|}{Malignancy Detection (Frozen)} & WithAI & 0.955 (0.918-0.985) & 59.318 (57.839-60.641) & 10.000 (10.000-10.000) &  \\
    \multicolumn{1}{l|}{Senior} & \multicolumn{1}{l|}{P7} & \multicolumn{1}{l|}{Malignancy Detection (Frozen)} & WithoutAI & 0.940 (0.897-0.976) & 71.888 (71.210-72.600) & 8.468 (8.370-8.570) &  \\
    \multicolumn{1}{l|}{Senior} & \multicolumn{1}{l|}{P8} & \multicolumn{1}{l|}{Malignancy Detection (Frozen)} & WithAI & 0.985 (0.963-1.000) & 56.248 (52.580-59.670) & 9.484 (9.330-9.620) &  \\
    \multicolumn{1}{l|}{Senior} & \multicolumn{1}{l|}{P8} & \multicolumn{1}{l|}{Malignancy Detection (Frozen)} & WithoutAI & 0.962 (0.919-0.993) & 75.643 (73.420-78.020) & 9.043 (8.900-9.190) &  \\
\cmidrule{1-7}    \multicolumn{3}{c|}{\multirow{2}[2]{*}{Mean}} & WithAI & 0.959$\pm$0.021 & 66.343$\pm$23.502 & 9.098$\pm$0.649 &  \\
    \multicolumn{3}{c|}{} & WithoutAI & 0.900$\pm$0.067 & 80.220$\pm$26.725 & 7.872$\pm$0.861 &  \\
    \bottomrule
    \end{tabular}%
    }
  \label{tab:all_reader_results_frozen}%
\end{table}%

\begin{table}[htbp]
  \centering
  \captionsetup{name=Extended Data Table}
  \caption{Balanced Accuracy, Time (s), Confidence Score and their 95\% CI from 8 pathologists on two conditions (with/without AI) for pCR Assessment (Surgical) using H\&E slides only, and Balanced Accuracy from BRAVE for referecne.}
  \scalebox{0.7}{
    \begin{tabular}{c|c|c|l|c|c|c|c}
    \toprule
    \textbf{Experience} & \textbf{Pathologist} & \textbf{Task} & \multicolumn{1}{c|}{\textbf{Group}} & \textbf{Balanced\_Accuracy} & \textbf{Time (s)} & \textbf{Confidence Score} & \textbf{BAccuracy\_Model} \\
\midrule    
\multicolumn{1}{l|}{Junior} & \multicolumn{1}{l|}{P1} & \multicolumn{1}{l|}{pCR Assessment (Surgical)} & WithAI & 0.885 (0.727-1.000) & 234.521 (229.050-240.616) & 8.741 (8.308-9.154) & \multirow{18}[18]{*}{0.878 (0.828-0.925)} \\
    \multicolumn{1}{l|}{Junior} & \multicolumn{1}{l|}{P1} & \multicolumn{1}{l|}{pCR Assessment (Surgical)} & WithoutAI & 0.791 (0.620-0.931) & 279.308 (269.810-289.693) & 7.513 (7.128-7.897) &  \\
\cmidrule{1-7}    \multicolumn{1}{l|}{Junior} & \multicolumn{1}{l|}{P2} & \multicolumn{1}{l|}{pCR Assessment (Surgical)} & WithAI & 0.869 (0.709-1.000) & 371.705 (330.612-414.498) & 8.283 (7.897-8.667) &  \\
    \multicolumn{1}{l|}{Junior} & \multicolumn{1}{l|}{P2} & \multicolumn{1}{l|}{pCR Assessment (Surgical)} & WithoutAI & 0.789 (0.621-0.922) & 399.050 (357.819-443.035) & 6.999 (6.641-7.333) &  \\
\cmidrule{1-7}    \multicolumn{1}{l|}{Junior} & \multicolumn{1}{l|}{P3} & \multicolumn{1}{l|}{pCR Assessment (Surgical)} & WithAI & 0.869 (0.709-1.000) & 291.952 (268.820-315.848) & 9.208 (9.000-9.410) &  \\
    \multicolumn{1}{l|}{Junior} & \multicolumn{1}{l|}{P3} & \multicolumn{1}{l|}{pCR Assessment (Surgical)} & WithoutAI & 0.798 (0.628-0.950) & 328.229 (303.728-355.165) & 8.028 (7.821-8.231) &  \\
\cmidrule{1-7}    \multicolumn{1}{l|}{Junior} & \multicolumn{1}{l|}{P4} & \multicolumn{1}{l|}{pCR Assessment (Surgical)} & WithAI & 0.944 (0.812-1.000) & 209.496 (198.047-221.675) & 9.153 (8.923-9.385) &  \\
    \multicolumn{1}{l|}{Junior} & \multicolumn{1}{l|}{P4} & \multicolumn{1}{l|}{pCR Assessment (Surgical)} & WithoutAI & 0.914 (0.773-1.000) & 253.677 (244.486-261.642) & 8.437 (8.103-8.769) &  \\
\cmidrule{1-7}    \multicolumn{1}{l|}{Senior} & \multicolumn{1}{l|}{P5} & \multicolumn{1}{l|}{pCR Assessment (Surgical)} & WithAI & 0.982 (0.944-1.000) & 216.381 (191.968-243.408) & 9.758 (9.622-9.892) &  \\
    \multicolumn{1}{l|}{Senior} & \multicolumn{1}{l|}{P5} & \multicolumn{1}{l|}{pCR Assessment (Surgical)} & WithoutAI & 0.982 (0.944-1.000) & 244.861 (216.703-272.135) & 8.573 (8.297-8.838) &  \\
\cmidrule{1-7}    \multicolumn{1}{l|}{Senior} & \multicolumn{1}{l|}{P6} & \multicolumn{1}{l|}{pCR Assessment (Surgical)} & WithAI & 0.930 (0.800-1.000) & 209.551 (199.792-219.822) & 9.620 (9.359-9.821) &  \\
    \multicolumn{1}{l|}{Senior} & \multicolumn{1}{l|}{P6} & \multicolumn{1}{l|}{pCR Assessment (Surgical)} & WithoutAI & 0.898 (0.764-0.984) & 284.959 (270.919-300.054) & 9.055 (8.692-9.385) &  \\
\cmidrule{1-7}    \multicolumn{1}{l|}{Senior} & \multicolumn{1}{l|}{P7} & \multicolumn{1}{l|}{pCR Assessment (Surgical)} & WithAI & 0.917 (0.762-1.000) & 218.305 (212.324-223.974) & 9.081 (8.865-9.297) &  \\
    \multicolumn{1}{l|}{Senior} & \multicolumn{1}{l|}{P7} & \multicolumn{1}{l|}{pCR Assessment (Surgical)} & WithoutAI & 0.917 (0.762-1.000) & 272.754 (266.351-280.328) & 7.890 (7.649-8.135) &  \\
\cmidrule{1-7}    \multicolumn{1}{l|}{Senior} & \multicolumn{1}{l|}{P8} & \multicolumn{1}{l|}{pCR Assessment (Surgical)} & WithAI & 0.944 (0.812-1.000) & 202.356 (183.456-221.027) & 9.586 (9.359-9.769) &  \\
    \multicolumn{1}{l|}{Senior} & \multicolumn{1}{l|}{P8} & \multicolumn{1}{l|}{pCR Assessment (Surgical)} & WithoutAI & 0.873 (0.711-1.000) & 277.501 (265.280-290.565) & 8.764 (8.487-9.026) &  \\
\cmidrule{1-7}    \multicolumn{3}{c|}{\multirow{2}[2]{*}{Mean}} & WithAI & 0.918$\pm$0.038 & 244.283$\pm$55.036 & 9.179$\pm$0.461 &  \\
    \multicolumn{3}{c|}{} & WithoutAI & 0.870$\pm$0.067 & 292.542$\pm$46.438 & 8.157$\pm$0.639 &  \\
    \bottomrule
    \end{tabular}%
}
  \label{tab:all_reader_results_surg}%
\end{table}%

\begin{table}[htbp]
  \centering
  \captionsetup{name=Extended Data Table}
  \caption{Outcomes with GEE analyses for 3 tasks based on the reference of without AI condition.}
    \begin{tabular}{l|c|l|l|l|c}
    \toprule
    \multicolumn{1}{c|}{\textbf{Task}} & \multicolumn{1}{p{6.585em}|}{\textbf{Baseline value\newline{}(Reference)}} & \multicolumn{1}{c|}{\textbf{Metric}} & \multicolumn{1}{c|}{\textbf{Value (OR/TR/Diff)}} & \multicolumn{1}{c|}{\textbf{95\% CI}} & \textbf{P-value} \\
    \midrule
    Malignancy Detection (Biopsy) & 1.0   & Accuracy (OR) & 4.8827 & 2.944 - 8.099 & P<0.001 \\
    Malignancy Detection (Biopsy) & 1.0   & Time Ratio & 0.8552 & 0.845 - 0.866 & P<0.001 \\
    Malignancy Detection (Biopsy) & 0.0   & Confidence Diff & 1.0100 & 0.943 - 1.077 & P<0.001 \\
    \midrule
    Malignancy Detection (Frozen) & 1.0   & Accuracy (OR) & 2.5032 & 1.501 - 4.175 & P<0.001 \\
    Malignancy Detection (Frozen) & 1.0   & Time Ratio & 0.8073 & 0.792 - 0.823 & P<0.001 \\
    Malignancy Detection (Frozen) & 0.0   & Confidence Diff & 1.2262 & 1.161 - 1.291 & P<0.001 \\
    \midrule
    pCR Assessment (Surgical) & 1.0   & Accuracy (OR) & 2.4139 & 1.009 - 5.776 & P=0.048 \\
    pCR Assessment (Surgical) & 1.0   & Time Ratio & 0.8204 & 0.808 - 0.833 & P<0.001 \\
    pCR Assessment (Surgical) & 0.0   & Confidence Diff & 1.0195 & 0.923 - 1.116 & P<0.001 \\
    \midrule
    Overall & 1.0   & Accuracy (OR) & 3.1432 & 2.215 - 4.461 & P<0.001 \\
    Overall & 1.0   & Time Ratio & 0.8292 & 0.821 - 0.838 & P<0.001 \\
    Overall & 0.0   & Confidence Diff & 1.1024 & 1.058 - 1.147 & P<0.001 \\
    \bottomrule
    \end{tabular}%
  \label{tab:gee}%
\end{table}%

\begin{table}[htbp]
  \centering
  \captionsetup{name=Extended Data Table}
  \caption{Inter-rater agreement among pathologists with and without AI assistance.}
    \begin{tabular}{c|c|c|c}
    \toprule
    \textbf{Task} & \textbf{Group} & \multicolumn{1}{c|}{\textbf{Fleiss's $\kappa$ (95\%CI)}} & \textbf{P-value} \\
    \midrule
    \multirow{3}[2]{*}{Malignancy Detection (Biopsy)} & Unassisted & 0.556 (0.452-0.653) & \multirow{3}[2]{*}{$P\leq$0.001} \\
          & AI-Assisted & 0.876 (0.804-0.935) &  \\
          & Difference & 0.320 (0.240-0.399) &  \\
    \midrule
    \multirow{3}[2]{*}{Malignancy Detection (Frozen)} & Unassisted & 0.592 (0.500-0.686) & \multirow{3}[2]{*}{$P\leq$0.001} \\
          & AI-Assisted & 0.841 (0.769-0.905) &  \\
          & Difference & 0.249 (0.166-0.332) &  \\
    \midrule
    \multirow{3}[2]{*}{pCR Assessment (Surgical)} & Unassisted & 0.534 (0.374-0.683) & \multirow{3}[2]{*}{$P\leq$0.001} \\
          & AI-Assisted & 0.841 (0.655-0.966) &  \\
          & Difference & 0.306 (0.121-0.505) &  \\
    \bottomrule
    \end{tabular}%
  \label{tab:kappa}%
\end{table}%

\begin{table}
  \centering
  \captionsetup{name=Extended Data Table}
  \caption{\textbf{Sequenced Effects Analysis of Pathologist Performance in the AI-Assisted Crossover Reader Study.} This table summarizes the GEE-based model estimates for sequenced effects across three diagnostic metrics: Diagnostic Accuracy (Odds Ratio, OR), Mean Reading Time (Time Ratio, TR), and Mean Confidence Level (Mean Difference).
  Values are presented as Estimate (95\% Confidence Interval). P-values indicate the statistical significance of the sequenced effects.}
  \scalebox{0.8}{  
  \begin{tabular}{c|c|c|c|c}
    \toprule
    \textbf{Task} & \textbf{Overall} & \textbf{Malignancy Detection (Biopsy)} & \textbf{Malignancy Detection (Frozen)} & \textbf{pCR Assessment (Surgical)} \\
    \midrule
    \textbf{Accuracy (OR)} & 0.961 (0.760-1.215) & 0.896 (0.590-1.361) & 1.013 (0.726-1.413) & 0.913 (0.513-1.625) \\
    \midrule
    \textbf{Confidence (Diff)} & 0.028 (-0.014-0.070) & 0.088 (0.029-0.148) & -0.029 (-0.098-0.040) & 0.019 (-0.086-0.125) \\
    \midrule
    \textbf{Time (TR)} & 1.020 (1.010-1.030) & 1.036 (1.022-1.050) & 1.014 (0.996-1.033) & 0.992 (0.974-1.011) \\
    \midrule
    \textbf{Accuracy P-value} & 0.739 & 0.607 & 0.940 & 0.757 \\
    \midrule
    \textbf{Confidence P-value} & 0.191 & 0.003 & 0.407 & 0.717 \\
    \midrule
    \textbf{Time P-value} & <0.001 & <0.001 & 0.128 & 0.413 \\
    \bottomrule
    \end{tabular}%
  }
  \label{tab:seq_effect_stat}%
\end{table}%

\begin{table}[htbp]
  \centering
  \captionsetup{name=Extended Data Table}
  \caption{\textbf{Descriptive Statistics of Pathologist Performance Stratified by Task and AI-Intervention Sequence.} 
  Performance metrics are shown for each diagnostic task, further stratified by the order of AI intervention (WithAIFirst: "Yes" indicating AI-assisted session followed by independent session; 
  "No" indicating independent session followed by AI-assisted session).}
  \scalebox{0.83}{
  \begin{tabular}{l|c|c|c|c}
    \toprule
    \multicolumn{1}{c|}{\textbf{Task}} & \textbf{WithAIFirst} & \textbf{withoutAI} & \textbf{withAI} & \textbf{delta / TR (Time)} \\
    \midrule
    \rowcolor[rgb]{ .957,  .886,  .835} \multicolumn{5}{c}{\textbf{Balanced Accuracy}} \\
    \midrule
    Overall & All   & 0.884 & 0.951 & 0.066 \\
    \midrule
    Overall & Yes   & 0.899 & 0.961 & 0.062 \\
    \midrule
    Overall & No    & 0.870 & 0.941 & 0.071 \\
    \midrule
    Malignancy Detection (Biopsy) & All   & 0.883 & 0.975 & 0.091 \\
    \midrule
    Malignancy Detection (Biopsy) & Yes   & 0.894 & 0.980 & 0.086 \\
    \midrule
    Malignancy Detection (Biopsy) & No    & 0.872 & 0.969 & 0.096 \\
    \midrule
    Malignancy Detection (Frozen) & All   & 0.900 & 0.959 & 0.059 \\
    \midrule
    Malignancy Detection (Frozen) & Yes   & 0.914 & 0.965 & 0.051 \\
    \midrule
    Malignancy Detection (Frozen) & No    & 0.886 & 0.952 & 0.066 \\
    \midrule
    pCR Assessment (Surgical) & All   & 0.870 & 0.919 & 0.049 \\
    \midrule
    pCR Assessment (Surgical) & Yes   & 0.889 & 0.936 & 0.047 \\
    \midrule
    pCR Assessment (Surgical) & No    & 0.851 & 0.902 & 0.051 \\
    \midrule
    \rowcolor[rgb]{ .906,  .922,  .847} \multicolumn{5}{c}{\textbf{Time (s)}} \\
    \midrule
    Overall & All   & 157.942 & 133.012 & 0.842 \\
    \midrule
    Overall & Yes   & 155.140 & 131.747 & 0.849 \\
    \midrule
    Overall & No    & 160.745 & 134.277 & 0.835 \\
    \midrule
    Malignancy Detection (Biopsy) & All   & 101.089 & 88.471 & 0.875 \\
    \midrule
    Malignancy Detection (Biopsy) & Yes   & 97.570 & 84.317 & 0.864 \\
    \midrule
    Malignancy Detection (Biopsy) & No    & 104.607 & 92.624 & 0.885 \\
    \midrule
    Malignancy Detection (Frozen) & All   & 80.193 & 66.298 & 0.827 \\
    \midrule
    Malignancy Detection (Frozen) & Yes   & 73.985 & 60.904 & 0.823 \\
    \midrule
    Malignancy Detection (Frozen) & No    & 86.402 & 71.692 & 0.830 \\
    \midrule
    pCR Assessment (Surgical) & All   & 292.546 & 244.267 & 0.835 \\
    \midrule
    pCR Assessment (Surgical) & Yes   & 293.865 & 250.019 & 0.851 \\
    \midrule
    pCR Assessment (Surgical) & No    & 291.226 & 238.515 & 0.819 \\
    \midrule
    \rowcolor[rgb]{ .851,  .922,  .953} \multicolumn{5}{c}{\textbf{Confidence}} \\
    \midrule
    Overall & All   & 8.103 & 9.189 & 1.086 \\
    \midrule
    Overall & Yes   & 8.033 & 9.093 & 1.060 \\
    \midrule
    Overall & No    & 8.172 & 9.285 & 1.112 \\
    \midrule
    Malignancy Detection (Biopsy) & All   & 8.276 & 9.287 & 1.011 \\
    \midrule
    Malignancy Detection (Biopsy) & Yes   & 8.275 & 9.197 & 0.922 \\
    \midrule
    Malignancy Detection (Biopsy) & No    & 8.278 & 9.378 & 1.100 \\
    \midrule
    Malignancy Detection (Frozen) & All   & 7.875 & 9.101 & 1.226 \\
    \midrule
    Malignancy Detection (Frozen) & Yes   & 7.631 & 8.886 & 1.255 \\
    \midrule
    Malignancy Detection (Frozen) & No    & 8.119 & 9.316 & 1.197 \\
    \midrule
    pCR Assessment (Surgical) & All   & 8.157 & 9.178 & 1.022 \\
    \midrule
    pCR Assessment (Surgical) & Yes   & 8.193 & 9.196 & 1.002 \\
    \midrule
    pCR Assessment (Surgical) & No    & 8.120 & 9.161 & 1.041 \\
    \bottomrule
    \end{tabular}%
  }
  \label{tab:seq_effect_desc}%
\end{table}%

\begin{table*}
  \centering
  \captionsetup{name=Extended Data Table}
  \caption{\textbf{Performance (C-Index) on Survival Prediction of BRAVE} for DFS and OS on 7 cohorts from 5 centers. 95\% CI is included in parentheses. * represents internal cohorts, otherwise external cohorts.}
    \begin{tabular}{l|l|l|l}
    \toprule
    Stage & Task  & Center & C-Index \\
    \midrule
    Pre   & DFS-Pre & H5+H8* & 0.6581 (0.4189, 0.8438) \\
    Pre   & OS-Pre & H5*  & 0.7520 (0.6095, 0.8874) \\
    \midrule
    Post  & DFS-Post & H6*   & 0.6467 (0.4461, 0.8160) \\
    Post  & DFS-Post & H13  & 0.6958 (0.2984, 0.9913) \\
    Post  & OS-Post & H6*   & 0.6836 (0.4941, 0.8403) \\
    Post  & OS-Post & H13  & 0.8294 (0.2438, 1.0000) \\
    Post  & OS-Post & H9 & 0.6223 (0.3446, 0.8693) \\
    \bottomrule
    \end{tabular}%
  \label{tab:surv_results}%
\end{table*}%

\begin{table*}
  \centering
  \captionsetup{name=Extended Data Table}
  \caption{\textbf{Data Distribution} of 8 cohorts from 5 centers for DFS and OS, including the number of cases (n\_cases), uncensored cases (n\_uncensored) and censored cases (n\_censored).}
    \begin{tabular}{c|c|c|c|c}
    \toprule
    Task  & Center & n\_cases & n\_uncensored & n\_censored \\
    \midrule
    DFS-Pre & H5    & 246   & 49    & 197 \\
    DFS-Pre & H8    & 57    & 11    & 46 \\
    DFS-Post & H6    & 451   & 71    & 380 \\
    DFS-Post & H13   & 79    & 3     & 76 \\
    OS-Pre & H5    & 1503  & 106   & 1397 \\
    OS-Post & H6    & 451   & 59    & 392 \\
    OS-Post & H13   & 79    & 2     & 77 \\
    OS-Post & H9    & 140   & 8     & 132 \\
    \bottomrule
    \end{tabular}%
  \label{tab:task_dist_surv_data}%
\end{table*}%

\begin{table*}
  \centering
  \captionsetup{name=Extended Data Table}
  \caption{\textbf{Hazard Ratios} via uni-variate analysis of various variables and BRAVE for DFS from Pre data in H5.  Neg. and Pos. refer to Negative and Positive, respectively. 95\% CI is included in parentheses.}
    \begin{tabular}{l|l|p{3.415em}|r|l}
    \toprule
    Variable & HR    & P-value & \multicolumn{1}{l|}{n\_samples} & Reference \\
    \midrule
    BRAVE & 3.9531 (2.0498,  5.8563) & P<0.001 & 234   & High-risk vs Low-risk \\
    IHC-PR & 1.7654 (0.9664,  2.5644) & P=0.064 & 234   & Negative vs Positive \\
    \midrule
    BRAVE & 4.0079 (2.0783,  5.9374) & P<0.001 & 235   & High-risk vs Low-risk \\
    IHC-ER & 2.6422 (1.4869,  3.7975) & P<0.001 & 235   & Negative vs Positive \\
    \midrule
    BRAVE & 3.3954 (1.7280,  5.0628) & P<0.001 & 204   & High-risk vs Low-risk \\
    IHC-HER2 & 1.1960 (0.6382,  1.7539) & P=0.576 & 204   & High Expression vs Low Expression \\
    \midrule
    BRAVE & 3.9376 (2.0374,  5.8379) & P<0.001 & 235   & High-risk vs Low-risk \\
    IHC-KI67 & 1.4836 (0.6635,  2.3037) & P=0.337 & 235   & High Expression vs Low Expression \\
    \midrule
    BRAVE & 3.6234 (1.9155,  5.3313) & P<0.001 & 245   & High-risk vs Low-risk \\
    TNM-Stage & 1.8239 (1.0102,  2.6375) & P=0.046 & 245   & III+IV vs I+II \\
    \midrule
    BRAVE (all) & 3.1339 (1.7814,  4.4863) & P<0.001 & 303   & High-risk vs Low-risk \\
    \bottomrule
    \end{tabular}%
  \label{tab:surv_uni_hr_dfs}%
\end{table*}%

\begin{table*}
  \centering
  \captionsetup{name=Extended Data Table}
  \caption{\textbf{Hazard Ratios} via multi-variate analysis of various variables and BRAVE for DFS from Pre data in H5.  Neg. and Pos. refer to Negative and Positive, respectively. 95\% CI is included in parentheses.}
    \begin{tabular}{l|l|p{3.415em}|r|l}
    \toprule
    \multicolumn{1}{c|}{Variable} & \multicolumn{1}{c|}{HR} & \multicolumn{1}{c|}{P-value} & \multicolumn{1}{c|}{n\_samples} & \multicolumn{1}{c}{Reference} \\
    \midrule
    IHC-ER & 0.2168 (0.1022,  0.3315) & P<0.001 & 303   & High- vs Low-risk \\
    IHC-HER2 & 1.0473 (0.5730,  1.5216) & P=0.881 & 303   & High- vs Low-risk \\
    IHC-KI67 & 1.1431 (0.5093,  1.7769) & P=0.746 & 303   & High- vs Low-risk \\
    IHC-PR & 2.7345 (1.2794,  4.1895) & P=0.009 & 303   & High- vs Low-risk \\
    BRAVE & 4.7872 (2.5304,  7.0441) & P<0.001 & 303   & High- vs Low-risk \\
    \bottomrule
    \end{tabular}%
  \label{tab:surv_multi_hr_dfs}%
\end{table*}%

\clearpage

\begin{center}
  \captionsetup{type=table,name=Extended Data Table}
  \captionof{table}{\textbf{Hazard Ratios} via uni-variate of various variables and BRAVE for OS from Pre data in H5.  Neg. and Pos. refer to Negative and Positive, respectively. 95\% CI is included in parentheses.}
  \label{tab:surv_uni_hr_os}%
    \begin{tabular}{l|l|p{5em}|r|l}
    \toprule
    Variable & HR    & P-value & \multicolumn{1}{l|}{n\_samples} & Reference \\
    \midrule
    BRAVE & 6.9527 (3.7086,  10.1968) & P<0.001 & 1457  & High- vs Low-risk \\
    IHC-ER & 1.5512 (1.0413,  2.0611) & P=0.031 & 1457  & Neg. vs Pos. \\
    \midrule
    BRAVE & 6.9362 (3.6985,  10.1739) & P<0.001 & 1456  & High- vs Low-risk \\
    IHC-PR & 1.6683 (1.1047,  2.2318) & P=0.015 & 1456  & Neg. vs Pos. \\
    \midrule
    BRAVE & 7.9503 (3.9729,  11.9276) & P<0.001 & 1301  & High- vs Low-risk \\
    IHC-HER2 & 1.1675 (0.7535,  1.5815) & P=0.488 & 1301  & Pos. vs Neg. \\
    \midrule
    BRAVE & 6.9526 (3.7085,  10.1968) & P<0.001 & 1453  & High- vs Low-risk \\
    IHC-KI67 & 0.7232 (0.4708,  0.9757) & P=0.139 & 1453  & Pos. vs Neg. \\
    \midrule
    BRAVE & 6.0598 (3.3848,  8.7348) & P<0.001 & 1487  & High- vs Low-risk \\
    TNM-Stage & 1.9910 (1.3585,  2.6234) & P<0.001 & 1487  & III+IV vs I+II \\
    \midrule
    BRAVE & 7.7599 (3.5489,  11.9708) & P<0.001 & 945   & High- vs Low-risk \\
    TNM-N & 1.9458 (1.1145,  2.7771) & P=0.019 & 945   & N+ vs N0 \\
    \midrule
    BRAVE (all) & 6.1286 (3.4232,  8.8339) & P<0.001 & 1503  & High- vs Low-risk \\
    \bottomrule
    \end{tabular}%
\end{center}

\begin{center}
  \captionsetup{type=table,name=Extended Data Table}
  \captionof{table}{\textbf{Hazard Ratios} via uni-variate of various variables and BRAVE for OS from Pre data in H5.  Neg. and Pos. refer to Negative and Positive, respectively. 95\% CI is included in parentheses.}
  \label{tab:surv_multi_hr_os}%
    \begin{tabular}{l|l|p{3.415em}|r|l}
    \toprule
    \multicolumn{1}{c|}{Variable} & \multicolumn{1}{c|}{HR} & \multicolumn{1}{c|}{P-value} & \multicolumn{1}{c|}{n\_samples} & \multicolumn{1}{c}{Reference} \\
    \midrule
    IHC-ER & 0.8162 (0.4588,  1.1737) & P=0.490 & 1265  & High- vs Low-risk \\
    IHC-HER2 & 1.0107 (0.6345,  1.3870) & P=0.964 & 1265  & High- vs Low-risk \\
    IHC-KI67 & 0.7408 (0.4511,  1.0305) & P=0.236 & 1265  & High- vs Low-risk \\
    IHC-PR & 0.8561 (0.4812,  1.2311) & P=0.597 & 1265  & High- vs Low-risk \\
    BRAVE & 8.1446 (3.9002,  12.3889) & P<0.001 & 1265  & High- vs Low-risk \\
    \bottomrule
    \end{tabular}%
\end{center}

\begin{center}
  \captionsetup{type=table,name=Extended Data Table}
  \captionof{table}{\textbf{Performance Comparison of Different Pathology Foundation Models on Pre-operative Classification Tasks in the Retrospective Cohorts.} The best performance for each task is highlighted in bold. The values are presented as mean (95\% confidence interval).}
  \label{tab:pre_cmp}%
  \scalebox{0.75}{
  \begin{tabular}{l|l|l|l|l|l|l}
    \toprule
    \multicolumn{1}{c|}{\textbf{Task}} & \multicolumn{1}{c|}{\textbf{Center}} & \multicolumn{1}{c|}{\textbf{PLIP}} & \multicolumn{1}{c|}{\textbf{CONCH}} & \multicolumn{1}{c|}{\textbf{UNI}} & \multicolumn{1}{c|}{\textbf{Virchow2}} & \multicolumn{1}{c}{\textbf{BRAVE}} \\
    \midrule
    Malignancy Detection (Bio) & H4 (Internal) & \textbf{1.000 (1.000-1.000)} & \textbf{1.000 (1.000-1.000)} & \textbf{1.000 (1.000-1.000)} & \textbf{1.000 (1.000-1.000)} & \textbf{1.000 (1.000-1.000)} \\
    \midrule
    Mass Differentiation (Bio) & H4 (Internal) & \textbf{1.000 (1.000-1.000)} & 0.980 (0.939-1.000) & \textbf{1.000 (1.000-1.000)} & \textbf{1.000 (1.000-1.000)} & \textbf{1.000 (0.998-1.000)} \\
    \midrule
    Metastasis Detection (Bio) & H5 (Internal) & 0.819 (0.556-1.000) & 0.840 (0.566-1.000) & 0.930 (0.831-1.000) & 0.854 (0.625-1.000) & \textbf{0.940 (0.820-1.000)} \\
    \midrule
    ER Prediction (Bio) & H5 (Internal) & 0.772 (0.715-0.827) & 0.838 (0.788-0.886) & 0.865 (0.818-0.910) & 0.869 (0.827-0.910) & \textbf{0.871 (0.826-0.912)} \\
    ER Prediction (Bio) & H10-Retro & 0.766 (0.727-0.804) & 0.804 (0.765-0.840) & 0.830 (0.802-0.868) & 0.833 (0.805-0.871) & \textbf{0.840 (0.805-0.874)} \\
    ER Prediction (Bio) & H17-Retro & 0.702 (0.660-0.746) & 0.796 (0.756-0.835) & 0.783 (0.744-0.819) & 0.817 (0.782-0.851) & \textbf{0.829 (0.798-0.862)} \\
    ER Prediction (Bio) & H19-Retro & 0.694 (0.643-0.742) & 0.757 (0.707-0.803) & 0.777 (0.731-0.822) & \textbf{0.808 (0.777-0.858)} & 0.801 (0.760-0.841) \\
    ER Prediction (Bio) & H20-Retro & 0.771 (0.738-0.802) & 0.793 (0.763-0.821) & 0.830 (0.804-0.855) & 0.821 (0.807-0.853) & \textbf{0.835 (0.810-0.860)} \\
    ER Prediction (Bio) & H21-Retro & 0.670 (0.573-0.764) & 0.750 (0.655-0.837) & 0.790 (0.701-0.861) & 0.833 (0.790-0.912) & \textbf{0.840 (0.772-0.900)} \\
    ER Prediction (Bio) & H4-Retro & 0.794 (0.699-0.879) & 0.769 (0.668-0.856) & \textbf{0.846 (0.799-0.928)} & 0.843 (0.788-0.921) & 0.843 (0.764-0.910) \\
    ER Prediction (Bio) & H7-Retro & 0.643 (0.586-0.697) & 0.727 (0.679-0.771) & 0.639 (0.585-0.693) & 0.749 (0.703-0.791) & \textbf{0.789 (0.749-0.826)} \\
    ER Prediction (Bio) & H8-Retro & 0.793 (0.730-0.854) & 0.831 (0.773-0.883) & 0.890 (0.846-0.928) & 0.885 (0.841-0.925) & \textbf{0.894 (0.847-0.935)} \\
    \midrule
    PR Prediction (Bio) & H5 (Internal) & 0.708 (0.639-0.775) & 0.734 (0.669-0.796) & 0.759 (0.698-0.817) & 0.771 (0.715-0.826) & \textbf{0.773 (0.709-0.829)} \\
    PR Prediction (Bio) & H10-Retro & 0.638 (0.591-0.684) & 0.755 (0.716-0.798) & 0.796 (0.759-0.831) & 0.822 (0.787-0.853) & \textbf{0.823 (0.789-0.854)} \\
    PR Prediction (Bio) & H17-Retro & 0.593 (0.548-0.638) & 0.728 (0.689-0.764) & 0.740 (0.702-0.776) & 0.746 (0.709-0.781) & \textbf{0.762 (0.725-0.798)} \\
    PR Prediction (Bio) & H4-Retro & 0.733 (0.637-0.816) & 0.750 (0.667-0.835) & 0.773 (0.687-0.853) & \textbf{0.812 (0.736-0.886)} & 0.811 (0.735-0.887) \\
    PR Prediction (Bio) & H8-Retro & 0.716 (0.641-0.782) & 0.761 (0.698-0.825) & 0.793 (0.726-0.850) & \textbf{0.810 (0.748-0.868)} & 0.802 (0.738-0.859) \\
    \midrule
    HER2 Prediction (Bio) & H5 (Internal) & 0.757 (0.697-0.816) & 0.804 (0.750-0.858) & 0.791 (0.735-0.845) & 0.801 (0.745-0.855) & \textbf{0.824 (0.771-0.872)} \\
    HER2 Prediction (Bio) & H10-Retro & 0.671 (0.630-0.714) & 0.734 (0.697-0.773) & 0.753 (0.712-0.791) & 0.768 (0.730-0.805) & \textbf{0.791 (0.755-0.827)} \\
    HER2 Prediction (Bio) & H4-Retro & 0.704 (0.569-0.825) & 0.716 (0.588-0.836) & 0.757 (0.633-0.872) & \textbf{0.818 (0.694-0.918)} & 0.801 (0.676-0.917) \\
    \midrule
    KI67 Prediction (Bio) & H5 (Internal) & 0.774 (0.718-0.828) & 0.810 (0.754-0.864) & \textbf{0.819 (0.759-0.872)} & 0.797 (0.733-0.858) & 0.816 (0.753-0.873) \\
    KI67 Prediction (Bio) & H17-Retro & 0.547 (0.480-0.615) & 0.701 (0.646-0.753) & 0.762 (0.725-0.816) & \textbf{0.767 (0.731-0.822)} & 0.765 (0.723-0.807) \\
    KI67 Prediction (Bio) & H4-Retro & 0.833 (0.742-0.919) & \textbf{0.875 (0.809-0.933)} & 0.832 (0.725-0.932) & 0.798 (0.659-0.920) & 0.843 (0.739-0.922) \\
    \midrule
    Molecular Subtyping (Bio) & H5 (Internal) & 0.817 (0.731-0.893) & 0.824 (0.741-0.899) & 0.822 (0.735-0.890) & 0.837 (0.752-0.911) & \textbf{0.841 (0.756-0.912)} \\
    Molecular Subtyping (Bio) & H4-Retro & 0.932 (0.853-0.993) & \textbf{0.968 (0.914-1.000)} & 0.942 (0.870-0.996) & 0.961 (0.896-1.000) & 0.967 (0.910-1.000) \\
    \midrule
    NAC Response Prediction (Bio) & H5 (Internal) & 0.681 (0.612-0.745) & 0.730 (0.656-0.792) & 0.731 (0.663-0.797) & 0.688 (0.611-0.763) & \textbf{0.740 (0.671-0.805)} \\
    \bottomrule
    \end{tabular}%
  }
\end{center}

\begin{center}
  \captionsetup{type=table,name=Extended Data Table}
  \captionof{table}{\textbf{Performance Comparison of Different Pathology Foundation Models on Intra-operative Classification Tasks in the Retrospective Cohorts.} The best performance for each task is highlighted in bold. The values are presented as mean (95\% confidence interval).}
  \label{tab:intra_cmp}%
  \scalebox{0.7}{%
    \begin{tabular}{l|l|l|l|l|l|l}
    \toprule
    \multicolumn{1}{c|}{\textbf{Task}} & \multicolumn{1}{c|}{\textbf{Center}} & \multicolumn{1}{c|}{\textbf{PLIP}} & \multicolumn{1}{c|}{\textbf{CONCH}} & \multicolumn{1}{c|}{\textbf{UNI}} & \multicolumn{1}{c|}{\textbf{Virchow2}} & \multicolumn{1}{c}{\textbf{BRAVE}} \\
    \midrule
    Malignancy Detection (Frozen) & H2 (Internal) & 0.964 (0.927-0.990) & 0.999 (0.996-1.000) & \textbf{1.000 (0.998-1.000)} & 0.998 (0.991-1.000) & 0.999 (0.996-1.000) \\
    Malignancy Detection (Frozen) & H12+H16-Retro & 0.668 (0.550-0.770) & 0.902 (0.811-0.973) & 0.934 (0.847-0.986) & 0.923 (0.839-0.975) & \textbf{0.936 (0.879-0.978)} \\
    Malignancy Detection (Frozen) & H15-Retro & 0.779 (0.691-0.865) & 0.794 (0.702-0.880) & \textbf{0.925 (0.862-0.972)} & 0.901 (0.836-0.956) & 0.908 (0.842-0.965) \\
    Malignancy Detection (Frozen) & H4-Retro & 0.689 (0.663-0.716) & 0.926 (0.913-0.940) & 0.938 (0.925-0.951) & 0.950 (0.939-0.960) & \textbf{0.985 (0.980-0.990)} \\
    \midrule
    Mass Differentiation (Frozen) & H4 (Internal) & 0.963 (0.932-0.989) & 0.983 (0.967-0.996) & 0.991 (0.978-1.000) & 0.987 (0.974-0.997) & \textbf{0.994 (0.986-0.999)} \\
    \midrule
    Margin Assessment (Frozen) & H2 (Internal) & 0.816 (0.712-0.917) & 0.900 (0.811-0.964) & 0.885 (0.790-0.965) & 0.862 (0.731-0.962) & \textbf{0.927 (0.862-0.981)} \\
    Margin Assessment (Frozen) & H12-Retro & 0.906 (0.841-0.955) & 0.929 (0.838-0.990) & 0.965 (0.913-0.995) & 0.964 (0.930-0.987) & \textbf{0.973 (0.948-0.990)} \\
    Margin Assessment (Frozen) & H4-Retro & 0.897 (0.843-0.945) & \textbf{0.942 (0.898-0.978)} & 0.836 (0.786-0.879) & 0.862 (0.800-0.912) & 0.909 (0.831-0.969) \\
    \midrule
    Lymph Node Metastasis Prediction (Frozen) & H2 (Internal) & \textbf{0.980 (0.944-0.999)} & 0.972 (0.922-1.000) & 0.974 (0.920-1.000) & 0.979 (0.931-1.000) & 0.976 (0.923-1.000) \\
    Lymph Node Metastasis Prediction (Frozen) & H12-Retro & 0.869 (0.812-0.934) & 0.898 (0.840-0.955) & \textbf{0.920 (0.891-0.989)} & 0.913 (0.886-0.975) & \textbf{0.920 (0.844-0.980)} \\
    Lymph Node Metastasis Prediction (Frozen) & H4-Retro & 0.905 (0.845-0.954) & \textbf{0.936 (0.893-0.968)} & 0.931 (0.884-0.975) & 0.915 (0.870-0.955) & 0.919 (0.869-0.959) \\
    \bottomrule
    \end{tabular}%
  }
\end{center}

\clearpage

\begin{center}
  \captionsetup{type=table,name=Extended Data Table}
  \captionof{table}{\textbf{Performance Comparison of Different Pathology Foundation Models on Post-operative Classification Tasks in the Retrospective Cohorts (Part 1).} The best performance for each task is highlighted in bold. The values are presented as mean (95\% confidence interval).}
  \label{tab:post_cmp}%
  \scalebox{0.67}{%
  \begin{tabular}{l|l|l|l|l|l|l}
    \toprule
    \multicolumn{1}{c|}{\textbf{Task}} & \multicolumn{1}{c|}{\textbf{Center}} & \multicolumn{1}{c|}{\textbf{PLIP}} & \multicolumn{1}{c|}{\textbf{CONCH}} & \multicolumn{1}{c|}{\textbf{UNI}} & \multicolumn{1}{c|}{\textbf{Virchow2}} & \multicolumn{1}{c}{\textbf{BRAVE}} \\
    \midrule
    Pathological Subtyping (Surgical) & TCGA (Internal) & 0.904 (0.849-0.951) & 0.967 (0.946-0.985) & 0.965 (0.941-0.986) & \textbf{0.981 (0.964-0.994)} & 0.980 (0.962-0.993) \\
    Pathological Subtyping (Surgical) & H4-Retro & 0.746 (0.591-0.889) & 0.946 (0.904-0.981) & 0.907 (0.852-0.953) & 0.917 (0.863-0.962) & \textbf{0.949 (0.907-0.984)} \\
    \midrule
    Tumor Grading (Surgical) & H4 (Internal) & 0.719 (0.556-0.867) & 0.746 (0.592-0.882) & 0.732 (0.600-0.856) & 0.770 (0.613-0.903) & \textbf{0.803 (0.662-0.918)} \\
    \midrule
    pCR Assessment (Surgical) & H4 (Internal) & 0.668 (0.438-0.857) & 0.890 (0.739-1.000) & 0.920 (0.800-1.000) & 0.925 (0.792-1.000) & \textbf{0.949 (0.841-1.000)} \\
    \midrule
    TNM-T Stage Prediction (Surgical) & H6 (Internal) & 0.704 (0.595-0.808) & 0.782 (0.684-0.869) & 0.754 (0.645-0.845) & 0.750 (0.644-0.842) & \textbf{0.825 (0.738-0.901)} \\
    \midrule
    Lymphovascular Invasion Prediction (Surgical) & H4 (Internal) & 0.678 (0.554-0.797) & \textbf{0.745 (0.638-0.848)} & 0.744 (0.634-0.847) & 0.703 (0.585-0.814) & 0.739 (0.615-0.848) \\
    \midrule
    Perineural Invasion Prediction (Surgical) & H4 (Internal) & 0.818 (0.665-0.943) & 0.783 (0.680-0.879) & 0.788 (0.665-0.894) & 0.796 (0.678-0.892) & \textbf{0.819 (0.719-0.903)} \\
    \midrule
    AR Prediction (Surgical) & H6 (Internal) & 0.734 (0.663-0.803) & 0.687 (0.617-0.755) & 0.729 (0.660-0.798) & 0.734 (0.663-0.802) & \textbf{0.748 (0.679-0.812)} \\
    \midrule
    CK5 Prediction (Surgical) & H6 (Internal) & 0.805 (0.725-0.876) & 0.753 (0.676-0.822) & 0.834 (0.753-0.900) & 0.817 (0.739-0.886) & \textbf{0.843 (0.775-0.901)} \\
    \midrule
    ER Prediction (Surgical) & H6 (Internal) & 0.852 (0.808-0.893) & 0.889 (0.849-0.922) & 0.899 (0.862-0.933) & 0.898 (0.863-0.930) & \textbf{0.908 (0.875-0.938)} \\
    ER Prediction (Surgical) & H18-Retro & 0.708 (0.672-0.745) & 0.773 (0.743-0.804) & 0.763 (0.732-0.792) & 0.798 (0.769-0.825) & \textbf{0.834 (0.807-0.860)} \\
    ER Prediction (Surgical) & H20-Retro & 0.729 (0.656-0.796) & 0.778 (0.717-0.834) & 0.756 (0.692-0.818) & 0.776 (0.710-0.836) & \textbf{0.807 (0.746-0.863)} \\
    \midrule
    PR Prediction (Surgical) & H6 (Internal) & 0.895 (0.856-0.926) & 0.865 (0.823-0.902) & 0.893 (0.853-0.928) & 0.891 (0.853-0.928) & \textbf{0.901 (0.866-0.936)} \\
    \midrule
    HER2 Prediction (Surgical) & H6 (Internal) & 0.797 (0.748-0.845) & 0.784 (0.727-0.836) & 0.817 (0.762-0.863) & 0.835 (0.788-0.882) & \textbf{0.850 (0.801-0.891)} \\
    HER2 Prediction (Surgical) & H18-Retro & 0.767 (0.722-0.807) & 0.654 (0.608-0.702) & 0.848 (0.812-0.881) & 0.877 (0.845-0.906) & \textbf{0.895 (0.866-0.922)} \\
    \midrule
    KI67 Prediction (Surgical) & H4 (Internal) & 0.729 (0.587-0.866) & 0.840 (0.721-0.928) & 0.688 (0.530-0.832) & 0.808 (0.678-0.925) & \textbf{0.843 (0.710-0.952)} \\
    KI67 Prediction (Surgical) & H18-Retro & 0.700 (0.667-0.732) & 0.761 (0.731-0.792) & 0.777 (0.749-0.806) & \textbf{0.807 (0.779-0.834)} & 0.789 (0.761-0.817) \\
    \bottomrule
    \end{tabular}%
  }
\end{center}

\begin{center}
  \captionsetup{type=table,name=Extended Data Table}
  \captionof{table}{\textbf{Performance Comparison of Different Pathology Foundation Models on Post-operative Classification Tasks in the Retrospective Cohorts (Part 2).} The best performance for each task is highlighted in bold. The values are presented as mean (95\% confidence interval).}
  \label{tab:post_cmp_cont}%
  \scalebox{0.67}{%
  \begin{tabular}{l|l|l|l|l|l|l}
    \toprule
    \multicolumn{1}{c|}{\textbf{Task}} & \multicolumn{1}{c|}{\textbf{Center}} & \multicolumn{1}{c|}{\textbf{PLIP}} & \multicolumn{1}{c|}{\textbf{CONCH}} & \multicolumn{1}{c|}{\textbf{UNI}} & \multicolumn{1}{c|}{\textbf{Virchow2}} & \multicolumn{1}{c}{\textbf{BRAVE}} \\
    \midrule
    \midrule
    Molecular Subtyping (Surgical) & H6 (Internal) & 0.889 (0.849-0.925) & 0.865 (0.815-0.911) & \textbf{0.920 (0.886-0.950)} & 0.907 (0.874-0.938) & 0.904 (0.867-0.935) \\
    Molecular Subtyping (Surgical) & H13-Retro & 0.870 (0.837-0.902) & 0.888 (0.852-0.920) & 0.877 (0.841-0.911) & 0.895 (0.865-0.924) & \textbf{0.903 (0.870-0.931)} \\
    Molecular Subtyping (Surgical) & H4+H14-Retro & 0.836 (0.757-0.906) & 0.920 (0.859-0.965) & \textbf{0.937 (0.884-0.975)} & 0.931 (0.865-0.983) & \textbf{0.937 (0.888-0.974)} \\
    Molecular Subtyping (Surgical) & H9-Retro & 0.824 (0.800-0.849) & 0.826 (0.799-0.849) & 0.830 (0.802-0.852) & 0.840 (0.816-0.860) & \textbf{0.842 (0.801-0.878)} \\
    \midrule
    Mutation CDH1 Prediction (Surgical) & TCGA (Internal) & 0.786 (0.672-0.892) & \textbf{0.881 (0.794-0.948)} & 0.841 (0.723-0.930) & 0.860 (0.752-0.936) & 0.874 (0.745-0.953) \\
    \midrule
    Mutation GATA3 Prediction (Surgical) & TCGA (Internal) & 0.556 (0.415-0.688) & 0.592 (0.465-0.723) & 0.649 (0.532-0.761) & 0.657 (0.559-0.753) & \textbf{0.711 (0.619-0.796)} \\
    \midrule
    Mutation PIK3CA Prediction (Surgical) & TCGA (Internal) & 0.635 (0.559-0.713) & \textbf{0.651 (0.572-0.724)} & 0.638 (0.562-0.714) & 0.588 (0.506-0.664) & 0.641 (0.565-0.712) \\
    Mutation PIK3CA Prediction (Surgical) & CPTAC-Retro & 0.605 (0.492-0.713) & 0.566 (0.419-0.700) & 0.618 (0.504-0.728) & 0.635 (0.519-0.748) & \textbf{0.713 (0.595-0.814)} \\
    \midrule
    Mutation TP53 Prediction (Surgical) & TCGA (Internal) & 0.719 (0.650-0.783) & 0.803 (0.745-0.859) & 0.797 (0.740-0.852) & 0.807 (0.746-0.864) & \textbf{0.829 (0.770-0.884)} \\
    Mutation TP53 Prediction (Surgical) & CPTAC-Retro & 0.789 (0.691-0.877) & 0.801 (0.701-0.885) & 0.814 (0.726-0.895) & \textbf{0.827 (0.745-0.896)} & 0.825 (0.733-0.903) \\
    \bottomrule
    \end{tabular}%
  }
\end{center}

\begin{center}
  \captionsetup{type=table,name=Extended Data Table}
  \captionof{table}{\textbf{Performance Comparison of Different Pathology Foundation Models on Survival Prediction Tasks in the Retrospective Cohorts.} The best performance for each task is highlighted in bold. The values are presented as mean (95\% confidence interval).}
  \label{tab:surv_cmp}%
  \scalebox{0.82}{%
    \begin{tabular}{l|l|l|l|l|l|l}
    \toprule
    \multicolumn{1}{c|}{\textbf{Task}} & \multicolumn{1}{c|}{\textbf{Center}} & \multicolumn{1}{c|}{\textbf{PLIP}} & \multicolumn{1}{c|}{\textbf{CONCH}} & \multicolumn{1}{c|}{\textbf{UNI}} & \multicolumn{1}{c|}{\textbf{Virchow2}} & \multicolumn{1}{c}{\textbf{BRAVE}} \\
    \midrule
    DFS-Pre & H5+H8 (Internal) & 0.588 (0.386-0.789) & 0.643 (0.415-0.855) & 0.631 (0.412-0.861) & 0.651 (0.446-0.842) & \textbf{0.658 (0.419-0.844)} \\
    \midrule
    OS-Pre & H5 (Internal) & 0.723 (0.572-0.840) & 0.744 (0.627-0.886) & 0.738 (0.600-0.864) & 0.742 (0.607-0.875) & \textbf{0.752 (0.610-0.887)} \\
    \midrule
    DFS-Post & H6 (Internal) & 0.553 (0.356-0.742) & 0.635 (0.467-0.791) & 0.638 (0.426-0.860) & 0.641 (0.447-0.821) & \textbf{0.647 (0.446-0.816)} \\
    DFS-Post & H13-Retro & 0.644 (0.341-0.909) & 0.690 (0.425-0.978) & 0.686 (0.373-0.980) & 0.652 (0.257-0.991) & \textbf{0.696 (0.298-0.991)} \\
    \midrule
    OS-Post & H6 (Internal) & 0.637 (0.455-0.800) & 0.664 (0.475-0.829) & 0.659 (0.445-0.839) & 0.665 (0.453-0.864) & \textbf{0.684 (0.494-0.840)} \\
    OS-Post & H13-Retro & 0.761 (0.410-1.000) & 0.802 (0.739-1.000) & 0.812 (0.776-1.000) & 0.819 (0.500-1.000) & \textbf{0.829 (0.244-1.000)} \\
    OS-Post & H9-Retro & 0.551 (0.230-0.838) & 0.609 (0.316-0.846) & 0.615 (0.394-0.861) & 0.610 (0.398-0.889) & \textbf{0.622 (0.345-0.869)} \\
    \bottomrule
    \end{tabular}%
  }
\end{center}

\end{document}